\documentclass[letterpaper]{article} 
\usepackage{aaai2026}  
\usepackage{times}  
\usepackage{helvet}  
\usepackage{courier}  
\usepackage[hyphens]{url}  
\usepackage{graphicx} 
\usepackage{natbib}  
\usepackage{caption} 
\usepackage{algorithm}
\usepackage{algorithmic}

\usepackage{newfloat}
\usepackage{listings}
\DeclareCaptionStyle{ruled}{labelfont=normalfont,labelsep=colon,strut=off} 
\floatstyle{ruled}
\newfloat{listing}{tb}{lst}{}
\floatname{listing}{Listing}
\usepackage[colorinlistoftodos,prependcaption,textsize=tiny]{todonotes}
\usepackage{booktabs}
\usepackage{multirow}
\usepackage{listings}
\usepackage{graphicx}
\usepackage{url}
\usepackage{subfig}
\usepackage{amsmath}
\usepackage{caption}
\usepackage{subcaption}
\usepackage{subfig}
\usepackage{rotating}
\usepackage{dblfloatfix}
\usepackage{amssymb}
\usepackage{comment}

\newcommand{\system}{{\sc VRD-UQA}}
\newcommand{\systemext}{{\sc \textbf{V}isually \textbf{R}ich \textbf{D}ocument \textbf{U}nanswerable \textbf{Q}uestion \textbf{A}nswering}}

\title{Benchmarking Visual LLMs Resilience to \\Unanswerable Questions on Visually Rich Documents}
\author{
    Davide Napolitano,
    Luca Cagliero,
    Fabrizio Battiloro
}
\affiliations{
    Politecnico di Torino, Torino, Italy\\

    name.surname@polito.it
}

\begin{document}

\maketitle

\begin{abstract}
The evolution of Visual Large Language Models (VLLMs) has revolutionized the automatic understanding of Visually Rich Documents (VRDs), which contain both textual and visual elements.
Although VLLMs excel in Visual Question Answering (VQA) on multi-page VRDs, their ability to detect unanswerable questions is still an open research question.
Our research delves into the robustness of the VLLMs to plausible yet unanswerable questions, i.e., questions that appear valid but cannot be answered due to subtle corruptions
caused by
swaps between related concepts
or plausible question formulations. 
Corruptions are generated by replacing the original natural language entities with other ones of the same type, belonging to different document elements, and in different layout positions or pages of the related document.
To this end, we present \system\ (\systemext), a benchmark for evaluating VLLMs' resilience to plausible yet unanswerable questions across multiple dimensions. 
It automatically alters the questions of existing VQA datasets consisting of multi-page VRDs,
verifies their unanswerability using a VLLM-as-a-judge approach, and then thoroughly evaluates VLLMs' performance. 
Experiments, run on 12 models, analyze:
(1) The VLLMs' accuracy in detecting unanswerable questions at both page and document levels;
(2) The effect of different types of corruption (NLP entity, document element, layout);
(3) The effectiveness of different knowledge injection strategies based on in-context learning (OCR, multi-page selection, or the possibility of unanswerability).
Our findings reveal VLLMs' limitations and demonstrate that \system\ can serve as an 
evaluation framework for developing resilient document VQA systems. 
\end{abstract}

\begin{links}
    \link{Code}{https://github.com/DavideNapolitano/VRD-UQA}
\end{links}

\section{Introduction}
\label{sec:introduction}

Visual Large Language Models (VLLMs) are trained and specialized to produce accurate answers, in textual form, to questions about a mix of visual and textual content~\cite{Luo2024CVPR}. 
These models are particularly valuable for analyzing Visually Rich Documents (VRDs)~\cite{WangZWLT23}, i.e., documents that combine textual content (paragraphs, titles) with structured visual elements (e.g., figures, tables). 
They encompass various document types, such as PDF files and printed or scanned copies, and cover a variety of domains and sources (e.g., news, financial reports).


Visual Question Answering (VQA) from Visually Rich Documents (VRDs) is particularly challenging because it requires not only an advanced comprehension of the question 
but also the ability to link the linguistic concepts mentioned in the question 
to contents, either textual and visual, available in pages with complex layout structures or even in different pages. 
In this work, we focus on the zero-shot VQA capabilities of VLLMs on multi-page VRDs, which is one the most representative real-world scenarios.

Even though a question on a multi-page VRD may seem plausible and well-formed, its answer can be undetermined.
For example, given the question 
\textit{What is the future projection of sea level in the figure?}, document pages may not contain figures regarding sea level or the information could be embedded in different elements, like a table~\cite{davis2020unanswerable}. 

The ability of VQA models to determine whether a question is answerable or not is at least as important as providing correct and pertinent answers~\cite{GuoJSNK24,vardi2025clip}.
In our research, we aim to mimic human 
questions that are unanswerable due to small errors caused by swaps between related concepts or due to the inherent formulations. They are known to be quite common~\cite{xie-etal-2024-typos} and not as trivial to detect as small typos or meaningless sentences because they pass grammar and semantic checks~\cite{jia-liang-2017-adversarial}. 
Notice that NLP entity swaps may also involve document elements (e.g., {\it Caption} instead of {\it Footnote}) or layout information (e.g., {\it Bottom} instead of {\it Top}), making their detection even worse. 

We verify the VLLMs' robustness in correctly detecting these unanswerable cases both separately for each page and at the document level. To this end, we alter the answerable questions of the multi-page VQA datasets~\cite{MPDocVQA,LandeghemPTJBBC23} with a controlled level of corruption. 
Specifically, we recognize NLP entities in the original question and replace them with others of the same type,
within different multimodal elements, and in different layout positions or pages of the related document.

The purpose of the present work is to address the following Research Questions (RQs):

\begin{itemize}
\item RQ1: Are VLLMs capable of accurately detecting question unanswerability due to the entity corruption?
\item RQ2: What is the effect of different corruption types 
on models’ performance?
\item RQ3: Which in-context learning strategies are able to mitigate the limitations of VLLMs in identifying unanswerable questions?
\end{itemize}

To address the RQs, we propose \system\  (\systemext ), a new framework aimed to evaluate VLLMs performance in detecting unanswerable questions.
Given a multi-page document VQA dataset and a set of models,
\system\ automatically corrupts the questions, 
verifies their actual unanswerability using a VLLM-as-a-judge approach~\cite{li2024llms, zheng2023judging}, 
and then evaluates the document- and page-level accuracies of distinct models 
by analyzing the separate and combined effects of different corruption types.
The experiments carried out on 12 models and 2 datasets showcase:
\begin{itemize}
    \item The models' performance, underlying the importance of the model pretraining strategy which is paramount even compared to the number of model parameters (RQ1); 
    \item The models' strengths and weaknesses with specific NLP entities, 
    (e.g., fairly robust to perturbations of location entities, weak on document structure-related entities), 
    the variable resilience in handling document elements 
    (e.g., higher resilience with headers and footnotes, lower with tables), 
    and the difficulty to circumvent corruptions in long documents and caused by in-page entities (RQ2);
    \item The benefits of adopting in-context learning strategies, such as providing the document OCR or stating the possibility of unanswerability, to mitigate the limitations of state-of-the-art models in tackling the unanswerability detection problem (RQ3). 
\end{itemize}

\noindent The main paper contributions can be summarized as:

\begin{itemize}
\itemsep0em 
\item An open source \textbf{new evaluation framework}  (\system ) focused on evaluating VLLMs' robustness to unanswerable questions on multi-page VRDs.
\item We present a \textbf{pipeline} aimed to alter answerable questions available in VQA benchmark datasets
with controlled levels of corruptions regarding \textbf{NLP entities}, \textbf{document elements}, and \textbf{document layout}.
\item We \textbf{release} the extended and corrupted versions of the established DUDE~\cite{LandeghemPTJBBC23} and MPDocVQA~\cite{MPDocVQA}  datasets.
\item An extensive empirical \textbf{evaluation} carried out on 12 VLLMs, and 2 VQA datasets for multi-page VRDs.  
\end{itemize}

The rest of the paper is organized as follows.
Section~\ref{sec:relatedworks} reports related work.
Sections~\ref{sec:preliminaries} and~\ref{sec:corruption} introduce preliminary notions and corruption strategies.
Section~\ref{sec:methodology} describes the \system\ benchmark.
Section~\ref{sec:results} discusses the results.
Section~\ref{sec:conclusion} draws conclusions and discusses future works. 

\section{Related Work}
\label{sec:relatedworks}

Recently, the research community has released several VQA benchmarks for VRDs~\cite{DocVQA, LandeghemPTJBBC23,MPDocVQA, mathew2021infographicvqa, choi2018quacquestionanswering, deng-etal-2025-longdocurl}.
A comprehensive taxonomy can be found in~\cite{rogers2023qa}.
Parallel works have focused on assessing the VQA models' capability to detect unanswerable questions
using corrupted images and questions~\cite{GuoJSNK24, DBLP:conf/eccv/WhiteheadPS0DRR22, ICCV2023, akter-etal-2024-visreas}.
Specifically, Reliable-VQA and UNK-VQA~\cite{GuoJSNK24} are designed to handle single images or text without contextual knowledge, whereas our approach (VRD-UQA) is capable of processing documents including
multiple images. 
While VRD-UQA dynamically corrupts the input questions through a mix of NLP and multimodal learning techniques, UNK-VQA applies predefined perturbations, RGQA~\cite{ICCV2023} applies self-supervised contrastive learning to generate image-question pairs, whereas VisReas~\cite{akter-etal-2024-visreas} generates unanswerable queries using Visual Genome data.

Alternative approaches, such as 
MMLongBench-Doc~\cite{ma2024mmlongbench},  TUBench~\cite{he2024tubench}
and LongDocURL~\cite{deng-etal-2025-longdocurl}, evaluate VQA models' robustness through natively unanswerable questions. In contrast, our approach generates unanswerable questions by corrupting answerable ones. 
Furthermore, we explore multiple dimensions (e.g., document elements and layout), either separately or jointly, 
while preserving question plausibility. 

Other studies have examined the frequency and kind of typing errors~\cite{cucerzan2004spelling}, showing that entity substitution errors occur through mechanisms like autocorrect interference, phonetic similarity, and memory lapses~\cite{shi2025simulating}.
Human transcription errors~\cite{hong2013error, mays2019measuring} are alternative sources of corruption, 
which potentially preserve coherence and plausibility of the unanswerable question. 
Previous studies on the robustness of QA models~\cite{belinkov2017synthetic, ribeiro2018semantically} have shown that these models are highly sensitive to corrupted inputs, with minor substitutions causing significant performance degradation in document understanding~\cite{jia2017adversarial}. This calls for new VQA testing benchmarks aimed to evaluate VLLMs performance with corrupted questions on VRDs.

\section{Preliminaries}
\label{sec:preliminaries}

A VRD $D$ consists of one or multiple pages $p_1$,  $p_2$,  $\ldots$, $p_{|D|}$. It includes not only textual elements, such as paragraphs and headlines, but also visual elements (e.g., charts and tables). Given a natural language question $Q$ on a $D$, VQA from VRDs exploits a model to generate the answer $A$ to $Q$ based on the $D$'s content. In this work, we focus on questions $Q$ with no answer, i.e., the \textit{unanswerable questions}. We ask VLLMs to detect these cases and return 
\textit{No answer} as corresponding response.
To evaluate the models' capabilities to accurately identify unanswerable questions, we define the following experimental setting.
Firstly, we leverage the VQA model with 
(1) a specific instruction prompt, where additional information such as OCR and unanswerability information may be included,
(2) an unanswerable question and 
(3) a window sliding over the document pages (the window size $w$ is a user-specified parameter). 
Then, we verify the correctness of the provided answer (correct: \textit{No answer}, incorrect: \textit{otherwise}). Next, we repeat the test over different page windows.
Finally, we evaluate the model's performance 
according to the following performance metrics:
    (1) \textit{Document-Level Accuracy} ($Acc_D$), i.e.,  the percentage of unanswerable questions 
    for which \textit{all} the associated document page-level answers are correct. 
    (2) \textit{Page-Level Accuracy} ($Acc_P$), i.e.,  the average rate of correct page-level answers for each corrupted question. 


\section{Question Corruption}
\label{sec:corruption}

\begin{figure}[th!]
    \centering
    \includegraphics[width=0.95\linewidth]{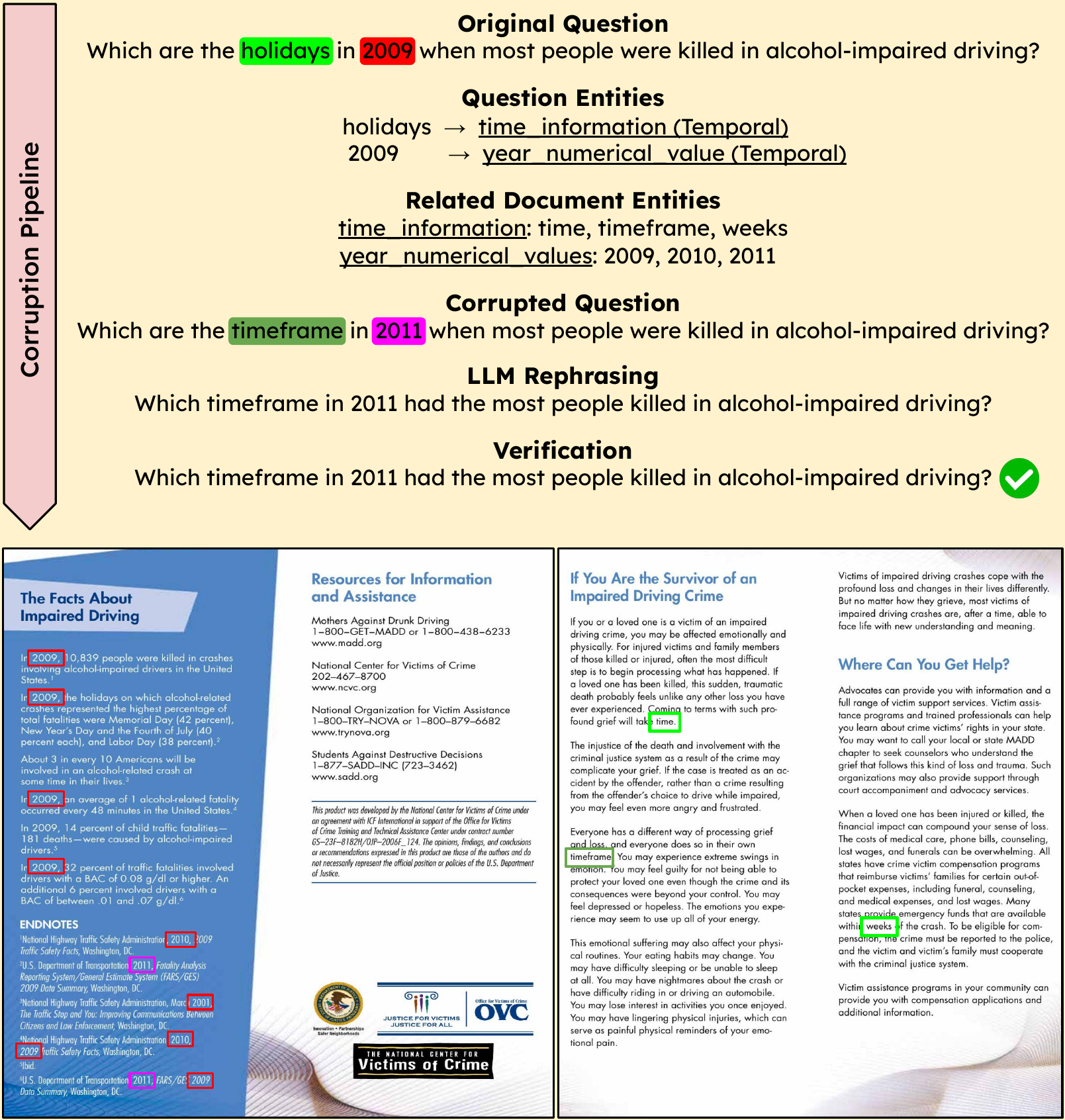}
    \caption{
    \system\ generates unanswerable questions starting from an answerable question and the reference document. Example from DUDE~\cite{LandeghemPTJBBC23}). 
    }
    \label{fig:runningexample}
\end{figure}

Questions on VRD may contain errors due to typos, misunderstandings, and memory lapses, or they may be intrinsically unanswerable.
In this work, we focus on unanswerable questions built by swapping related concepts, identified by NLP entities. These questions mimic user queries that appear plausible and contextually relevant but cannot be answered based on the VRD content.
This allows us not only to evaluate the robustness to the presence of semantically similar but incorrect entity matchings between questions and document contents but also to evaluate the effect of incorrect references to document elements or layout information. 

We design a systematic approach to generate plausible yet unanswerable questions. 
Specifically, we 
inject a controlled level of corruption into the (answerable) questions of a VQA document dataset. 
To this end, we consider (separately or mixed) three main corruption types: (1) \textit{NLP entities}, (2) \textit{Document elements}, (3) \textit{Document layout}. 
Our framework is motivated by the unique challenges of VRDs, which require models to jointly reason over textual semantics, structural composition (i.e., functional elements), and spatial layout to enable effective question answering~\cite{vardi2025clip, wang2023docllm}.
To evaluate such capabilities, we propose a multi-level corruption framework that systematically analyzes multiple facets of documents and models.

\paragraph{NLP entities} In Natural Language Processing, named entities are well-known concepts that are typically described by one or more words in the document text~\cite{ManningBook}. Based on their semantic meaning, entities are usually categorized into predefined types (e.g., numbers, temporal information).  For example, in Figure~\ref{fig:runningexample}, we highlight in red and green the identified temporal entities.
A possible typo in writing consists of replacing an entity with another of the same type, such as reporting 
\textit{2011} instead of \textit{2009} (see Figure~\ref{fig:runningexample}). 
Similar human errors in entity value specification are common in document information retrieval~\cite{shi2025simulating}.
In most cases, these subtle textual modifications would preserve the plausibility of the question, thus making the detection of unanswerable cases particularly challenging. 
Hence, we evaluate models' resilience to entity-level corruptions by replacing an entity occurring in the original question with another one of the same type occurring in the document (regardless of its relative position). This approach simulates the most challenging settings for models, as all relevant elements in the question are contained within the document.

\begin{figure*}[th!]
    \centering
    \includegraphics[width=0.87\linewidth]{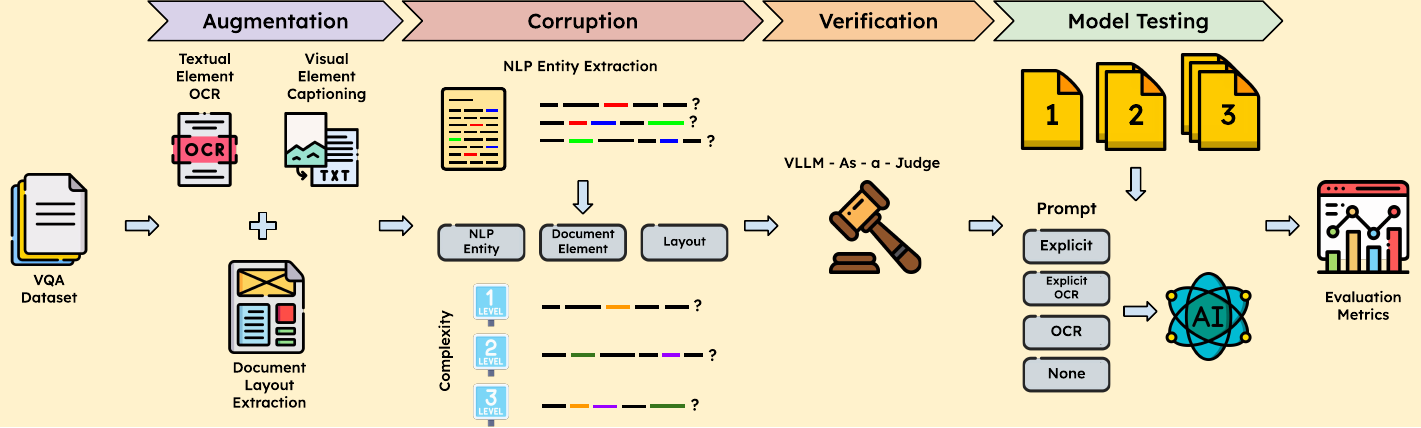}
    \caption{The  \systemext\  framework.}
    \label{fig:framework}
\end{figure*}

\paragraph{Document elements} Elements in VRDs encompass both textual items (e.g., paragraphs, captions) and visual ones (tables, figures). When the entities in question are corrupted,
unanswerability becomes particularly difficult to grasp if the substitutes are placed in different document elements. 
To simulate element-wise corruptions, we 
replace an NLP entity in the original question with any one appearing 
in the different elements present in the document.
For example, to corrupt the entities in the question, we pick 2 out of 11 temporal entities belonging to \textit{Abandon} elements (i.e., headers, footers, footnotes, and marginal notes) from the infographic appearing in the left-hand side document page in Figure~\ref{fig:runningexample}.

\paragraph{Document layouts} Question entities can appear in multiple layout positions and pages.
While evaluating the unanswerability of a question with a corrupted entity on a given document page, 
the presence of similar entities within that page, eventually in different positions (i.e., the in-page corruption), 
can be challenging because the model may struggle to detect the error as layout information becomes diriment.
We simulate layout-related errors by replacing an entity in the original question with 
both in- and out-page entities.
For example, in Figure~\ref{fig:runningexample}, the question is corrupted with the green and red entities 
belonging to different pages.


\section{The evaluation benchmark}
\label{sec:methodology}

This section describes the \system\ framework, which generates unanswerable questions to test VQA models.
Its architecture is depicted in Figure~\ref{fig:framework}. 
Given a VQA dataset consisting of VRDs (see Section~\ref{sec:preliminaries}), it performs:

\begin{enumerate}
    \item \textit{Augumentation}, which extracts auxiliary information from VRDs, like OCR or visual element captions, necessary for the next steps;  
    \item \textit{Corruption}, which corrupts the questions as described in Section~\ref{sec:corruption}, both separately for each entity type and mixed;
    \item \textit{Verification}, verifies the actual questions' unanswerability by employing a VLLM-as-a-judge approach;
    \item \textit{Evaluation}, which tests the corrupted questions on several VQA models, with and without enriching the prompt with the auxiliary information and collecting the document- and page-level accuracies.
\end{enumerate}

Each pipeline step, thoroughly described in the following sections, is designed to be modular and highly customizable. 
Additional information, including prompts, on each of the following steps is reported in the Supplementary Material.

\subsection{Augmentation}
\label{sec:augment}

This step focuses on enriching the input VQA examples with the following auxiliary information:




\begin{enumerate}
    \item D\textit{ocument} L\textit{ayout} A\textit{nalysis} (DLA): it extracts document elements (including metadata) within each VRD page;
    \item E\textit{lement} C\textit{aptions} (EC): it generates textual captions of visual elements (e.g., plots, diagrams, figures) using the state-of-the-art Qwen 2.5 VL model~\cite{Qwen2VL}. The purpose is to allow the automatic extraction of entities from multimodal document elements;
    \item O\textit{ptical} C\textit{haracter} R\textit{ecognition} (OCR): it generates transcriptions of the textual elements present in the document image, such as paragraphs or titles, using the cost-effective GOT-OCR 2 model~\cite{wei2024general_GOTOCR}.
\end{enumerate}

Our structured pipeline enables a comprehensive textual representation of document pages by systematically processing visual elements, such as tables and figures. This approach enhances document comprehension compared to existing methods, which extract information from visual elements without preserving their structural relationships.

\subsection{Corruption}
\label{sec:corrupt}

This stage transforms each answerable question $Q$ in the original VQA dataset into an unanswerable one $\hat{Q}$ by applying the corruption types described in Section~\ref{sec:corruption}.

To detect NLP entities, we leverage the GliNER model~\cite{zaratiana-etal-2024-gliner} on a set of user-defined entities.
We define the following macro categories: Numerical, Temporal, Miscellaneous, Location, and Structural. 
To test VLLMs' resilience to varying perturbation levels, we generate corrupted questions where a single corruption is applied (C=1) or two/three corruptions (C=2/3) are present in the generated question.  
We decided not to exceed C=3 to avoid excessively severe corruptions (uncommon for humans). 

To ensure grammatical coherence and semantic plausibility while maintaining specific corrupted elements, we leverage Qwen 2.5~\cite{qwen2}. 
Our methodology provides the LLM with a structured framework comprising the original question for contextual reference, the corrupted version requiring refinement, a comprehensive list of corruption items that must be preserved, explicit directives emphasizing readability and linguistic naturalness, and curated examples of adequate and suboptimal refinements. 

\subsection{Verification}
\label{sec:verify } 

To prevent \system\ from collecting results from not truly unanswerable questions, we employ a VLLM-as-a-judge approach~\cite{li2024llms, zheng2023judging}, i.e., we inquire the established Gemini 2.5 Flash VLLM~\cite{team2023gemini} to double-check whether each new question is actually unanswerable on each page of the analyzed VRDs and is unlikely to contain hallucinations. 
To limit evaluation circularity, we deliberately use a model different from those tested during the main experiments. We selected Gemini 2.5 Flash due to its strong trade-off between performance on multimodal document understanding tasks and efficiency. 
The verification process utilizes a structured prompt that incorporates several critical components to ensure accurate assessment. In detail, the prompt includes a detailed task description, comprehensive OCR output from the document page, and explicit entity mapping that shows the relationship between original and corrupted entities.
Questions marked as unanswerable by Gemini ($\thicksim$30\%) are manually reviewed by human experts to evaluate the quality of the model’s judgment. 
On this step, we find an average precision of 96.97\%, indicating strong alignment between Gemini's predictions and human assessment. 
Notably, we observe that the discarded questions are predominantly associated with lower complexity levels, suggesting that simpler corruptions are more prone to accidental answerability.

\subsection{Evaluation}
\label{sec:testing}

Given the set of models under evaluation, we prompt each of them with the corrupted and verified questions and collect their outcomes. We test the pre-trained model versions under a zero-shot setting. 
To enrich the  benchmarking phase, beyond  the question complexity, we also consider for VLLMs the following parameters:
    (1) \textit{Page window size} ($w$), which indicates the number of consecutive pages that are processed (1 to 3), reflecting the multi-page nature of documents;
    (2) \textit{OCR-inclusion}, which indicates whether the text transcription is included 
    or not~\cite{wei2024general_GOTOCR};
    (3) \textit{Explicit}, which indicates whether the prompt 
    indicates the possibility of unanswerability of the given question or not.  

\section{Experiments}
\label{sec:results}

We run experiments on a machine equipped with NVIDIA A6000 GPUs, 192GB of RAM, and an AMD 7950X CPU.
The total computational budget was around 90 hours
to perform experiments with a single execution per dataset, model, and setting.
Due to the lack of space, the statistics about the original and sampled datasets, the
models settings and additional results are reported in the Supplementary Material.

\subsection{Original datasets}
\label{subsec:datasets}

We analyze two open-source benchmark datasets for multi-page VQA from VRDs: (1) \textit{MPDocVQA}~\cite{MPDocVQA}, which collects 5,131 documents of varying length along with 36,230 question-answer pairs in its train set; (2) \textit{DUDE}~\cite{LandeghemPTJBBC23}, which consists of 5,017 documents and 40,000+ questions. 
To limit the computational and human efforts,
hereafter we will consider a representative sample of 300 questions. 

\subsection{Augmented, corrupted, and verified data}
\label{subsec:corrupt}

For both datasets' samples we process 424 documents containing 595 questions and generate 2176 potentially unanswerable questions
as well as the necessary auxiliary information (more details in the Supplementary Material).
Then, we perform verification, identifying 593 genuinely unanswerable questions with variable complexity (318 level-1 questions, 201 level-2 questions, and 74 level-3 questions). 
In both datasets, we achieve a significant variety in the number of pages per document. 
The elements of types 
Abandon 
and Plain Text are predominant, whereas figures, tables, and titles are relatively rare but with non-negligible peaks. 
Similar to prior works~\cite{MPDocVQA, LandeghemPTJBBC23}, we neglect other elements, such as formulas, as they are statistically irrelevant.
\begin{table*}[th!]
\centering
\footnotesize
\setlength{\tabcolsep}{5.2pt}

\begin{tabular}{cccccccccccccc}
\toprule
& & \multicolumn{12}{c}{\textbf{DUDE}} \\ \midrule
& & Phi4 & Molmo & Ovis & Llama & \begin{tabular}[c]{@{}c@{}}Llava\\  34B\end{tabular}  & \begin{tabular}[c]{@{}c@{}}Gemma3\\  27B\end{tabular} & \begin{tabular}[c]{@{}c@{}}Qwen2.5\\ VL 7B\end{tabular} &\begin{tabular}[c]{@{}c@{}}Qwen2.5\\  VL 72B\end{tabular} & \begin{tabular}[c]{@{}c@{}}InterVL3\\  9B\end{tabular} & \begin{tabular}[c]{@{}c@{}}InterVL3\\  78B\end{tabular} & \begin{tabular}[c]{@{}c@{}}GPT4.1\\  mini\end{tabular} & O3\\ \midrule
\multicolumn{2}{c|}{$Acc_D$}        & 0.070 & 0.230 & 0.241 & 0.289 & 0.401 & \underline{0.503} & 0.460 & \textbf{0.599} & 0.267 & 0.326 & 0.214 & 0.239 \\ 
\multicolumn{2}{c|}{$Acc_P$}        & 0.248 & 0.554 & 0.674 & 0.680 & 0.717 & \underline{0.786} & \textbf{0.835} & 0.754 & 0.713 & 0.781 & 0.638 & 0.663\\ \midrule
\multirow{3}{*}{\rotatebox[origin=c]{0}{$Acc_D$}} 
& \multicolumn{1}{c|}{C1}           & 0.079 & 0.254 & 0.281 & 0.342 & 0.377 & \underline{0.482} & 0.465 & \textbf{0.588} & 0.281 & 0.342 & 0.202 & 0.227\\
& \multicolumn{1}{c|}{C2}           & 0.052 & 0.190 & 0.172 & 0.224 & 0.483 & \underline{0.586} & 0.517 & \textbf{0.707} & 0.259 & 0.328 & 0.276 & 0.301\\
& \multicolumn{1}{c|}{C3}           & 0.067 & 0.200 & 0.200 & 0.133 & 0.267 & \textbf{0.333} & 0.200 & \underline{0.267} & 0.200 & 0.200 & 0.067 & 0.092\\ \midrule
\multirow{3}{*}{\rotatebox[origin=c]{0}{$Acc_P$}}
& \multicolumn{1}{c|}{C1}           & 0.266 & 0.577 & 0.723 & 0.712 & 0.701 & \underline{0.810} & \textbf{0.843} & 0.753 & 0.738 & 0.805 & 0.636 & 0.661\\
& \multicolumn{1}{c|}{C2}           & 0.240 & 0.542 & 0.615 & 0.655 & 0.760 & 0.764 & \textbf{0.847} & \underline{0.816} & 0.684 & 0.760 & 0.669 & 0.694\\
& \multicolumn{1}{c|}{C3}           & 0.141 & 0.423 & 0.513 & 0.526 & \underline{0.692} & 0.679 & \textbf{0.731} & 0.519 & 0.628 & 0.667 & 0.538 & 0.563\\ \midrule
& & \multicolumn{12}{c}{\textbf{MPDocVQA}} \\ \midrule
\multicolumn{2}{c|}{$Acc_D$}        & 0.037 & 0.340 & 0.217 & 0.325 & 0.357 & 0.394 & \underline{0.490} & \textbf{0.581} & 0.241 & 0.219 & 0.264 & 0.163\\ 
\multicolumn{2}{c|}{$Acc_P$}        & 0.211 & 0.780 & 0.792 & 0.796 & 0.708 & 0.838 & \textbf{0.881} & \underline{0.842} & 0.782 & 0.818 & 0.775 & 0.738\\ \midrule
\multirow{3}{*}{\rotatebox[origin=c]{0}{$Acc_D$}} 
& \multicolumn{1}{c|}{C1}           & 0.044 & 0.358 & 0.221 & 0.314 & 0.309 & 0.402 & \underline{0.500} & \textbf{0.613} & 0.255 & 0.275 & 0.294 & 0.18\\
& \multicolumn{1}{c|}{C2}           & 0.028 & 0.329 & 0.189 & 0.322 & 0.441 & 0.420 & \underline{0.497} & \textbf{0.538} & 0.259 & 0.175 & 0.259 & 0.16\\
& \multicolumn{1}{c|}{C3}           & 0.034 & 0.305 & 0.271 & 0.373 & 0.322 & 0.305 & \underline{0.441} & \textbf{0.576} & 0.153 & 0.136 & 0.169 & 0.06\\ \midrule
\multirow{3}{*}{\rotatebox[origin=c]{0}{$Acc_P$}}
& \multicolumn{1}{c|}{C1}           & 0.225 & 0.830 & 0.815 & 0.845 & 0.707 & 0.852 & \textbf{0.901} & \underline{0.855} & 0.829 & 0.849 & 0.827 & 0.780 \\
& \multicolumn{1}{c|}{C2}           & 0.188 & 0.699 & 0.740 & 0.724 & 0.729 & \underline{0.824} & \textbf{0.850} & 0.791 & 0.725 & 0.757 & 0.691 & 0.669\\
& \multicolumn{1}{c|}{C3}           & 0.205 & 0.749 & 0.808 & 0.749 & 0.663 & 0.808 & \underline{0.865} & \textbf{0.885} & 0.712 & 0.824 & 0.741 & 0.712\\ 
\bottomrule
\end{tabular}
\caption{Document- and Page- Accuracy for each dataset and complexity. 
Best models are in bold, the second are underlined.
}
\label{tab:accdoc}
\end{table*}

\subsection{Models used in the framework}
\label{subsec:models}

Due to the nature of the datasets, which primarily consist of scanned documents, including handwritten documents, we consider the DocLayout-YOLO model a reference after an empirical evaluation of over 100 documents for each dataset. 
We leverage an additional phase to extract textual representations upon identifying document elements.
In particular, we employ the lightweight state-of-the-art GOT-OCR 2~\cite{wei2024general_GOTOCR} for OCR on textual elements, while Qwen 2.5 VL 7B~\cite{Qwen2VL} for visually rich items captioning. 
These operations are performed at the image patch level to provide reliable results.
We employ GliNER Large V2~\cite{zaratiana-etal-2024-gliner} for NER on both VRD elements and questions in order to perform the corruption.  
Additionally, to post-process corrupted questions, we leverage Qwen 2.5~\cite{qwen2}.
For the verification phase, we employ a state-of-the-art Gemini 2.5 Flash~\cite{team2023gemini}.

\subsection{Evaluated models}
We test a variety of VLLMs with different sizes, pretraining procedures, and optimize their parameter settings. In detail, we analyze Phi 4 Multimodal~\cite{abdin2024phi}, Qwen 2.5 VL 7B and 72B~\cite{Qwen2VL}, Molmo 7B~\cite{deitke2024molmo}, InternVL 3 9B and 78B~\cite{zhu2025internvl3}, 
Ovis1.6 9B~\cite{lu2024ovis},  LLama 3.2 11B~\cite{grattafiori2024llama}, Gemma3 27B~\cite{team2025gemma}, Llava1.6 34B~\cite{liu2024llavanext}, GPT-4.1-mini and O3. 

\subsection{Results discussion}

\begin{figure*}[th!]
\centering
\scriptsize
  \includegraphics[width=0.975\textwidth]{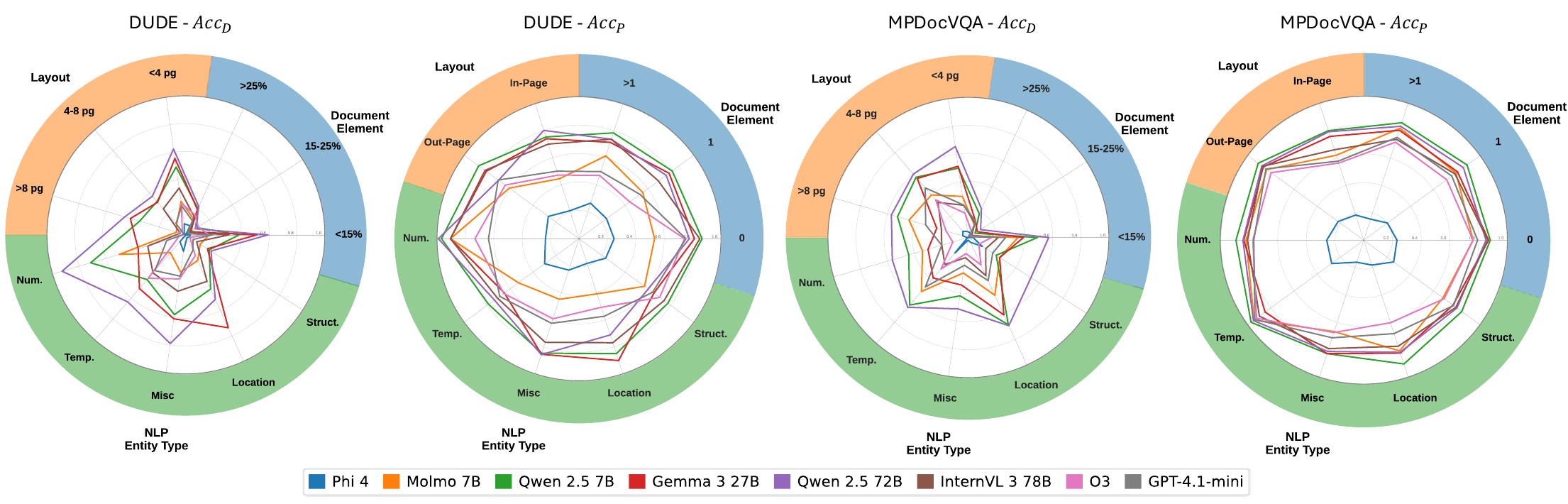}
\caption{Impact of corruption type on VQA performance across datasets and sizes of the models (representative subset).}
\label{fig:paramanalysis}
\end{figure*}

\paragraph{\textbf{RQ1:  Are VLLMs capable of accurately detecting question unanswerability due to the entity corruption?}} 
To answer RQ1, we analyze the $Acc_D$ and $Acc_P$ achieved by the tested models (see Table~\ref{tab:accdoc}). 
$Acc_D$ performance is consistently lower than $Acc_P$. This trend is particularly evident in long documents, where the likelihood of misclassifying an unanswerable question is higher (see Figure~\ref{fig:paramanalysis}).
Due to their highly specialized pretraining, Qwen and Gemma models demonstrate superior performance metrics.
Our analysis indicates that model size is not the most discriminant factor influencing the performance, 
suggesting that architectural features and training strategies are paramount 
and often yield more substantial gains than the VLLM scale.
Our findings show that a comprehensive understanding of the document is crucial to effectively address VQA on VRDs. 
This is confirmed by the results achieved with complexities 1 and 2, as all models roughly perform similarly. 
Converseley, at Complexity 3 the overall performance degrades because the corruption severely alters questions' meaning.

\begin{table*}[th!]
\centering
\footnotesize
\setlength{\tabcolsep}{4.4pt}

\begin{tabular}{cp{1mm}lcccccccccccc}
\toprule
& & & \multicolumn{12}{c}{\textbf{DUDE}} \\ \midrule
& 
\multicolumn{2}{c}{}  & Phi4 & Molmo & Ovis & Llama & \begin{tabular}[c]{@{}c@{}}Llava\\  34B\end{tabular}  & \begin{tabular}[c]{@{}c@{}}Gemma3\\  27B\end{tabular} & \begin{tabular}[c]{@{}c@{}}Qwen2.5\\ VL 7B\end{tabular} &\begin{tabular}[c]{@{}c@{}}Qwen2.5\\ VL 72B\end{tabular} & \begin{tabular}[c]{@{}c@{}}InterVL3\\  9B\end{tabular} & \begin{tabular}[c]{@{}c@{}}InterVL3\\  78B\end{tabular} & \begin{tabular}[c]{@{}c@{}}GPT4.1\\  mini\end{tabular} & O3\\ \midrule
\multirow{5}{*}{\rotatebox[origin=c]{90}{Doc Elem}} 
& \multicolumn{2}{c|}{Title}            & 0.143 & 0.357 & 0.643 & 0.500 & 0.571 & 0.714 & \underline{0.786} & \textbf{0.846} & 0.429 & \underline{0.786} & 0.357 & 0.407\\
& \multicolumn{2}{c|}{Text}             & 0.152 & 0.411 & 0.601 & 0.506 & 0.715 & \underline{0.766} & \underline{0.766} & \textbf{0.835} & 0.576 & 0.728 & 0.500 & 0.486\\
& \multicolumn{2}{c|}{Figure}           & 0.328 & 0.406 & 0.438 & 0.531 & 0.625 & \underline{0.781} & 0.750 & \textbf{0.831} & 0.656 & 0.625 & 0.344 & 0.409\\
& \multicolumn{2}{c|}{Table}            & 0.150 & 0.483 & 0.400 & 0.483 & 0.567 & 0.617 & 0.683 & \textbf{0.732} & 0.417 & 0.617 & 0.533 & \underline{0.686}\\
& \multicolumn{2}{c|}{Abandon}          & 0.233 & 0.567 & \textbf{0.767} & 0.667 & 0.533 & \underline{0.733} & \underline{0.733} & 0.675 & 0.700 & \underline{0.733} & 0.633 & 0.610\\ \midrule
\multirow{4}{*}{\rotatebox[origin=c]{90}{Layout}}
& \multicolumn{2}{c|}{Top Left}         & 0.072 & 0.387 & 0.361 & 0.428 & 0.680 & \underline{0.768} & 0.655 & \textbf{0.892} & 0.392 & 0.649 & 0.670 & 0.720\\
& \multicolumn{2}{c|}{Top Right}        & 0.131 & 0.377 & 0.492 & 0.492 & 0.705 & \underline{0.869} & 0.770 & \textbf{0.896} & 0.525 & 0.754 & 0.426 & 0.412\\
& \multicolumn{2}{c|}{Bottom Left}      & 0.131 & 0.381 & 0.452 & 0.548 & 0.685 & \textbf{0.750} & 0.667 & \underline{0.742} & 0.500 & 0.708 & 0.435 & 0.499\\
& \multicolumn{2}{c|}{Bottom Right}     & 0.321 & 0.543 & 0.514 & 0.500 & 0.693 & 0.664 & \underline{0.707} & \textbf{0.752} & 0.629 & 0.643 & 0.471 & 0.624\\ \toprule
& & & \multicolumn{12}{c}{\textbf{MPDocVQA}} \\ \midrule
\multirow{5}{*}{\rotatebox[origin=c]{90}{Doc Elem}} 
& \multicolumn{2}{c|}{Title}            & 0.056 & 0.444 & 0.389 & 0.389 & \underline{0.667} & 0.583 & 0.528 & \textbf{0.682} & 0.333 & 0.361 & 0.361 & 0.306\\
& \multicolumn{2}{c|}{Text}             & 0.208 & 0.625 & 0.709 & 0.646 & 0.721 & 0.800 & \underline{0.823} & \textbf{0.828} & 0.654 & 0.723 & 0.613 & 0.609\\
& \multicolumn{2}{c|}{Figure}           & 0.176 & 0.576 & 0.353 & 0.647 & \underline{0.694} & 0.624 & \textbf{0.800} & 0.658 & 0.600 & 0.612 & 0.447 & 0.388\\
& \multicolumn{2}{c|}{Table}            & 0.060 & 0.641 & 0.530 & 0.581 & 0.607 & 0.675 & \textbf{0.735} & \underline{0.719} & 0.350 & 0.325 & 0.487 & 0.462\\
& \multicolumn{2}{c|}{Abandon}          & 0.433 & 0.767 & 0.700 & 0.800 & 0.767 & 0.833 & 0.800 & \underline{0.833} & 0.667 & \textbf{0.867} & 0.733 & 0.600\\ \midrule
\multirow{4}{*}{\rotatebox[origin=c]{90}{Layout}}
& \multicolumn{2}{c|}{Top Left}         & 0.105 & 0.499 & 0.506 & 0.491 & \textbf{0.686} & 0.663 & \underline{0.676} & 0.670 & 0.410 & 0.388 & 0.469 & 0.416  \\
& \multicolumn{2}{c|}{Top Right}        & 0.297 & 0.669 & 0.699 & 0.720 & 0.665 & 0.803 & \textbf{0.845} & \underline{0.834} & 0.628 & 0.628 & 0.632 & 0.573  \\
& \multicolumn{2}{c|}{Bottom Left}      & 0.246 & 0.652 & 0.696 & 0.672 & 0.701 & 0.736 & \underline{0.812} & \textbf{0.849} & 0.661 & 0.690 & 0.559 & 0.614  \\
& \multicolumn{2}{c|}{Bottom Right}     & 0.231 & 0.790 & 0.706 & 0.803 & 0.714 & 0.824 & \textbf{0.899} & \underline{0.844} & 0.744 & 0.832 & 0.689 & 0.664  \\
\bottomrule
\end{tabular}
\caption{Effect of the corruption type on the Page-Level Accuracy. 
Best results are in bold, the second are underlined.
}
\label{tab:accinpage}
\end{table*}

\begin{figure}[th!]
\centering
\scriptsize
  \begin{tabular}{cccc}
   \subfloat[Effect of augmentation on $Acc_D$] 
   {\includegraphics[width=0.47\textwidth]{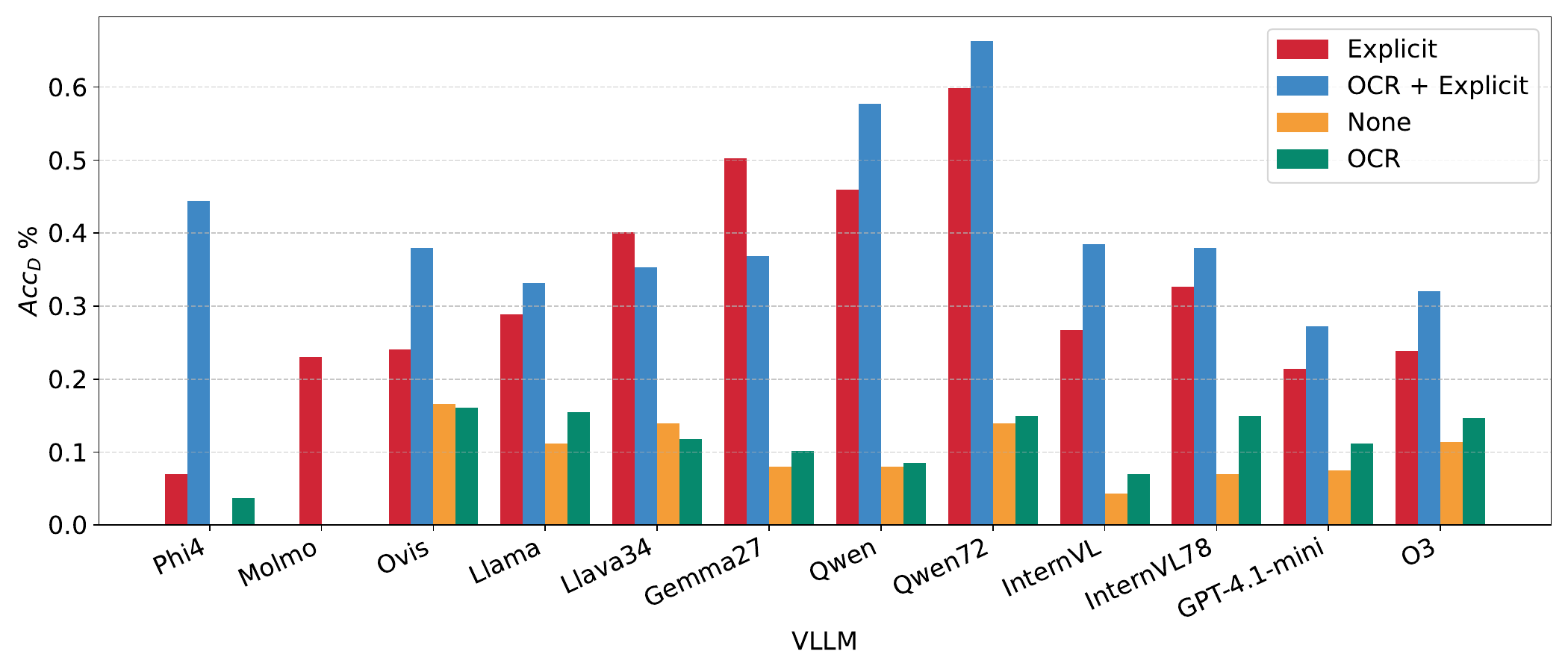}} \\
   \subfloat[Effect of window size on $Acc_D$] 
   {\includegraphics[width=0.47\textwidth]{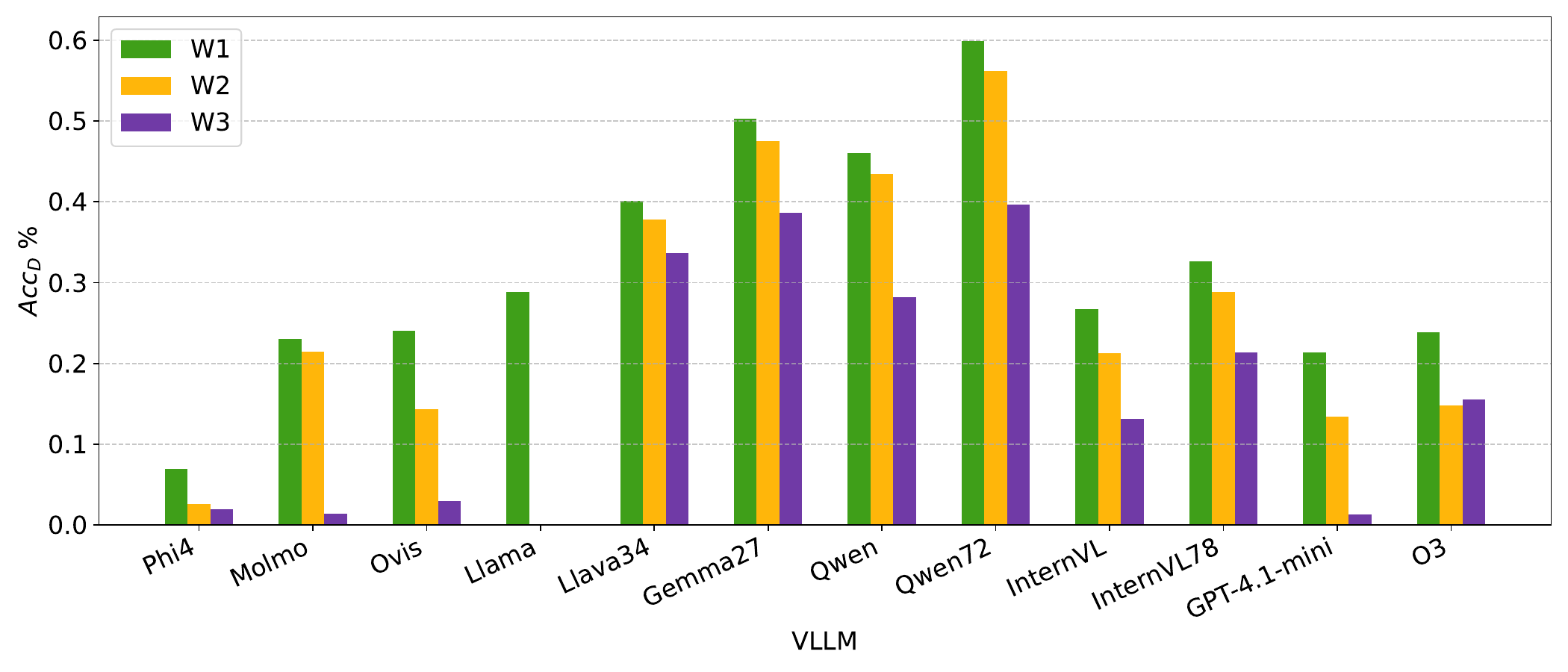}} 
  \end{tabular}
\caption{Effect of augmented information and window size on $Acc_D$ 
performance. DUDE dataset.}
\label{fig:ablationDUDE}
\end{figure}


\paragraph{\textbf{RQ2: What is the effect of different corruption types on
models’ performance?}} 
In Figure~\ref{fig:paramanalysis} we 
 analyze the effect of the corrupted entity type (i.e., numeric, temporal, miscellaneous, location, structure), the number/percentage of different visually reach document elements, and the document length on the document-level performance. Furthermore, we also compare the page-level performance across the NLP entity types. 
For the layout and document element analysis  
we group the outcomes by the page-level presence of document elements 
and by the presence of corrupted entities on a page for the layout. 

The results show the models' resilience to location and numerical entity corruptions.
Oppositely, their performance drop while dealing with structural entity modifications, particularly when structural layout-related information is manipulated (e.g., replacing the word "Figure" with "Table"). 
The composition of document elements also significantly affects models' performance. 
As the ratio of visual elements to the total number of elements increases, we observe a consistent decrease in
both accuracy metrics. 
As expected, all the tested models generally perform better on unanswerable questions related to text-only pages.
The document length emerges as a critical factor, with longer documents proving to be more challenging to process. 
Corruptions involving in-page entities turn out to be the most challenging. This difficulty stems from the proximity of misleading information, which creates false contexts that models struggle to discern.

Table~\ref{tab:accinpage} deepens the analysis of the $Acc_P$ performance for specific types of element- and layout-wise corruptions.
Our findings indicate that Abandon elements have a negligible impact on unanswerability detection capabilities, as models consistently demonstrate higher proficiency in distinguishing this secondary information from core content. 
Conversely, models exhibit poor performance on title elements on the MPDocVQA dataset, which directly correlates with the diminished accuracy in the top-left quadrant (50.63\%, with the remainder distributed between the top-right and bottom-left regions). 
The performance gap between Abandon and Title elements indicates that the specificity of the document type
significantly influences models' behavior.
Similarly, we detect a correlation between the presence of tabular elements 
and the fairly low performance of left-hand side page quarters, motivated
by the apperance of almost half of the tables in the top-left quadrant.

\hspace{-2cm}

\paragraph{\textbf{RQ3: Which in-context learning strategies are able to
mitigate the limitations of VLLMs in identifying unanswerable questions?}} 
We compare different in-context learning strategies for mitigating VLLM limitations in identifying unanswerable questions 
(see Figure~\ref{fig:ablationDUDE}). 
Regarding the prompt setting, explicitly stating the possibility of unanswerability significantly improves VLLMs' performance. 
As expected, models demonstrate significantly higher accuracy when explicitly instructed that questions may not be answerable based on the provided context. Additionally, including OCR-extracted text generally improves performance across all experimental settings, suggesting that textual information from images provides valuable context for determining question answerability. 
The combination of explicit unanswerability instructions and OCR integration yields the best overall performance, indicating a synergistic effect between semantic task understanding and comprehensive information access. 
We also analyze the effect of using different window sizes. Page-level accuracies consistently drop as window size increases. 
The tested models struggle to handle larger contexts as they may introduce noise or excessively spread the relevant information.  
Similar results on MPDocVQA are given in the Supplementary Material.

\section{Conclusions and future work}
\label{sec:conclusion}

The paper presents an evaluation framework for comprehensively analyzing the VLLM models' capabilities 
to detect unanswerable questions.
The framework can be used to benchmark the robustness of models in realistic scenarios, where questions are built by entity swaps.
To comprehensively explore the challenges of the VRDs, 
we leverage corruptions within different multimodal elements, layout positions, and pages.
The results provide insights into the model performance, 
highlighting gaps between models with different pretraining and size and their resilience to various corruption settings. 
The main limitations of this work are 
(1) the testing of zero-shot settings only; 
(2) the use of general-purpose in-context learning strategies. 
To address these issues, as future work, we plan to address VLLM fine-tuning and to design in-context learning mitigation strategies to overcome the most common weaknesses. 


\section{Acknowledgments}
This study was carried out within the FAIR (Future Artificial Intelligence Research) and received funding from Next-GenerationEU (Italian PNRR – M4 C2, Invest 1.3 – D.D. 1555.11-10-2022, PE00000013).
This manuscript reflects only the authors' views and opinions, neither the European Union nor the European Commission can be considered responsible for them.

\bibliography{biblio}

\newpage

\twocolumn[
  \begin{center}
    \LARGE\textbf{Supplementary Material}
    \vspace{1.5em} 
  \end{center}
]

\section{Additional datasets' statistics}
\label{sec:appendixdataset}

Table~\ref{tab:data} reports the statistics of the original dataset and the selected subset.
Table~\ref{tab:corrver} reports the statistics for both the corrupted and verified datasets.

Tables~\ref{tab:layout},~\ref{tab:entities},~\ref{tab:elements} report additional statistics about 
the NLP entities, document elements, and layout information relative to each dataset.

Analysis of document element distribution (see Table~\ref{tab:elements}) reveals a predominance of Abandon(headers, footers, footnotes, and marginal notes) and Text elements across both datasets, reflecting the underlying document types in the collections. 
Regarding entity (see Table~\ref{tab:entities}), on both datasets, the most predominant ones are Numeric, Miscellaneous, and Location.
The fine-grained entity distribution demonstrates both shared and distinct characteristics between the datasets. 
Familiar entities include measure units, person names, company names, spatial information, and document entity types.
MPDocVQA shows higher frequencies of percentage-related entities, product references, and chemical elements, while DUDE exhibits a notable emphasis on means-of-transport-related entities. 
About layout characteristics (see Table~\ref{tab:layout}), we observe an asymmetric distribution of entities, with a higher concentration in the left portions of documents. These distributional patterns persist consistently across both Corrupted and Verified versions of the datasets.

\begin{table}[th!]
\footnotesize
\centering

\setlength{\tabcolsep}{6pt}
\begin{tabular}{@{}cc|cc|cc@{}}
\toprule
& & \multicolumn{2}{c|}{\textbf{MPDocVQA}} & \multicolumn{2}{c}{\textbf{DUDE}} \\ \midrule
& & \multicolumn{1}{c}{\textbf{Full}} & \textbf{Sample} & \multicolumn{1}{c}{\textbf{Full}} & \textbf{Sample} \\ \midrule
\multicolumn{2}{c|}{\begin{tabular}[c]{@{}c@{}}N° documents\end{tabular}}
& \multicolumn{1}{c}{5131} & 147 & \multicolumn{1}{c}{5017} & 277 \\ \midrule
\begin{tabular}[c]{@{}c@{}}N° pages\end{tabular} 
& \begin{tabular}[c]{@{}c@{}}Avg\\Min\\Max\end{tabular} 
& \multicolumn{1}{c}{\begin{tabular}[c]{@{}c@{}}10.55\\1\\793\end{tabular}} 
& \begin{tabular}[c]{@{}c@{}}10.52\\1\\160\end{tabular} 
& \multicolumn{1}{c}{\begin{tabular}[c]{@{}c@{}}5.68\\1\\50\end{tabular}} 
& \begin{tabular}[c]{@{}c@{}}5.99\\1\\25\end{tabular} \\ \midrule
\multicolumn{2}{c|}{\begin{tabular}[c]{@{}c@{}}N° questions\end{tabular}}
& \multicolumn{1}{c}{36230} & 300 & \multicolumn{1}{c}{41453} & 300 \\ \midrule
\begin{tabular}[c]{@{}c@{}}N° questions / \\document\end{tabular} 
& \begin{tabular}[c]{@{}c@{}}Avg\\Min\\Max\end{tabular} 
& \multicolumn{1}{c}{\begin{tabular}[c]{@{}c@{}}7.06\\1\\606\end{tabular}} 
& \begin{tabular}[c]{@{}c@{}}2.03\\1\\11\end{tabular} 
& \multicolumn{1}{c}{\begin{tabular}[c]{@{}c@{}}8.26\\1\\38\end{tabular}} 
& \begin{tabular}[c]{@{}c@{}}1.07\\1\\3\end{tabular} \\ \bottomrule
\end{tabular}
\caption{Statistics about the original and sampled datasets} 
\label{tab:data}
\end{table}

\begin{table*}[h!]
\centering
\footnotesize
\setlength{\tabcolsep}{2.4pt}
\begin{tabular}{@{}cc|cccc|cccc|cccccccccccccc@{}}
\toprule
\textbf{Dataset} & \textbf{Version} & \multicolumn{4}{c|}{\begin{tabular}[c]{@{}c@{}}{\bf Number of}\\ {\bf questions}\end{tabular}} & \multicolumn{4}{c|}{\begin{tabular}[c]{@{}c@{}}{\bf Number of}\\ {\bf documents}\end{tabular}} & \multicolumn{12}{c}{\begin{tabular}[c]{@{}c@{}}{\bf Number of pages}\end{tabular}} \\ \midrule
& & \multirow{2}{*}{\textbf{Count}} & \multirow{2}{*}{\textbf{C1}} & \multirow{2}{*}{\textbf{C2}} & \multirow{2}{*}{\textbf{C3}} & \multirow{2}{*}{\textbf{Count}} & \multirow{2}{*}{\textbf{C1}} & \multirow{2}{*}{\textbf{C2}} & \multirow{2}{*}{\textbf{C3}} & \multicolumn{3}{c}{\textbf{Count}} & \multicolumn{3}{c}{\textbf{C1}} & \multicolumn{3}{c}{\textbf{C2}} & \multicolumn{3}{c}{\textbf{C3}} \\
& & & & & & & & & & \textbf{Avg} & \textbf{Min} & \textbf{Max} & \textbf{Avg} & \textbf{Min} & \textbf{Max} & \textbf{Avg} & \textbf{Min} & \textbf{Max} & \textbf{Avg} & \textbf{Min} & \textbf{Max} \\ \midrule
\multirow{2}{*}{MPDocVQA}   & Corrupted   & 1408 & 840 & 434 & 134 & 82 & 82 & 65 & 25 & 5.95 & 1 & \multicolumn{1}{c|}{40} & 6.00 & 1 & \multicolumn{1}{c|}{40} & 5.65 & 1 & \multicolumn{1}{c|}{21} & 6.22 & 1 & 21 \\
                            & Verified    & 406  & 204 & 143 & 59  & 69 & 50 & 49 & 17 & 6.93 & 1 & \multicolumn{1}{c|}{40} & 7.80 & 1 & \multicolumn{1}{c|}{40} & 5.83 & 1 & \multicolumn{1}{c|}{17} & 6.54 & 1 & 21 \\ \midrule
\multirow{2}{*}{DUDE}       & Corrupted   & 768  & 495 & 199 & 74  & 87 & 87 & 44 & 15 & 5.33 & 1 & \multicolumn{1}{c|}{20} & 5.45 & 1 & \multicolumn{1}{c|}{20} & 4.85 & 1 & \multicolumn{1}{c|}{17} & 5.89 & 1 & 10 \\
                            & Verified    & 187  & 114 & 58  & 15  & 54 & 46 & 26 & 11 & 5.04 & 1 & \multicolumn{1}{c|}{20} & 5.18 & 1 & \multicolumn{1}{c|}{20} & 4.74 & 1 & \multicolumn{1}{c|}{17} & 5.20 & 1 & 10 \\ \bottomrule
\end{tabular}
\caption{Statistics about the corrupted and verified datasets. C\textit{X} stands for Complexity=X.}
\label{tab:corrver}
\end{table*}

\begin{table}[th!]
\centering
\tiny
\setlength{\tabcolsep}{2.3pt}

\begin{tabular}{@{}ccccccccccccc@{}}
\toprule
                                          & \multicolumn{6}{c}{\textbf{MPDocVQA}}                                                                              & \multicolumn{6}{c}{\textbf{DUDE}}                                                           \\ \midrule
 \multicolumn{1}{c|}{}                     & \multicolumn{3}{c|}{\textbf{Corrupted}}                         & \multicolumn{3}{c|}{\textbf{Verified}}                    & \multicolumn{3}{c|}{\textbf{Corrupted}}                        & \multicolumn{3}{c}{\textbf{Verified}}     \\ \midrule
 \multicolumn{1}{c|}{}                     & Avg       & Min      & \multicolumn{1}{c|}{Max}        & Avg       & Min      & \multicolumn{1}{c|}{Max}  & Avg       & Min      & \multicolumn{1}{c|}{Max}       & Avg       & Min      & Max       \\ \midrule
 \multicolumn{1}{c|}{Top Left}     & 13.02    & 0    & \multicolumn{1}{c|}{89.00}     & 13.44    & 0    & \multicolumn{1}{c|}{70.00}  & 10.56    & 0    & \multicolumn{1}{c|}{49.00}    & 10.14    & 0    & 36.00    \\
 \multicolumn{1}{c|}{Top Right}    & 7.74     & 0 & \multicolumn{1}{c|}{105.00}    & 8.02     & 0    & \multicolumn{1}{c|}{105.00} & 7.79    & 0    & \multicolumn{1}{c|}{61.00}    & 5.72     & 0    & 35.00    \\
 \multicolumn{1}{c|}{Bottom Left}  & 10.90    & 0    & \multicolumn{1}{c|}{104.00}    & 11.05    & 0    & \multicolumn{1}{c|}{59.00}  & 10.08    & 0    & \multicolumn{1}{c|}{49.00}    & 9.74     & 0    & 38.00    \\
 \multicolumn{1}{c|}{Bottom Right} & 7.41     & 0    & \multicolumn{1}{c|}{98.00}     & 6.85     & 0    & \multicolumn{1}{c|}{79.00}  & 7.87     & 0    & \multicolumn{1}{c|}{54.00}    & 6.25     & 0    & 38.00    \\
\bottomrule
\end{tabular}
\caption{Detailed layout information about the analyzed datasets.}
\label{tab:layout}
\end{table}

\begin{table}[h!]
\centering
\tiny
\caption{NLP entity statistics over the datasets under analysis.}
\setlength{\tabcolsep}{2pt}
\begin{tabular}{cc|cccccc|cccccc}
\toprule
                                & & \multicolumn{6}{c|}{\textbf{MPDocVQA}}                                                              & \multicolumn{6}{c}{\textbf{DUDE}}                                                                   \\ \midrule
                                & & \multicolumn{3}{c|}{\textbf{Corrupted}}                         & \multicolumn{3}{c|}{\textbf{Verified}}     & \multicolumn{3}{c|}{\textbf{Corrupted}}                         & \multicolumn{3}{c}{\textbf{Verified}}      \\ 
                                & & Avg       & Min      & \multicolumn{1}{c|}{Max}        & Avg       & Min      & Max        & Avg       & Min      & \multicolumn{1}{c|}{Max}        & Avg       & Min      & Max        \\ \midrule
 \multirow{5}{*}{\rotatebox[origin=c]{90}{Macro Entities}} 
 &\textbf{Numeric}                 & 6.4     & 0   & \multicolumn{1}{c|}{117.7}   & 7.7     & 0    & 112.2    & 3.8     & 0    & \multicolumn{1}{c|}{83.7}    & 3.5     & 0    & 80.1     \\
& \textbf{Temporal}                & 3.8     & 0   & \multicolumn{1}{c|}{64.8}    & 4.3     & 0    & 64.6     & 3.3     & 0    & \multicolumn{1}{c|}{46.6}    & 3.5     & 0    & 44.3     \\
& \textbf{Misc}                    & 9.9     & 0   & \multicolumn{1}{c|}{175.1}   & 10.5    & 0    & 128.5    & 7.7     & 0    & \multicolumn{1}{c|}{154.0}   & 6.0     & 0    & 61.6     \\
& \textbf{Location}                & 6.4     & 0   & \multicolumn{1}{c|}{99.5}    & 6.7     & 0    & 72.0     & 8.8     & 0    & \multicolumn{1}{c|}{129.1}   & 7.5     & 0    & 65.0     \\
& \textbf{Structure}               & 5.2     & 0   & \multicolumn{1}{c|}{55.1}    & 5.5     & 0    & 47.5     & 6.5     & 0    & \multicolumn{1}{c|}{73.3}    & 5.2     & 0    & 25.8     \\ \midrule
\multirow{7}{*}{\rotatebox[origin=c]{90}{Numeric}} 
& number                            & 5.0     & 0    & \multicolumn{1}{c|}{137.0}   & 5.3     & 0    & 137.0    & 3.0     & 0    & \multicolumn{1}{c|}{57.0}    & 2.7     & 0    & 39.0     \\
& measure\_unit                     & 18.8    & 0    & \multicolumn{1}{c|}{170.0}   & 21.4    & 0    & 133.0    & 14.0    & 0    & \multicolumn{1}{c|}{351.0}   & 12.3    & 0    & 351.0    \\
& price                             & 0.9     & 0    & \multicolumn{1}{c|}{33.0}    & 1.2     & 0    & 33.0     & 0.6     & 0    & \multicolumn{1}{c|}{14.0}    & 0.5     & 0    & 10.0     \\
& percentage                        & 11.5    & 0    & \multicolumn{1}{c|}{245.0}   & 15.5    & 0    & 245.0    & 2.5     & 0    & \multicolumn{1}{c|}{47.0}    & 2.1     & 0    & 44.0     \\
& temperature                       & 0.9     & 0    & \multicolumn{1}{c|}{15.0}    & 0.9     & 0    & 14.0     & 1.2     & 0    & \multicolumn{1}{c|}{25.0}    & 1.0     & 0    & 25.0     \\
& currency                          & 7.5     & 0    & \multicolumn{1}{c|}{224.0}   & 9.8     & 0    & 224.0    & 5.5     & 0    & \multicolumn{1}{c|}{92.0}    & 5.6     & 0    & 92.0     \\\cmidrule{2-14}
\multirow{6}{*}{\rotatebox[origin=c]{90}{Temporal}} 
& date                              & 4.0     & 0    & \multicolumn{1}{c|}{38.0}    & 3.8     & 0    & 37.0     & 3.7     & 0   & \multicolumn{1}{c|}{33.0}    & 3.8     & 0    & 33.0     \\
& time\_info                        & 8.4     & 0    & \multicolumn{1}{c|}{104.0}   & 8.7     & 0    & 104.0    & 8.3     & 0   & \multicolumn{1}{c|}{105.0}   & 9.9     & 0    & 105.0    \\
& time\_value                       & 0.6     & 0    & \multicolumn{1}{c|}{13.0}    & 0.4     & 0    & 13.0     & 0.7     & 0   & \multicolumn{1}{c|}{15.0}    & 1.0     & 0    & 15.0     \\
& year\_info                        & 1.5     & 0    & \multicolumn{1}{c|}{47.0}    & 1.6     & 0    & 47.0     & 1.0     & 0   & \multicolumn{1}{c|}{21.0}    & 0.8     & 0    & 7.0      \\
& year\_value                       & 8.5     & 0    & \multicolumn{1}{c|}{187.0}   & 11.1    & 0    & 187.0    & 6.2     & 0   & \multicolumn{1}{c|}{106.0}   & 5.7     & 0    & 106.0    \\\cmidrule{2-14}
\multirow{13}{*}{\rotatebox[origin=c]{90}{Miscellaneous}} 
& person                            & 23.5    & 0   & \multicolumn{1}{c|}{648.0}    & 17.1    & 0    & 143.0    & 35.9    & 0    & \multicolumn{1}{c|}{697.0}   & 24.6    & 0    & 129.0    \\
& company                           & 24.7    & 0   & \multicolumn{1}{c|}{347.0}    & 26.6    & 0    & 347.0    & 14.1    & 0    & \multicolumn{1}{c|}{112.0}   & 11.7    & 0    & 63.0     \\
& event                             & 7.4     & 0   & \multicolumn{1}{c|}{187.0}    & 6.0     & 0    & 86.0     & 8.9    & 0    & \multicolumn{1}{c|}{71.0}    & 10.5    & 0    & 71.0     \\
& product                           & 13.9    & 0   & \multicolumn{1}{c|}{109.0}    & 17.1    & 0    & 109.0    & 6.7     & 0    & \multicolumn{1}{c|}{273.0}   & 3.0     & 0    & 42.0     \\
& food                              & 5.8     & 0   & \multicolumn{1}{c|}{154.0}    & 7.6     & 0    & 154.0    & 1.1    & 0    & \multicolumn{1}{c|}{33.0}    & 1.0     & 0    & 33.0     \\
& chemical\_elem                    & 37.3    & 0   & \multicolumn{1}{c|}{485.0}    & 43.3    & 0    & 485.0    & 6.5     & 0    & \multicolumn{1}{c|}{158.0}   & 5.2     & 0    & 56.0     \\
& job\_title\_name                  & 5.7     & 0   & \multicolumn{1}{c|}{104.0}    & 6.2     & 0    & 104.0    & 6.2    & 0    & \multicolumn{1}{c|}{61.0}    & 6.2     & 0    & 39.0     \\
& job\_title\_info                  & 0.1     & 0    & \multicolumn{1}{c|}{2.0}     & 0.1     & 0    & 2.0      & 0.2    & 0    & \multicolumn{1}{c|}{8.0}     & 0.3     & 0    & 8.0      \\
& animal                            & 1.0     & 0   & \multicolumn{1}{c|}{18.0}     & 1.1     & 0    & 18.0     & 2.1   & 0    & \multicolumn{1}{c|}{54.0}    & 2.5     & 0    & 54.0     \\
& plant                             & 6.3     & 0   & \multicolumn{1}{c|}{143.0}    & 7.8     & 0    & 143.0    & 3.7   & 0    & \multicolumn{1}{c|}{128.0}   & 2.6     & 0    & 79.0     \\
& movie                             & 0.1     & 0   & \multicolumn{1}{c|}{6.0}      & 0.2     & 0    & 6.0      & 0.3   & 0    & \multicolumn{1}{c|}{6.0}     & 0.4     & 0    & 6.0      \\
& book                              & 1.3     & 0   & \multicolumn{1}{c|}{25.0}     & 1.4     & 0    & 25.0     & 3.3   & 0    & \multicolumn{1}{c|}{190.0}   & 1.0     & 0    & 9.0      \\
& transport                         & 2.3     & 0   & \multicolumn{1}{c|}{49.0}     & 2.1     & 0    & 49.0     & 11.1  & 0    & \multicolumn{1}{c|}{212.0}   & 9.0     & 0    & 212.0    \\\cmidrule{2-14}
\multirow{7}{*}{\rotatebox[origin=c]{90}{Location}} 
& country                           & 7.9     & 0    & \multicolumn{1}{c|}{196.0}   & 6.3     & 0    & 78.0     & 5.5     & 0    & \multicolumn{1}{c|}{88.0}    & 5.2     & 0    & 88.0     \\
& city                              & 7.7     & 0    & \multicolumn{1}{c|}{137.0}   & 7.2     & 0    & 62.0     & 7.0     & 0    & \multicolumn{1}{c|}{68.0}    & 6.6     & 0    & 63.0     \\
& street                            & 0.8     & 0    & \multicolumn{1}{c|}{20.0}    & 0.8     & 0    & 20.0     & 2.7     & 0    & \multicolumn{1}{c|}{67.0}    & 2.2     & 0    & 67.0     \\
& spatial\_info                     & 22.1    & 0    & \multicolumn{1}{c|}{163.0}   & 24.3    & 0    & 163.0    & 43.2    & 0    & \multicolumn{1}{c|}{609.0}   & 35.8    & 0    & 201.0    \\
& continent                         & 4.4     & 0    & \multicolumn{1}{c|}{153.0}   & 5.4     & 0    & 153.0    & 2.0     & 0    & \multicolumn{1}{c|}{30.0}    & 1.9     & 0    & 27.0     \\
& postal\_code\_info                & 2.1     & 0    & \multicolumn{1}{c|}{26.0}    & 2.5     & 0    & 26.0     & 1.4     & 0    & \multicolumn{1}{c|}{41.0}    & 0.7     & 0    & 9.0      \\
& postal\_code\_val                 & 0.0     & 0    & \multicolumn{1}{c|}{2.0}     & 0.0     & 0    & 2.0      & 0.0     & 0    & \multicolumn{1}{c|}{1.0}     & 0.0     & 0    & 0.0      \\\cmidrule{2-14}
\multirow{6}{*}{\rotatebox[origin=c]{90}{Structure}} 
& doc\_pos\_info                & 4.6    & 0    & \multicolumn{1}{c|}{64.0}    & 4.4     & 0    & 34.0     & 4.6     & 0    & \multicolumn{1}{c|}{50.0}    & 3.2     & 0    & 28.0     \\
& page\_num\_info               & 0.7    & 0    & \multicolumn{1}{c|}{21.0}    & 0.6     & 0    & 6.0      & 3.2     & 0    & \multicolumn{1}{c|}{131.0}   & 1.3     & 0    & 17.0     \\
& page\_num                     & 0.0    & 0    & \multicolumn{1}{c|}{1.0}     & 0.0     & 0    & 0.0     & 0.0     & 0    & \multicolumn{1}{c|}{6.0}     & 0.0     & 0    & 0.0      \\
& doc\_elem\_type               & 25.7   & 0    & \multicolumn{1}{c|}{239.0}   & 27.5    & 0    & 239.0    & 30.9    & 0    & \multicolumn{1}{c|}{244.0}   & 26.5    & 0    & 107.0    \\
& doc\_elem\_info               & 0.3    & 0    & \multicolumn{1}{c|}{4.0}     & 0.3     & 0    & 4.0      & 0.4     & 0    & \multicolumn{1}{c|}{9.0}     & 0.2     & 0    & 3.0      \\
& doc\_struct\_info             & 0.0    & 0    & \multicolumn{1}{c|}{2.0}     & 0.0     & 0    & 2.0      & 0.0     & 0    & \multicolumn{1}{c|}{0.0}     & 0.0     & 0    & 0.0      \\
\bottomrule
\end{tabular}
\label{tab:entities}
\end{table}

\begin{table*}[h!]
\centering
\scriptsize

\begin{tabular}{@{}c|ccc|ccc|ccc|ccc|ccc|ccc@{}}
\toprule
& \multicolumn{9}{c|}{\textbf{MPDocVQA}} & \multicolumn{9}{c}{\textbf{DUDE}} \\ \midrule
& \multicolumn{3}{c|}{\textbf{Augmented}} & \multicolumn{3}{c|}{\textbf{Corrupted}} & \multicolumn{3}{c|}{\textbf{Verified}} & \multicolumn{3}{c|}{\textbf{Augmented}} & \multicolumn{3}{c|}{\textbf{Corrupted}} & \multicolumn{3}{c}{\textbf{Verified}} \\ \midrule
& Avg & Min & Max & Avg & Min & Max & Avg & Min & Max & Avg & Min & Max & Avg & Min & Max & Avg & Min & Max \\ \midrule
abandon & 16.59 & 0 & 218 & 17.74 & 0 & 218 & 19.52 & 0 & 218 & 10.50 & 0 & 75 & 8.22 & 0 & 36 & 7.09 & 0 & 36 \\
figure & 2.12 & 0 & 16 & 1.91 & 0 & 16 & 2.13 & 0 & 16 & 4.03 & 0 & 121 & 2.92 & 0 & 51 & 2.30 & 0 & 15 \\
isolate\_formula & 0.10 & 0 & 3 & 0.12 & 0 & 3 & 0.09 & 0 & 3 & 0.17 & 0 & 6 & 0.10 & 0 & 4 & 0.13 & 0 & 4 \\
plain text & 27.50 & 0 & 312 & 29.29 & 0 & 312 & 28.49 & 0 & 213 & 31.18 & 0 & 285 & 29.52 & 0 & 192 & 25.59 & 0 & 121 \\
table & 1.52 & 0 & 38 & 1.81 & 0 & 38 & 2.06 & 0 & 38 & 1.37 & 0 & 19 & 1.47 & 0 & 19 & 1.26 & 0 & 13 \\
title & 5.89 & 0 & 64 & 6.68 & 0 & 64 & 7.23 & 0 & 64 & 8.19 & 0 & 97 & 6.14 & 0 & 32 & 5.54 & 0 & 25 \\ \bottomrule
\end{tabular}
\caption{Document elements' statistics.}
\label{tab:elements}
\end{table*}

\section{List of  NLP entities}
\label{sec:entities}

We analyze the effect of corrupting different NLP entities. To this  end, 
we perform an extensive analysis of the sample datasets to identify prevalent topics and entity categories. 
Based on this analysis, we define a taxonomy of entities consisting of five categories:

\begin{itemize}
\item Numerical Corruption: "percentage", "currency", "temperature",  "measure\_unit", "numerical\_value\_number", "price\_number\_information", "price\_numerical\_value".

\item Temporal Corruption: "date\_information", "date\\\_numerical\_value", "time\_information", "time\\\_numerical\_value", "year\_number\_information", "year\\\_numerical\_value"

\item Entity Corruption: "person\_name", "company\\\_name", "product", "food", "chemical\_element", "job\_title\_name", "job\_title\_information", "animal", "plant", "movie", "book", "transport\_means", "event"

\item Location Corruption: "country", "city", "street", \\"spatial\_information", "continent", "postal\\\_code\_information", "postal\_code\_numerical\_value"

\item Document Structure Corruption: "document\_position\\\_information", "page\_number\_information", "page\\\_number\_numerical\_value", "document\_element\_type", "document\_element\_information", "document\_structure\\\_information"
\end{itemize}

The implementation of the entity extraction phase based on GliNER (large v2) requires careful calibration of detection thresholds for specific entity types to optimize extraction quality. We establish entity-specific confidence thresholds with a default threshold of 0.75 for general entities. 
Document structure elements require a higher threshold (0.8) for "document\_element\_type", "document\_element\_\\information", and "document\_structure\_information". 
Similarly, for "postal\_code\_information" we set the threshold to 0.8, while for "postal\_code\_numerical\_value" we set it to 0.78. 
For temporal entities, "date\_information" we set the threshold to  0.75, while "year\_numerical\_value" we set it  oto 0.7. Job-related entities required particularly stringent thresholds, i.e., for "job\_title\_name" 0.9, for "job\_title\_information" the threshold is 0.8, reflecting the complexity of accurately identifying these elements.

Given the absence of a comprehensive ground truth dataset for entity extraction in this context, we carry out a manual evaluation and iterative refinement of both entity definitions and their associated detection thresholds. This process ensured high-quality entity extraction while maintaining the contextual relevance necessary for effective question corruption.

\section{Experimental setup}
\label{sec:details}

In our experiment, we ensure maximal reproducibility and consistent evaluations across all models. For VLLMs, we standardize the token generation length to 1024 tokens to allow possible complete answers, while maintaining default settings for other parameters. The Qwen model implementation incorporated dynamic image scaling between 256 and 1440 pixels to optimize processing efficiency while preserving image quality. Llama 3.2 and Llava 1.6 are leveraged through the Ollama framework.
To ensure comprehensive evaluation, each model is tested across all possible combinations of prompt configurations and window sizes. 
Concerning VLM, they are tested on the default setting, with a binary prompt and page-by-page. The binary prompting is forced to get that some corrupted questions are unanswerable, otherwise not possible.

\textbf{Document Layout Analysis.} Our document analysis pipeline employs DocLayout-YOLO for layout detection, configured with a deliberately low confidence threshold of 0.1 to maximize object detection coverage. This configuration ensures comprehensive capture of document elements, though it frequently results in overlapping detection boxes. To address this overlap, we implemented a refinement process that compares pairs of overlapping elements. When the intersection-over-union ratio exceeds 0.6, we retain the larger bounding box, ensuring optimal coverage while eliminating redundant detections.

\noindent\textbf{OCR} The text extraction process utilizes two specialized models based on content type. For standard textual elements, we employ GOT-OCR 2 with its OCR-specific configuration to ensure accurate text recognition. Visual elements, specifically figures and tables, undergo analysis using Qwen 2.5 VL 7B, configured with a 1024-token generation limit to produce detailed descriptive content. This dual-model approach ensures appropriate processing for both textual and visual document components while maintaining high-quality information extraction throughout the pipeline.

\section{Prompt engineering}
\label{sec:prompt}

\textbf{Corruption} The corruption process occasionally produces syntactically or semantically challenged questions that require refinement to ensure human readability while maintaining their unanswerable nature. To address this challenge, we leverage the Qwen 2.5 7B language model. The model receives a carefully structured prompt that includes original and corrupted questions and explicit preservation instructions for corrupted elements.
Our prompt engineering approach provides the model with several key components to ensure optimal refinement: (1) the original question for context, (2) the corrupted version requiring refinement, (3) a comprehensive list of corrupted elements that must remain unchanged, (4) specific refinement directives focusing on readability and natural language flow, and (5) carefully selected exemplars demonstrating both successful and unsuccessful refinements. This structured approach ensures that the refined questions maintain their intended unanswerable characteristics while achieving natural linguistic quality suitable for human evaluation.
\vspace{2mm}
\begin{lstlisting}
PROMPT:
You are given two questions. The first one is the original one, the second one is the corrupted one.
The corruption is done based on entities extracted from the original question.

Original question: "{original_question}"
Corrupted question: "{corrupted_question}"

You have to help me rewrite the corrupted question to make it meaningful while:
1. Making it coherent and natural, while strictly keeping the exact same meaning
2. Ensuring it makes sense in the context of the original question
3. The following corrupted entities must be preserved in the rewritten question: {list(all_corrupted_entities)}
4. Editing the question minimally - only what's needed to make it coherent
5. Guaranteeing that the final output is meaningful

Original: "What is the highest temperature recorded?"
Bad corruption: "What is the 85 F temperature recorded?"
Correct rewrite: "Was 85 F the highest temperature recorded?"

Good Examples:
Original: "Which year is mentioned first in the x axis?"
Bad corruption: "Which 1975 is mentioned first in the x axis?"
Good rewrite: "Is 1975 the first year mentioned in the x axis?"

Original: "Which company had the most sales in 2022?"
Bad corruption: "Which Microsoft had the most sales in 2022?"
Correct rewrite: "Did Microsoft have the most sales in 2022?"

Important: Return only the rewritten question without any explanation or introductions.
\end{lstlisting}

\noindent\textbf{Verification} Our verification pipeline employs Gemini 2.5 Flash as an automated judge to evaluate the validity of corrupted questions. The verification process utilizes a structured prompt that incorporates several critical components to ensure accurate assessment. The prompt includes a detailed task description, comprehensive OCR output from the document page, and explicit entity mapping that shows the relationship between original and corrupted entities.
To maintain spatial coherence during verification, we reconstruct the document's OCR content following the natural reading order, organizing text elements from top to bottom and left to right. This reconstruction approach is consistently applied across both the verification stage and subsequent VQA model evaluation, ensuring uniform document representation throughout the pipeline. 
The verification prompt specifies a standardized output format, facilitating automated processing of verification results while maintaining consistency across the evaluation pipeline. This structured approach ensures reliable identification by looking at "verification\_result" field, set to false if the corrupted question is unanswerable.
\vspace{2mm}
\begin{lstlisting}
PROMPT:
You are an expert in Visual Question Answering on Document images. 
We are working on a project to verify the answerability of questions based on the information provided in a given image. 
In detail we have taken questions from a multipage VQA dataset and we have corrupted the questions based on the entities found in the whole document associated to the question. 
Now, given the corrupted question and each image of the document, we want to verify if the question is answerable based solely on the information provided in the given image. 
Your task is to help us to determine if the following corrupted question is answerable based solely on the information provided in the given image. 
The question answer must be explicitly stated in the image. 
In order to have a better document understanding, we extracted the following OCR text from the document:\n{ocr_text}

In addition here we provide the original entities found in the question and the corrupted ones in order to allow you to place special focus on the corrupted ones. The entities are reported with the format: ORIGINAL --$>$ CORRUPTED:\n{entities_string}

Respond with a structured response in JSON format with the following fields:
{
    "verification_result": "true if the question is answerable based solely on the information provided in the given image, or 'false' if it's not answerable",
    "question_answer": "The answer to the question or only the words 'not found' if the answer is not explicitly stated in the image"
}
Return only the JSON response. Without any other text or explanation.
Question: {question}
\end{lstlisting}
Questions marked as unanswerable were manually validated by three NLP experts (MSc or higher), achieving 96.97\% precision

\noindent\textbf{VLLM} For Vision-Language Large Language Models (VLLMs), we implemented a comprehensive evaluation framework that systematically tests different prompt configurations within defined context windows.
Our experimental design explores the impact of two key factors: explicit notification of potential question unanswerability and the inclusion of document OCR text.
The base prompt template establishes a clear task context and role definition for the model while maintaining flexibility for our experimental conditions:
\vspace{2mm}
\begin{lstlisting}
PROMPT:
You are an AI assistant specialized in analyzing document images and text.
Your task is to answer questions about the document image content precisely.

For this question, you have the following OCR text: {ocr_text} #OPTIONAL

Guidelines:
- Provide concise, focused answers (single word or short phrase preferred)
- Base your answer on both the image and the provided OCR text
- If uncertain, return 'Unable to determine' # OPTIONAL
- If you can't find the answer, return 'Unable to determine' # OPTIONAL
Question: {question}
\end{lstlisting}

This template incorporates several key elements: task specification, role definition, optional OCR context, and structured response guidelines. The optional components allow for a systematic evaluation of how different context levels affect model performance. 
To ensure optimal performance while maintaining comparability, we adapted the base prompt structure according to each model's author-recommended prompting patterns, while preserving the core evaluation framework.


\noindent\textbf{Ouput Standardization} To process metrics, we need a standard output. Although properly prompted, VLLMs may not follow output format directives. To overcome this issue, we leverage an LLM-as-a-judge that standardizes outputs that are not properly formatted.
This is done by exploiting Gemini 2.0 Flash with the following prompt:
\vspace{2mm}
\begin{lstlisting}
PROMPT:
I'm performing an evaluation test on the ability of different models to answer VQA questions from document images.
The model could return different answers to determine if the answer is 'unable to determine' or not. 
Your task is to detect if the answer means that the model is unable to determine the answer or not. 
Examples of answers that mean that the model is unable to determine the answer: 
- Not available. 
- Not provided in document. 
- The image does not provide information to answer the question. 
- I cannot provide an answer based on the given text. 
- The document does not provide information 
If the answer means 'unable to determine', respond with 'unable to determine', otherwise return the original answer. 
The answer is: {answer} 
Please respond only with the original answer or 'unable to determine' only.
\end{lstlisting}

\section{Additional  results}

\paragraph{RQ2 - Document and Page-Level Accuracy}
Table~\ref{tab:accdoc} and  ~\ref{tab:accpage} provide fine details about performances on analyzed metrics, respectively at document and page level. In detail, they extend the radar plots reported in the main paper by adding VLM performance. As expected, they perform poorly due to their nature and task settings.

\paragraph{RQ2/RQ3 - Document-Level Ablation}

In Table~\ref{tab:ablationDUDEDoc} and ~\ref{tab:ablationMPDoc}, we report the ablation study on the different models for different prompts and complexity levels. To reduce the cumbersome quantity of data and focus on relevant results, we decide to place focus on the two prompt types where the unanswertability is made explicit since providing the most relevant results (see Research Question 3 in the main paper).

The reported results demonstrate a clear performance advantage for Qwen when augmented with OCR explicit information, consistently achieving superior document-level accuracy across varying complexity conditions. 
This suggests that the integration of explicit text recognition significantly enhances document comprehension capabilities beyond what can be achieved through visual processing alone.
Performance degradation is evident as document complexity increases from C1 to C3, though this effect varies across models.
The substantial gap between OCR-enhanced and standard approaches underscores the importance of text recognition in document understanding tasks. Models exhibit heterogeneous performance patterns based on document characteristics, with notable sensitivities to document length, where accuracy typically diminishes as page count increases beyond 8 pages.
Entity-based analysis reveals differential performance across semantic categories. Location entities are generally processed more effectively, while Structure entities present consistent challenges across most models. 
This pattern manifests similarly in both datasets, suggesting fundamental limitations in how current vision-language models process structural document information. 
Interestingly, documents with lower element density ($<$15\%) yield better performance, indicating that visual clutter adversely affects comprehension capabilities.
The comparative analysis between DUDE and MPDocVQA demonstrates that while general performance trends remain consistent, the latter dataset shows less pronounced degradation across complexity levels for certain models, suggesting dataset-specific characteristics influence model robustness. 

\paragraph{RQ2/RQ3 - Page-Level Ablation}
The ablation studies on page-level accuracy across DUDE and MPDocVQA datasets (Table~\ref{tab:ablationDUDEPage}, ~\ref{tab:ablationMPPage}) demonstrate consistent superiority of Qwen with OCR explicit integration, highlighting the transformative impact of combining visual processing with textual recognition. This performance advantage persists across varying complexity levels, though it becomes less pronounced at C3, where models like Llava and Gemma sometimes outperform Qwen, suggesting these models possess enhanced resilience to extreme complexity.
The integration of OCR capabilities produces asymmetric benefits across document characteristics. For instance, while providing substantial improvements for most models on text-heavy elements, its impact on figures and tables is less consistent. This pattern indicates fundamental differences in how models process textual versus visual information in documents, with OCR integration primarily enhancing text extraction capabilities rather than comprehensive visual understanding.
Document element density emerges as a significant performance determinant, with most models achieving superior results on documents with lower element density ($<$15\%). This finding suggests that visual clutter presents a substantial challenge for current vision-language models. The spatial positioning of information also significantly impacts performance, with bottom-right positions generally yielding better results, potentially due to reading pattern biases in model training data.
Entity type analysis reveals pronounced performance differentials, with Numeric and Temporal entities being processed effectively while Structure entities remain challenging. This disparity indicates that current architectures excel at extracting discrete information but struggle with understanding document organization and hierarchical relationships. Notably, the MPDocVQA dataset shows less pronounced performance degradation across complexity levels compared to DUDE, suggesting dataset-specific characteristics influence model robustness.
In-page analysis further demonstrates that document understanding is highly context-dependent, with models exhibiting different strengths based on element type and position. 

\paragraph{RQ2/RQ3 - In-Page Ablation}

\begin{figure*}[h!]
\centering
\scriptsize
  \begin{tabular}{cc}
    \subfloat[MPDocVQA - Document Level Accuracy - Ablation parameters] 
    {\includegraphics[width=0.47\textwidth]{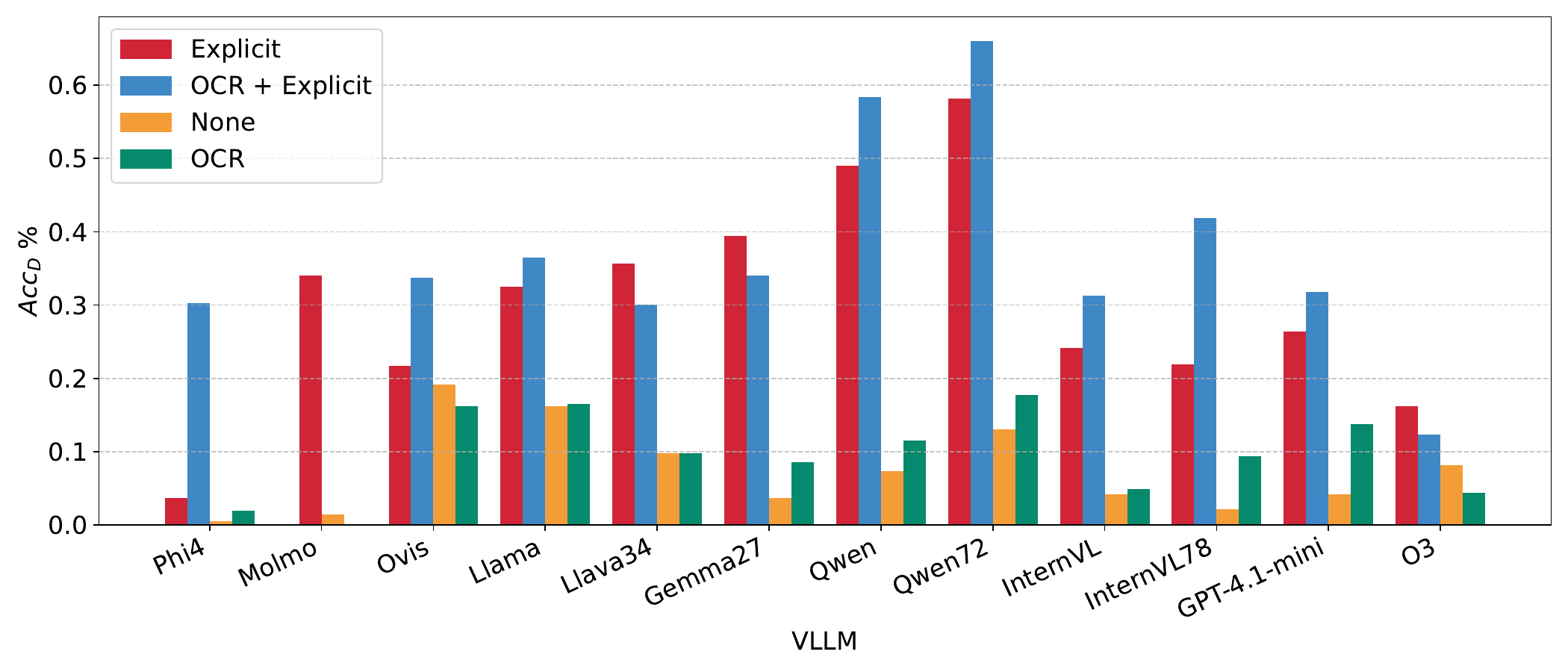}} &
    \subfloat[MPDocVQA - Document Level Accuracy - Ablation windows] 
    {\includegraphics[width=0.47\textwidth]{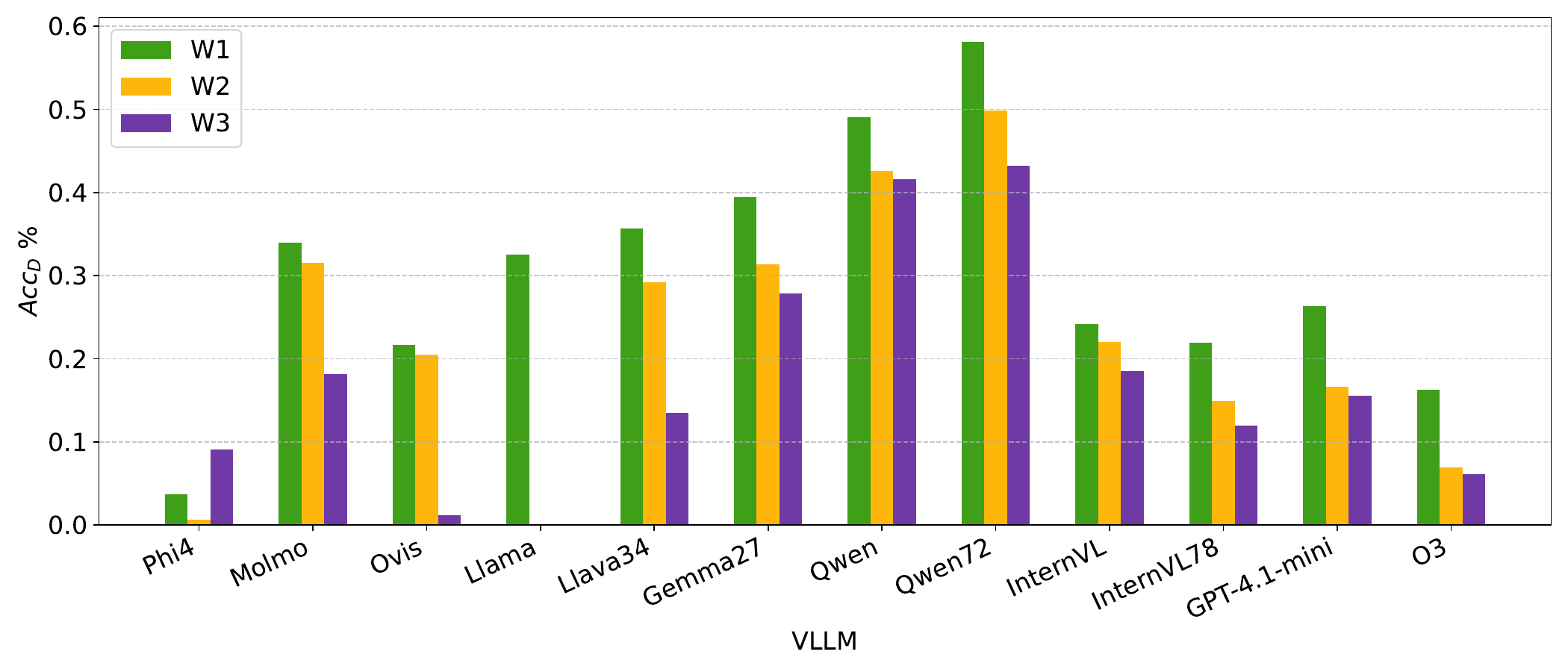}} \\
    \subfloat[MPDocVQA - Page Level Accuracy - Ablation parameters] 
    {\includegraphics[width=0.47\textwidth]{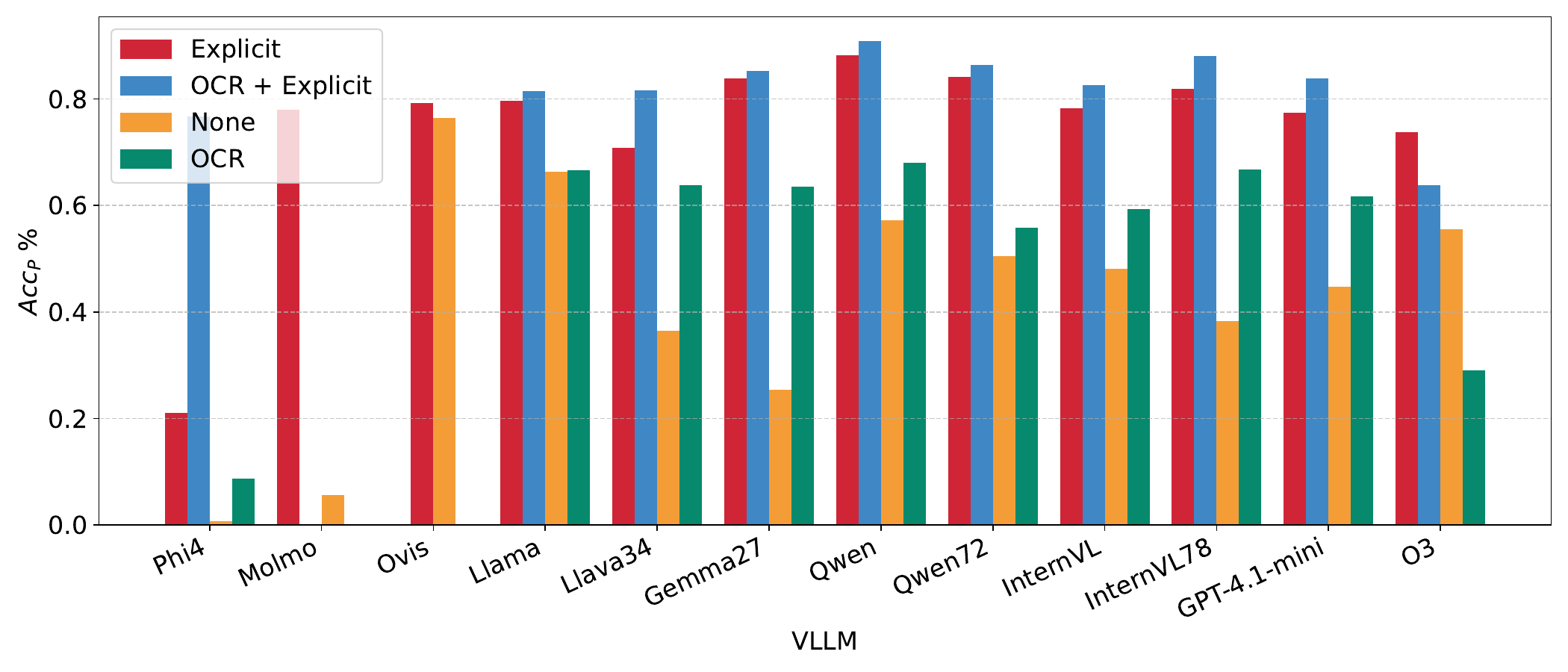}} &
    \subfloat[MPDocVQA - Page Level Accuracy - Ablation windows] 
    {\includegraphics[width=0.47\textwidth]{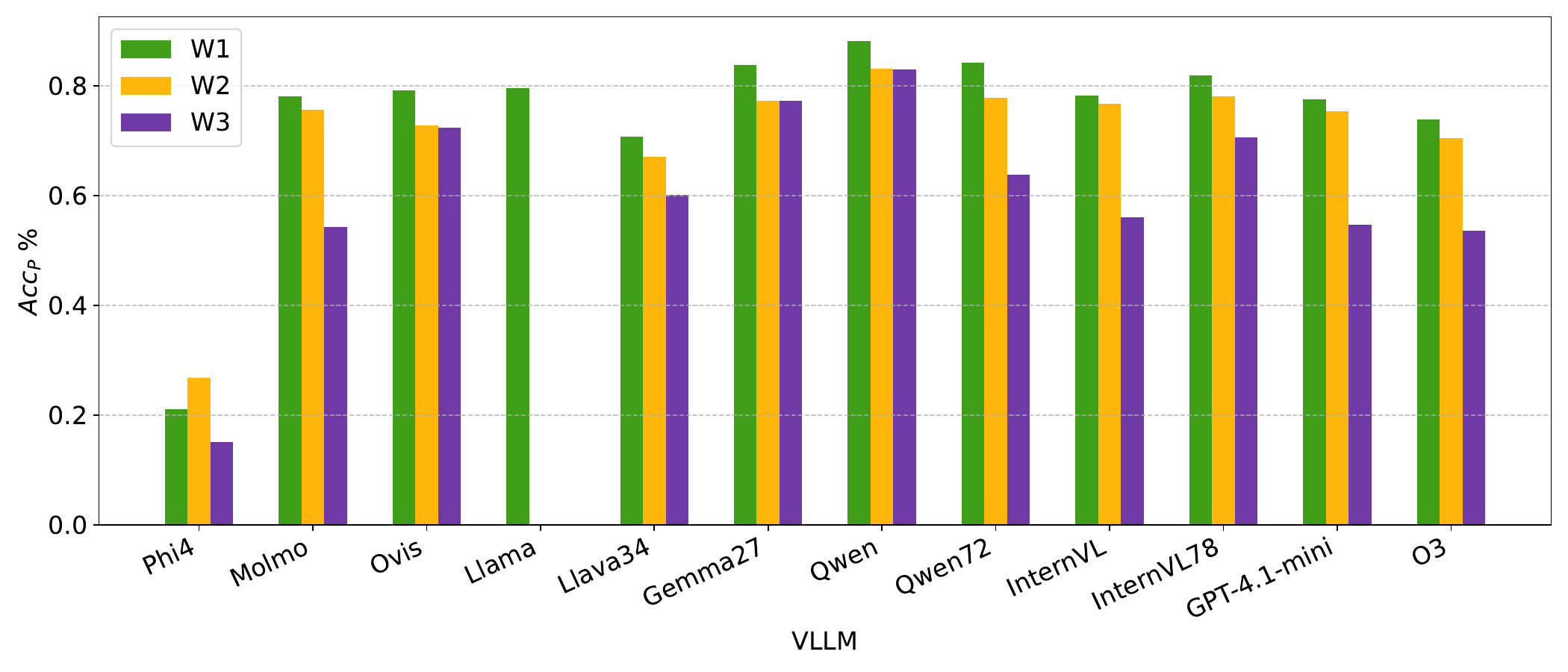}} \\
  \end{tabular}
\caption{Ablation study on in-context learning strategy and window size. MPDocVQA dataset (addressing RQ3)}
\label{fig:ablationMP}
\end{figure*}
The in-page analyses on Table~\ref{tab:AblDUDEInpage}, ~\ref{tab:AblMPInpage} reveal that document understanding is highly element-dependent and spatially nuanced, with consistent patterns emerging across both datasets despite their distinct characteristics.
Element-type analysis demonstrates that contemporary models exhibit specialized processing capabilities for different document components. Title elements generally yield the highest accuracy in DUDE, likely due to their distinctive visual formatting and semantic importance, while tables present persistent challenges that suggest limitations in structural reasoning. Interestingly, MPDocVQA shows strong table recognition capabilities for several models, indicating dataset-specific training or representation factors influence element processing capabilities.
Spatial positioning emerges as a critical factor in document understanding, with elements positioned in the bottom-right quadrant consistently achieving higher accuracy across models and complexity levels. 
This phenomenon reflects the same correlation between document elements and layout observed in the main paper.
OCR integration provides substantial but non-uniform benefits across elements and positions. Text-heavy elements show the most consistent improvements with OCR, while the benefits for figures are less pronounced. This differential impact highlights the complementary nature of visual and textual processing in document understanding tasks. The integration appears more consistently beneficial in MPDocVQA compared to DUDE, suggesting dataset characteristics influence the utility of explicit text recognition.
Complexity resilience varies significantly across element types and spatial positions. While performance generally degrades from C1 to C3, certain elements and positions maintain robust accuracy even at higher complexity levels. MPDocVQA demonstrates superior complexity resilience compared to DUDE, particularly for abandoned elements and bottom-positioned content. This difference suggests that dataset design characteristics substantially impact model robustness to document complexity.

These findings collectively underscore the multifaceted nature of document understanding, revealing that current vision-language models process documents through a complex interplay of element recognition, spatial reasoning, and textual integration. Future architectural improvements should focus on enhancing structural understanding capabilities and mitigating spatial biases to advance fine-grained document comprehension performance.

\paragraph{RQ3 - Ablation study on MPDocVQA}
Our study on in-context learning for vision-language models reveals striking patterns in unanswerable question detection (Figure~\ref{fig:ablationMP}). Explicitly stating that questions may be unanswerable dramatically improves model performance. Including OCR-extracted text substantially boosts accuracy across all conditions, providing critical context for answerability determination.
Combining explicit unanswerability instructions with OCR integration produces the strongest results, revealing powerful synergy between task understanding and information access. Counterintuitively, page-level accuracy plummets as window size increases—suggesting current models struggle when larger contexts dilute essential information.


\begin{table*}[h!]
\centering
\footnotesize
\setlength{\tabcolsep}{3.1pt}

\begin{tabular}{cccccccccccccc}
\toprule
& & \multicolumn{12}{c}{\textbf{DUDE}} \\ \midrule
\multicolumn{2}{c}{}  & Phi4 & Molmo & Ovis & Llama & \begin{tabular}[c]{@{}c@{}}Llava\\  34B\end{tabular}  & \begin{tabular}[c]{@{}c@{}}Gemma3\\  27B\end{tabular} & \begin{tabular}[c]{@{}c@{}}Qwen2.5 VL\\  7B\end{tabular} &\begin{tabular}[c]{@{}c@{}}Qwen2.5 VL\\  72B\end{tabular} & \begin{tabular}[c]{@{}c@{}}InterVL3\\  9B\end{tabular} & \begin{tabular}[c]{@{}c@{}}InterVL3\\  78B\end{tabular} & \begin{tabular}[c]{@{}c@{}}GPT4.1\\  mini\end{tabular} & O3\\ \midrule
\multicolumn{2}{c|}{$Acc_D$}        & 0.070 & 0.230 & 0.241 & 0.289 & 0.401 & 0.503 & 0.460 & 0.599 & 0.267 & 0.326 & 0.214 & 0.239\\ \midrule
\multirow{3}{*}{\rotatebox[origin=c]{90}{Doc El}} 
& \multicolumn{1}{c|}{$<$15\%}      & 0.032 & 0.168 & 0.248 & 0.272 & 0.408 & 0.512 & 0.384 & 0.592 & 0.216 & 0.312 & 0.192 & 0.177 \\
& \multicolumn{1}{c|}{15\%-25\%}    & 0.000 & 0.048 & 0.008 & 0.048 & 0.072 & 0.040 & 0.096 & 0.080 & 0.048 & 0.032 & 0.016 & 0.073 \\
& \multicolumn{1}{c|}{$>$25\%}      & 0.072 & 0.128 & 0.104 & 0.112 & 0.120 & 0.200 & 0.208 & 0.224 & 0.136 & 0.144 & 0.112 & 0.045 \\ \midrule
\multirow{3}{*}{\rotatebox[origin=c]{90}{Layout}}
& \multicolumn{1}{c|}{$<$4 pages}   & 0.080 & 0.243 & 0.309 & 0.273 & 0.416 & 0.556 & 0.493 & 0.624 & 0.240 & 0.340 & 0.210 & 0.122 \\
& \multicolumn{1}{c|}{4-8 pages}    & 0.019 & 0.101 & 0.178 & 0.197 & 0.267 & 0.310 & 0.317 & 0.368 & 0.202 & 0.248 & 0.108 & 0.116 \\
& \multicolumn{1}{c|}{$>$8 pages}   & 0.000 & 0.069 & 0.000 & 0.069 & 0.232 & 0.417 & 0.347 & 0.458 & 0.200 & 0.042 & 0.014 & 0.039 \\ \midrule
\multirow{5}{*}{\rotatebox[origin=c]{90}{NLP Entity}}
& \multicolumn{1}{c|}{Numeric}      & 0.001 & 0.500 & 0.143 & 0.357 & 0.286 & 0.357 & 0.714 & 0.929 & 0.357 & 0.286 & 0.143 & 0.185 \\
& \multicolumn{1}{c|}{Temporal}     & 0.064 & 0.170 & 0.170 & 0.191 & 0.553 & 0.511 & 0.426 & 0.638 & 0.191 & 0.362 & 0.340 & 0.248 \\
& \multicolumn{1}{c|}{Misc}         & 0.120 & 0.270 & 0.390 & 0.340 & 0.460 & 0.610 & 0.580 & 0.790 & 0.350 & 0.410 & 0.300 & 0.312 \\
& \multicolumn{1}{c|}{Location}     & 0.020 & 0.204 & 0.204 & 0.306 & 0.469 & 0.735 & 0.429 & 0.510 & 0.306 & 0.367 & 0.122 & 0.196 \\
& \multicolumn{1}{c|}{Structure}    & 0.031 & 0.123 & 0.031 & 0.123 & 0.185 & 0.185 & 0.200 & 0.215 & 0.108 & 0.092 & 0.062 & 0.003 \\ \toprule
& & \multicolumn{12}{c}{\textbf{MPDocVQA}} \\ \midrule
\multicolumn{2}{c}{}  & Phi4 & Molmo & Ovis & Llama & \begin{tabular}[c]{@{}c@{}}Llava\\  34B\end{tabular}  & \begin{tabular}[c]{@{}c@{}}Gemma3\\  27B\end{tabular} & \begin{tabular}[c]{@{}c@{}}Qwen2.5 VL\\  7B\end{tabular} &\begin{tabular}[c]{@{}c@{}}Qwen2.5 VL\\  72B\end{tabular} & \begin{tabular}[c]{@{}c@{}}InterVL3\\  9B\end{tabular} & \begin{tabular}[c]{@{}c@{}}InterVL3\\  78B\end{tabular} & \begin{tabular}[c]{@{}c@{}}GPT4.1\\  mini\end{tabular} & O3\\ \midrule
\multicolumn{2}{c|}{$Acc_D$}        & 0.037 & 0.340 & 0.217 & 0.325 & 0.357 & 0.394 & 0.490 & 0.581 & 0.241 & 0.219 & 0.264 & 0.163\\ \midrule
\multirow{3}{*}{\rotatebox[origin=c]{90}{Doc El}} 
& \multicolumn{1}{c|}{$<$15\%}        & 0.033 & 0.354 & 0.227 & 0.331 & 0.340 & 0.392 & 0.492 & 0.572 & 0.243 & 0.213 & 0.265 & 0.166\\
& \multicolumn{1}{c|}{15\%-25\%}      & 0.000 & 0.019 & 0.019 & 0.028 & 0.094 & 0.075 & 0.056 & 0.104 & 0.019 & 0.028 & 0.009 & 0.019\\
& \multicolumn{1}{c|}{$>$25\%}        & 0.037 & 0.100 & 0.050 & 0.112 & 0.149 & 0.124 & 0.187 & 0.224 & 0.100 & 0.112 & 0.124 & 0.050\\ \midrule
\multirow{3}{*}{\rotatebox[origin=c]{90}{Layout}}
& \multicolumn{1}{c|}{$<$4 pages}     & 0.042 & 0.296 & 0.218 & 0.316 & 0.567 & 0.514 & 0.500 & 0.655 & 0.252 & 0.232 & 0.234 & 0.176\\
& \multicolumn{1}{c|}{4-8 pages}      & 0.059 & 0.402 & 0.360 & 0.414 & 0.159 & 0.556 & 0.569 & 0.598 & 0.331 & 0.331 & 0.468 & 0.355\\
& \multicolumn{1}{c|}{$>$8 pages}     & 0.041 & 0.440 & 0.141 & 0.370 & 0.131 & 0.235 & 0.526 & 0.571 & 0.286 & 0.200 & 0.317 & 0.166\\ \midrule
\multirow{5}{*}{\rotatebox[origin=c]{90}{NLP Entity}}
& \multicolumn{1}{c|}{Numeric}      & 0.007 & 0.340 & 0.163 & 0.340 & 0.313 & 0.299 & 0.442 & 0.565 & 0.143 & 0.197 & 0.197 & 0.116 \\
& \multicolumn{1}{c|}{Temporal}     & 0.149 & 0.511 & 0.277 & 0.553 & 0.340 & 0.383 & 0.638 & 0.660 & 0.362 & 0.255 & 0.468 & 0.298 \\
& \multicolumn{1}{c|}{Misc}         & 0.019 & 0.256 & 0.207 & 0.298 & 0.369 & 0.343 & 0.421 & 0.515 & 0.184 & 0.146 & 0.201 & 0.117 \\
& \multicolumn{1}{c|}{Location}     & 0.038 & 0.454 & 0.308 & 0.346 & 0.431 & 0.608 & 0.685 & 0.692 & 0.400 & 0.300 & 0.338 & 0.215 \\
& \multicolumn{1}{c|}{Structure}    & 0.118 & 0.265 & 0.176 & 0.265 & 0.412 & 0.265 & 0.235 & 0.529 & 0.176 & 0.147 & 0.206 & 0.088 \\
\bottomrule
\end{tabular}

\caption{Effect of the corruption type on the Document-Level Accuracy. Coarse-grained analysis (addressing RQ2).}
\label{tab:accdoc}
\end{table*}

\begin{table*}[h!]
\centering
\footnotesize
\setlength{\tabcolsep}{3.1pt}
\caption{Effect of the corruption type on the Page-Level Accuracy. Fine-grained analysis (addressing RQ2).}
\begin{tabular}{cccccccccccccc}
\toprule
& & \multicolumn{12}{c}{\textbf{DUDE}} \\ \midrule
\multicolumn{2}{c}{}  & Phi4 & Molmo & Ovis & Llama & \begin{tabular}[c]{@{}c@{}}Llava\\  34B\end{tabular}  & \begin{tabular}[c]{@{}c@{}}Gemma3\\  27B\end{tabular} & \begin{tabular}[c]{@{}c@{}}Qwen2.5 VL\\  7B\end{tabular} &\begin{tabular}[c]{@{}c@{}}Qwen2.5 VL\\  72B\end{tabular} & \begin{tabular}[c]{@{}c@{}}InterVL3\\  9B\end{tabular} & \begin{tabular}[c]{@{}c@{}}InterVL3\\  78B\end{tabular} & \begin{tabular}[c]{@{}c@{}}GPT4.1\\  mini\end{tabular} & O3\\ \midrule
\multicolumn{2}{c|}{$Acc_P$}       & 0.248 & 0.554 & 0.674 & 0.680 & 0.717 & 0.786 & 0.835 & 0.754 & 0.713 & 0.781 & 0.638 & 0.663 \\\midrule
\multirow{3}{*}{\rotatebox[origin=c]{90}{Doc El}} 
& \multicolumn{1}{c|}{0}    & 0.247 & 0.532 & 0.746 & 0.692 & 0.755 & 0.813 & 0.867 & 0.777 & 0.773 & 0.850 & 0.753 & 0.709 \\
& \multicolumn{1}{c|}{1}    & 0.240 & 0.555 & 0.627 & 0.658 & 0.685 & 0.784 & 0.812 & 0.759 & 0.661 & 0.692 & 0.531 & 0.575 \\
& \multicolumn{1}{c|}{$>$1} & 0.263 & 0.614 & 0.550 & 0.684 & 0.667 & 0.713 & 0.784 & 0.737 & 0.632 & 0.737 & 0.497 & 0.577 \\ \midrule
\multirow{2}{*}{\rotatebox[origin=c]{90}{Lay}}
& \multicolumn{1}{c|}{In-Page}      & 0.207 & 0.444 & 0.566 & 0.536 & 0.655 & 0.740 & 0.753 & 0.802 & 0.579 & 0.701 & 0.500 & 0.522 \\
& \multicolumn{1}{c|}{Out-Page}     & 0.267 & 0.606 & 0.725 & 0.748 & 0.747 & 0.808 & 0.873 & 0.700 & 0.777 & 0.819 & 0.703 & 0.712 \\ \midrule
\multirow{5}{*}{\rotatebox[origin=c]{90}{NLP Entity}}
& \multicolumn{1}{c|}{Numeric}      & 0.236 & 0.906 & 0.890 & 0.866 & 0.661 & 0.906 & 0.969 & 0.990 & 0.906 & 0.906 & 0.638 & 0.662 \\
& \multicolumn{1}{c|}{Temporal}     & 0.299 & 0.528 & 0.492 & 0.563 & 0.787 & 0.650 & 0.787 & 0.736 & 0.614 & 0.711 & 0.690 & 0.652 \\
& \multicolumn{1}{c|}{Misc}         & 0.233 & 0.448 & 0.724 & 0.678 & 0.724 & 0.856 & 0.848 & 0.858 & 0.701 & 0.767 & 0.626 & 0.529 \\
& \multicolumn{1}{c|}{Location}     & 0.183 & 0.409 & 0.668 & 0.545 & 0.800 & 0.902 & 0.851 & 0.713 & 0.749 & 0.770 & 0.574 & 0.484 \\
& \multicolumn{1}{c|}{Structure}    & 0.233 & 0.571 & 0.568 & 0.682 & 0.673 & 0.652 & 0.774 & 0.602 & 0.647 & 0.741 & 0.641 & 0.727 \\ \toprule
& & \multicolumn{12}{c}{\textbf{MPDocVQA}} \\ \midrule
& & Phi & Qwen & Molmo & InternVL & DocOwl & Ovis & Llama & Gemma & Llava & UDOP & LayoutLMv3 & BLIP\\ \midrule
\multicolumn{2}{c|}{$Acc_P$}        & 0.211 & 0.780 & 0.792 & 0.796 & 0.708 & 0.838 & 0.881 & 0.842 & 0.782 & 0.818 & 0.775 & 0.738 \\ \midrule
\multirow{3}{*}{\rotatebox[origin=c]{90}{Doc El}} 
& \multicolumn{1}{c|}{0}    & 0.231 & 0.761 & 0.839 & 0.772 & 0.726 & 0.869 & 0.881 & 0.851 & 0.808 & 0.864 & 0.793 & 0.761 \\
& \multicolumn{1}{c|}{1}    & 0.203 & 0.794 & 0.769 & 0.823 & 0.700 & 0.807 & 0.889 & 0.858 & 0.776 & 0.784 & 0.758 & 0.714 \\
& \multicolumn{1}{c|}{$>$1} & 0.154 & 0.817 & 0.667 & 0.812 & 0.658 & 0.803 & 0.858 & 0.832 & 0.699 & 0.736 & 0.751 & 0.716 \\ \midrule
\multirow{2}{*}{\rotatebox[origin=c]{90}{Lay}}
& \multicolumn{1}{c|}{In-Page}      & 0.184 & 0.620 & 0.638 & 0.638 & 0.705 & 0.758 & 0.800 & 0.792 & 0.609 & 0.661 & 0.577 & 0.563 \\
& \multicolumn{1}{c|}{Out-Page}     & 0.221 & 0.835 & 0.844 & 0.850 & 0.709 & 0.865 & 0.909 & 0.878 & 0.842 & 0.872 & 0.842 & 0.798 \\\midrule
\multirow{5}{*}{\rotatebox[origin=c]{90}{NLP Entity}}
& \multicolumn{1}{c|}{Numeric}       & 0.258 & 0.820 & 0.799 & 0.829 & 0.715 & 0.836 & 0.890 & 0.842 & 0.766 & 0.810 & 0.766 & 0.773 \\
& \multicolumn{1}{c|}{Temporal}      & 0.276 & 0.944 & 0.897 & 0.949 & 0.774 & 0.850 & 0.970 & 0.950 & 0.909 & 0.899 & 0.937 & 0.923 \\
& \multicolumn{1}{c|}{Misc}          & 0.161 & 0.668 & 0.776 & 0.702 & 0.657 & 0.829 & 0.829 & 0.813 & 0.702 & 0.792 & 0.713 & 0.675 \\
& \multicolumn{1}{c|}{Location}      & 0.182 & 0.809 & 0.703 & 0.752 & 0.749 & 0.825 & 0.904 & 0.819 & 0.797 & 0.775 & 0.682 & 0.603 \\
& \multicolumn{1}{c|}{Structure}     & 0.258 & 0.682 & 0.732 & 0.778 & 0.783 & 0.768 & 0.843 & 0.801 & 0.758 & 0.793 & 0.773 & 0.692 \\
\bottomrule
\end{tabular}
\label{tab:accpage}
\end{table*}

\begin{sidewaystable*}[h!]
\tiny
\centering
\setlength{\tabcolsep}{3pt}

\begin{tabular}{@{}p{2mm}cccc|c|cc|cc|cc|cc|cc|cc|cc|cc|cc|cc@{}}
\toprule
\multicolumn{3}{c}{}  & \multicolumn{2}{c}{Phi4} & \multicolumn{1}{c}{Molmo} & \multicolumn{2}{c}{Ovis} & \multicolumn{2}{c}{Llama} & \multicolumn{2}{c}{Llava 34B}  & \multicolumn{2}{c}{Gemma 27B} & \multicolumn{2}{c}{Qwen 2.5 7B} & \multicolumn{2}{c}{Qwen 2.5 72B} & \multicolumn{2}{c}{InternVL 3 9B} & \multicolumn{2}{c}{InternVL 3 78B} & \multicolumn{2}{c}{GPT-4.1-mini} & \multicolumn{2}{c}{O3}\\ \midrule
                       
& & 
  & Explicit & \begin{tabular}[c]{@{}c@{}}OCR\\  Explicit\end{tabular}      
  & Explicit   
  & Explicit & \begin{tabular}[c]{@{}c@{}}OCR\\  Explicit\end{tabular}  
  & Explicit & \begin{tabular}[c]{@{}c@{}}OCR \\ Explicit\end{tabular}          
  & Explicit & \begin{tabular}[c]{@{}c@{}}OCR \\ Explicit\end{tabular}       
  & Explicit & \begin{tabular}[c]{@{}c@{}}OCR \\ Explicit\end{tabular}       
  & Explicit & \begin{tabular}[c]{@{}c@{}}OCR\\  Explicit\end{tabular} 
  & Explicit & \begin{tabular}[c]{@{}c@{}}OCR\\  Explicit\end{tabular} 
  & Explicit & \begin{tabular}[c]{@{}c@{}}OCR \\ Explicit\end{tabular}
  & Explicit & \begin{tabular}[c]{@{}c@{}}OCR \\ Explicit\end{tabular}
  & Explicit & \begin{tabular}[c]{@{}c@{}}OCR \\ Explicit\end{tabular}
  & Explicit & \begin{tabular}[c]{@{}c@{}}OCR \\ Explicit\end{tabular}\\ \midrule

\multicolumn{2}{c}{\multirow{3}{*}{$Acc_D$}}   
& \multicolumn{1}{c|}{C1}                                                                                                                                  &  0.079 & 0.439 & 0.254 & 0.281 & 0.509 & 0.342 & 0.325 & 0.377 & 0.325 & 0.482 & 0.430 & 0.465 & 0.570 & 0.588 & 0.649 & 0.281 & 0.404 & 0.342 & 0.404 & 0.202 & 0.298 & 0.227 & 0.319 \\
\multicolumn{2}{c}{} & \multicolumn{1}{c|}{C2}                                                                                                             &  0.052 & 0.534 & 0.190 & 0.172 & 0.190 & 0.224 & 0.362 & 0.483 & 0.431 & 0.586 & 0.310 & 0.517 & 0.672 & 0.707 & 0.741 & 0.259 & 0.397 & 0.328 & 0.397 & 0.276 & 0.276 & 0.301 & 0.080 \\
\multicolumn{2}{c}{} & \multicolumn{1}{c|}{C3}                                                                                                             &  0.067 & 0.133 & 0.200 & 0.200 & 0.133 & 0.133 & 0.267 & 0.267 & 0.267 & 0.333 & 0.133 & 0.200 & 0.267 & 0.267 & 0.467 & 0.200 & 0.200 & 0.200 & 0.133 & 0.067 & 0.067 & 0.092 & 0.066 \\ \midrule
\multirow{10}{*}{\rotatebox[origin=c]{90}{{\scriptsize Document Element}}} & \multicolumn{1}{c}{\multirow{3}{*}{$<$15\%}} & \multicolumn{1}{c|}{C1}        &  0.042 & 0.306 & 0.208 & 0.306 & 0.500 & 0.333 & 0.278 & 0.389 & 0.278 & 0.514 & 0.319 & 0.389 & 0.431 & 0.583 & 0.611 & 0.236 & 0.347 & 0.333 & 0.347 & 0.167 & 0.278 & 0.319 & 0.297 \\
\multicolumn{2}{c}{} & \multicolumn{1}{c|}{C2}                                                                                                             &  0.023 & 0.419 & 0.093 & 0.186 & 0.186 & 0.209 & 0.302 & 0.442 & 0.395 & 0.535 & 0.233 & 0.442 & 0.605 & 0.698 & 0.698 & 0.186 & 0.395 & 0.302 & 0.395 & 0.279 & 0.279 & 0.256 & 0.353 \\
\multicolumn{2}{c}{} & \multicolumn{1}{c|}{C3}                                                                                                             &  0.023 & 0.419 & 0.093 & 0.186 & 0.186 & 0.209 & 0.302 & 0.442 & 0.395 & 0.535 & 0.233 & 0.442 & 0.605 & 0.698 & 0.698 & 0.186 & 0.395 & 0.302 & 0.395 & 0.279 & 0.279 & 0.256 & 0.296 \\ \cmidrule{2-26}
& \multicolumn{1}{c}{\multirow{3}{*}{15\%-25\%}} & \multicolumn{1}{c|}{C1}                                                                                 &  0.000 & 0.040 & 0.040 & 0.010 & 0.110 & 0.040 & 0.020 & 0.080 & 0.010 & 0.020 & 0.080 & 0.100 & 0.130 & 0.080 & 0.120 & 0.040 & 0.070 & 0.030 & 0.090 & 0.010 & 0.010 & 0.168 & 0.002 \\
\multicolumn{2}{c}{} & \multicolumn{1}{c|}{C2}                                                                                                             &  0.000 & 0.080 & 0.160 & 0.000 & 0.080 & 0.160 & 0.080 & 0.080 & 0.080 & 0.160 & 0.160 & 0.160 & 0.160 & 0.160 & 0.160 & 0.160 & 0.160 & 0.080 & 0.160 & 0.080 & 0.080 & 0.157 & 0.050 \\
\multicolumn{2}{c}{} & \multicolumn{1}{c|}{C3}                                                                                                             &  0.000 & 0.080 & 0.160 & 0.000 & 0.080 & 0.160 & 0.080 & 0.080 & 0.080 & 0.160 & 0.160 & 0.160 & 0.160 & 0.160 & 0.160 & 0.160 & 0.160 & 0.080 & 0.160 & 0.080 & 0.080 & 0.033 & 0.068 \\ \cmidrule{2-26}
& \multicolumn{1}{c}{\multirow{3}{*}{$>$25\%}} & \multicolumn{1}{c|}{C1}                                                                                   &  0.078 & 0.310 & 0.129 & 0.116 & 0.142 & 0.142 & 0.194 & 0.090 & 0.207 & 0.207 & 0.233 & 0.194 & 0.271 & 0.220 & 0.233 & 0.142 & 0.181 & 0.155 & 0.155 & 0.129 & 0.168 & 0.183 & 0.096 \\
\multicolumn{2}{c}{} & \multicolumn{1}{c|}{C2}                                                                                                             &  0.052 & 0.310 & 0.129 & 0.052 & 0.052 & 0.052 & 0.181 & 0.207 & 0.181 & 0.233 & 0.155 & 0.233 & 0.284 & 0.233 & 0.284 & 0.129 & 0.103 & 0.129 & 0.103 & 0.078 & 0.078 & 0.031 & 0.031 \\
\multicolumn{2}{c}{} & \multicolumn{1}{c|}{C3}                                                                                                             &  0.052 & 0.310 & 0.129 & 0.052 & 0.052 & 0.052 & 0.181 & 0.207 & 0.181 & 0.233 & 0.155 & 0.233 & 0.284 & 0.233 & 0.284 & 0.129 & 0.103 & 0.129 & 0.103 & 0.078 & 0.078 & 0.031 & 0.183 \\ \midrule
\multirow{10}{*}{\rotatebox[origin=c]{90}{{\scriptsize Layout}}} & \multicolumn{1}{c}{\multirow{3}{*}{$<$4 pages}} & \multicolumn{1}{c|}{C1}               &  0.099 & 0.626 & 0.273 & 0.424 & 0.308 & 0.363 & 0.319 & 0.446 & 0.403 & 0.525 & 0.516 & 0.606 & 0.581 & 0.643 & 0.836 & 0.282 & 0.667 & 0.373 & 0.542 & 0.210 & 0.280 & 0.234 & 0.315 \\
\multicolumn{2}{c}{} & \multicolumn{1}{c|}{C2}                                                                                                             &  0.036 & 0.446 & 0.179 & 0.107 & 0.143 & 0.143 & 0.304 & 0.429 & 0.357 & 0.536 & 0.268 & 0.393 & 0.536 & 0.607 & 0.607 & 0.179 & 0.196 & 0.250 & 0.250 & 0.196 & 0.179 & 0.005 & 0.002 \\
\multicolumn{2}{c}{} & \multicolumn{1}{c|}{C3}                                                                                                             &  0.062 & 0.125 & 0.125 & 0.188 & 0.125 & 0.125 & 0.250 & 0.375 & 0.438 & 0.312 & 0.125 & 0.188 & 0.312 & 0.250 & 0.562 & 0.125 & 0.188 & 0.188 & 0.125 & 0.062 & 0.062 & 0.110 & 0.095 \\ \cmidrule{2-26}
& \multicolumn{1}{c}{\multirow{3}{*}{4-8 pages}} & \multicolumn{1}{c|}{C1}                                                                                 &  0.000 & 0.258 & 0.113 & 0.148 & 0.547 & 0.220 & 0.258 & 0.330 & 0.258 & 0.261 & 0.148 & 0.294 & 0.398 & 0.368 & 0.401 & 0.187 & 0.223 & 0.223 & 0.148 & 0.038 & 0.181 & 0.018 & 0.143 \\
\multicolumn{2}{c}{} & \multicolumn{1}{c|}{C2}                                                                                                             &  0.056 & 0.333 & 0.000 & 0.222 & 0.111 & 0.167 & 0.222 & 0.111 & 0.222 & 0.389 & 0.111 & 0.444 & 0.611 & 0.444 & 0.333 & 0.278 & 0.889 & 0.278 & 0.556 & 0.222 & 0.278 & 0.121 & 0.210 \\
\multicolumn{2}{c}{} & \multicolumn{1}{c|}{C3}                                                                                                             &  0.000 & 0.000 & 0.167 & 0.000 & 0.000 & 0.000 & 0.167 & 0.167 & 0.000 & 0.167 & 0.000 & 0.000 & 0.000 & 0.000 & 0.000 & 0.000 & 0.000 & 0.000 & 0.000 & 0.000 & 0.000 & 0.031 & 0.061 \\ \cmidrule{2-26}
& \multicolumn{1}{c}{\multirow{3}{*}{$>$8 pages}} & \multicolumn{1}{c|}{C1}                                                                                &  0.000 & 0.103 & 0.026 & 0.000 & 0.709 & 0.026 & 0.000 & 0.167 & 0.051 & 0.359 & 0.251 & 0.344 & 0.410 & 0.462 & 0.436 & 0.118 & 0.318 & 0.077 & 0.251 & 0.000 & 0.077 & 0.091 & 0.180 \\
\multicolumn{2}{c}{} & \multicolumn{1}{c|}{C2}                                                                                                             &  0.000 & 0.111 & 0.333 & 0.000 & 0.333 & 0.333 & 0.000 & 0.111 & 0.111 & 0.370 & 0.333 & 0.407 & 0.481 & 0.481 & 0.556 & 0.370 & 0.370 & 0.000 & 0.407 & 0.037 & 0.000 & 0.104 & 0.093 \\
\multicolumn{2}{c}{} & \multicolumn{1}{c|}{C3}                                                                                                             &  0.000 & 0.000 & 0.000 & 0.000 & 0.000 & 0.000 & 0.000 & 0.333 & 0.000 & 0.333 & 0.000 & 0.000 & 0.000 & 0.000 & 0.167 & 0.333 & 0.000 & 0.000 & 0.000 & 0.000 & 0.000 & 0.147 & 0.084 \\ \midrule
\multirow{18}{*}{\rotatebox[origin=c]{90}{{\scriptsize NLP Entity}}} & \multicolumn{1}{c}{\multirow{3}{*}{Numeric}} & \multicolumn{1}{c|}{C1}              &  0.000 & 0.000 & 0.250 & 0.250 & 0.750 & 0.250 & 0.250 & 0.125 & 0.250 & 0.125 & 0.500 & 0.750 & 0.875 & 1.000 & 1.000 & 0.250 & 0.500 & 0.250 & 0.500 & 0.125 & 0.125 & 0.102 & 0.094 \\
\multicolumn{2}{c}{} & \multicolumn{1}{c|}{C2}                                                                                                             &  0.000 & 0.500 & 0.833 & 0.000 & 0.333 & 0.500 & 0.500 & 0.500 & 0.333 & 0.667 & 0.667 & 0.667 & 0.833 & 0.833 & 1.000 & 0.500 & 0.500 & 0.333 & 0.500 & 0.167 & 0.167 & 0.173 & 0.200 \\
\multicolumn{2}{c}{} & \multicolumn{1}{c|}{C3}                                                                                                             &  0.000 & 0.000 & 0.000 & 0.000 & 0.000 & 0.000 & 0.000 & 0.000 & 0.000 & 0.000 & 0.000 & 0.000 & 0.000 & 0.000 & 0.000 & 0.000 & 0.000 & 0.000 & 0.000 & 0.000 & 0.000 & 0.142 & 0.098 \\ \cmidrule{2-26}
& \multicolumn{1}{c}{\multirow{3}{*}{Temporal}} & \multicolumn{1}{c|}{C1}                                                                                  &  0.091 & 0.545 & 0.364 & 0.182 & 0.455 & 0.364 & 0.364 & 0.545 & 0.545 & 0.636 & 0.636 & 0.364 & 0.545 & 0.727 & 0.909 & 0.273 & 0.364 & 0.455 & 0.364 & 0.182 & 0.182 & 0.127 & 0.134 \\
\multicolumn{2}{c}{} & \multicolumn{1}{c|}{C2}                                                                                                             &  0.042 & 0.583 & 0.083 & 0.167 & 0.333 & 0.167 & 0.333 & 0.708 & 0.667 & 0.583 & 0.250 & 0.583 & 0.667 & 0.792 & 0.750 & 0.208 & 0.125 & 0.417 & 0.500 & 0.542 & 0.250 & 0.553 & 0.231 \\
\multicolumn{2}{c}{} & \multicolumn{1}{c|}{C3}                                                                                                             &  0.083 & 0.167 & 0.167 & 0.167 & 0.083 & 0.083 & 0.417 & 0.250 & 0.167 & 0.250 & 0.083 & 0.167 & 0.167 & 0.250 & 0.667 & 0.083 & 0.167 & 0.167 & 0.083 & 0.083 & 0.083 & 0.032 & 0.027 \\ \cmidrule{2-26}
& \multicolumn{1}{c}{\multirow{3}{*}{Misc}} & \multicolumn{1}{c|}{C1}                                                                                      &  0.139 & 0.583 & 0.361 & 0.528 & 0.556 & 0.444 & 0.528 & 0.528 & 0.556 & 0.667 & 0.528 & 0.556 & 0.722 & 0.861 & 0.917 & 0.389 & 0.500 & 0.472 & 0.556 & 0.389 & 0.611 & 0.426 & 0.491 \\
\multicolumn{2}{c}{} & \multicolumn{1}{c|}{C2}                                                                                                             &  0.098 & 0.569 & 0.196 & 0.275 & 0.196 & 0.275 & 0.392 & 0.451 & 0.412 & 0.667 & 0.373 & 0.627 & 0.745 & 0.784 & 0.784 & 0.333 & 0.529 & 0.353 & 0.451 & 0.275 & 0.431 & 0.214 & 0.513 \\
\multicolumn{2}{c}{} & \multicolumn{1}{c|}{C3}                                                                                                             &  0.154 & 0.308 & 0.308 & 0.462 & 0.308 & 0.308 & 0.308 & 0.308 & 0.692 & 0.231 & 0.308 & 0.462 & 0.462 & 0.615 & 0.538 & 0.308 & 0.462 & 0.462 & 0.308 & 0.154 & 0.154 & 0.125 & 0.289 \\ \cmidrule{2-26}
& \multicolumn{1}{c}{\multirow{3}{*}{Location}} & \multicolumn{1}{c|}{C1}                                                                                  &  0.036 & 0.536 & 0.179 & 0.250 & 0.750 & 0.393 & 0.214 & 0.393 & 0.143 & 0.679 & 0.464 & 0.500 & 0.643 & 0.429 & 0.464 & 0.321 & 0.429 & 0.357 & 0.393 & 0.143 & 0.250 & 0.083 & 0.243 \\
\multicolumn{2}{c}{} & \multicolumn{1}{c|}{C2}                                                                                                             &  0.000 & 0.750 & 0.250 & 0.125 & 0.125 & 0.188 & 0.438 & 0.562 & 0.438 & 0.812 & 0.375 & 0.375 & 0.750 & 0.750 & 0.812 & 0.125 & 0.250 & 0.438 & 0.188 & 0.125 & 0.188 & 0.310 & 0.288 \\
\multicolumn{2}{c}{} & \multicolumn{1}{c|}{C3}                                                                                                             &  0.000 & 0.000 & 0.200 & 0.200 & 0.200 & 0.200 & 0.200 & 0.600 & 0.200 & 0.800 & 0.200 & 0.200 & 0.200 & 0.200 & 0.400 & 0.800 & 0.200 & 0.200 & 0.200 & 0.000 & 0.000 & 0.001 & 0.036 \\ \cmidrule{2-26}
& \multicolumn{1}{c}{\multirow{3}{*}{Structure}} & \multicolumn{1}{c|}{C1}                                                                                 &  0.065 & 0.258 & 0.161 & 0.065 & 0.194 & 0.194 & 0.194 & 0.194 & 0.161 & 0.129 & 0.194 & 0.290 & 0.258 & 0.258 & 0.323 & 0.129 & 0.258 & 0.161 & 0.226 & 0.065 & 0.065 & 0.041 & 0.000 \\
\multicolumn{2}{c}{} & \multicolumn{1}{c|}{C2}                                                                                                             &  0.000 & 0.211 & 0.053 & 0.000 & 0.000 & 0.105 & 0.211 & 0.211 & 0.211 & 0.158 & 0.053 & 0.211 & 0.368 & 0.316 & 0.474 & 0.158 & 0.474 & 0.053 & 0.263 & 0.105 & 0.000 & 0.188 & 0.036 \\
\multicolumn{2}{c}{} & \multicolumn{1}{c|}{C3}                                                                                                             &  0.000 & 0.000 & 0.133 & 0.000 & 0.000 & 0.000 & 0.133 & 0.133 & 0.000 & 0.333 & 0.000 & 0.000 & 0.200 & 0.000 & 0.267 & 0.000 & 0.000 & 0.000 & 0.000 & 0.000 & 0.000 & 0.122 & 0.154 \\
\bottomrule
\end{tabular}

 \caption{Effect of the corruption type on the Document-Level Accuracy by varying in-context learning strategy and complexity level (addressing RQ2 and RQ3). DUDE dataset.}
\label{tab:ablationDUDEDoc}
\end{sidewaystable*}

\begin{sidewaystable*}[h!]
\tiny
\centering
\setlength{\tabcolsep}{3pt}
 
\begin{tabular}{@{}p{2mm}cccc|c|cc|cc|cc|cc|cc|cc|cc|cc|cc|cc@{}}
\toprule
\multicolumn{3}{c}{}  & \multicolumn{2}{c}{Phi4} & \multicolumn{1}{c}{Molmo} & \multicolumn{2}{c}{Ovis} & \multicolumn{2}{c}{Llama} & \multicolumn{2}{c}{Llava 34B}  & \multicolumn{2}{c}{Gemma 27B} & \multicolumn{2}{c}{Qwen 2.5 7B} & \multicolumn{2}{c}{Qwen 2.5 72B} & \multicolumn{2}{c}{InternVL 3 9B} & \multicolumn{2}{c}{InternVL 3 78B} & \multicolumn{2}{c}{GPT-4.1-mini} & \multicolumn{2}{c}{O3}\\ \midrule
                       
& & 
  & Explicit & \begin{tabular}[c]{@{}c@{}}OCR\\  Explicit\end{tabular}      
  & Explicit   
  & Explicit & \begin{tabular}[c]{@{}c@{}}OCR\\  Explicit\end{tabular}  
  & Explicit & \begin{tabular}[c]{@{}c@{}}OCR \\ Explicit\end{tabular}          
  & Explicit & \begin{tabular}[c]{@{}c@{}}OCR \\ Explicit\end{tabular}       
  & Explicit & \begin{tabular}[c]{@{}c@{}}OCR \\ Explicit\end{tabular}       
  & Explicit & \begin{tabular}[c]{@{}c@{}}OCR\\  Explicit\end{tabular} 
  & Explicit & \begin{tabular}[c]{@{}c@{}}OCR\\  Explicit\end{tabular} 
  & Explicit & \begin{tabular}[c]{@{}c@{}}OCR \\ Explicit\end{tabular}
  & Explicit & \begin{tabular}[c]{@{}c@{}}OCR \\ Explicit\end{tabular}
  & Explicit & \begin{tabular}[c]{@{}c@{}}OCR \\ Explicit\end{tabular}
  & Explicit & \begin{tabular}[c]{@{}c@{}}OCR \\ Explicit\end{tabular}\\ \midrule

\multicolumn{2}{c}{\multirow{3}{*}{$Acc_P$}}   & \multicolumn{1}{c|}{C1}                                                                                &  0.266 & 0.750 & 0.577 & 0.723 & 0.638 & 0.712 & 0.699 & 0.701 & 0.738 & 0.810 & 0.772 & 0.843 & 0.870 & 0.753 & 0.812 & 0.738 & 0.783 & 0.805 & 0.827 & 0.636 & 0.724 & 0.661 & 0.618   \\
\multicolumn{2}{c}{} & \multicolumn{1}{c|}{C2}                                                                                                          &  0.240 & 0.818 & 0.542 & 0.615 & 0.593 & 0.655 & 0.742 & 0.760 & 0.771 & 0.764 & 0.724 & 0.847 & 0.920 & 0.816 & 0.896 & 0.684 & 0.778 & 0.760 & 0.800 & 0.669 & 0.691 & 0.694 & 0.773   \\
\multicolumn{2}{c}{} & \multicolumn{1}{c|}{C3}                                                                                                          &  0.141 & 0.692 & 0.423 & 0.513 & 0.551 & 0.526 & 0.654 & 0.692 & 0.744 & 0.679 & 0.692 & 0.731 & 0.731 & 0.519 & 0.712 & 0.628 & 0.603 & 0.667 & 0.667 & 0.538 & 0.590 & 0.563 & 0.468   \\ \midrule
\multirow{10}{*}{\rotatebox[origin=c]{90}{{\scriptsize Document Element}}} & \multicolumn{1}{c}{\multirow{3}{*}{0}} & \multicolumn{1}{c|}{C1}           &  0.244 & 0.674 & 0.498 & 0.787 & 0.581 & 0.694 & 0.735 & 0.725 & 0.773 & 0.842 & 0.780 & 0.859 & 0.873 & 0.688 & 0.682 & 0.766 & 0.784 & 0.856 & 0.863 & 0.742 & 0.808 & 0.895 & 0.828   \\
\multicolumn{2}{c}{} & \multicolumn{1}{c|}{C2}                                                                                                          &  0.283 & 0.799 & 0.610 & 0.667 & 0.623 & 0.698 & 0.767 & 0.799 & 0.830 & 0.748 & 0.736 & 0.887 & 0.962 & 0.958 & 0.958 & 0.767 & 0.824 & 0.836 & 0.887 & 0.767 & 0.811 & 0.744 & 0.615   \\
\multicolumn{2}{c}{} & \multicolumn{1}{c|}{C3}                                                                                                          &  0.283 & 0.799 & 0.610 & 0.667 & 0.623 & 0.698 & 0.767 & 0.799 & 0.830 & 0.748 & 0.736 & 0.887 & 0.962 & 0.958 & 0.958 & 0.767 & 0.824 & 0.836 & 0.887 & 0.767 & 0.811 & 0.744 & 0.679    \\ \cmidrule{2-26}
& \multicolumn{1}{c}{\multirow{3}{*}{1}} & \multicolumn{1}{c|}{C1}                                                                                      &  0.264 & 0.862 & 0.621 & 0.707 & 0.667 & 0.741 & 0.672 & 0.678 & 0.730 & 0.787 & 0.822 & 0.833 & 0.868 & 0.719 & 0.842 & 0.736 & 0.816 & 0.759 & 0.816 & 0.557 & 0.667 & 0.715 & 0.686   \\
\multicolumn{2}{c}{} & \multicolumn{1}{c|}{C2}                                                                                                          &  0.198 & 0.840 & 0.469 & 0.556 & 0.556 & 0.556 & 0.691 & 0.704 & 0.716 & 0.827 & 0.691 & 0.815 & 0.877 & 0.881 & 0.881 & 0.593 & 0.741 & 0.605 & 0.691 & 0.519 & 0.543 & 0.595 & 0.617   \\
\multicolumn{2}{c}{} & \multicolumn{1}{c|}{C3}                                                                                                          &  0.198 & 0.840 & 0.469 & 0.556 & 0.556 & 0.556 & 0.691 & 0.704 & 0.716 & 0.827 & 0.691 & 0.815 & 0.877 & 0.881 & 0.881 & 0.593 & 0.741 & 0.605 & 0.691 & 0.519 & 0.543 & 0.472 & 0.560   \\ \cmidrule{2-26}
& \multicolumn{1}{c}{\multirow{3}{*}{$>$1}} & \multicolumn{1}{c|}{C1}                                                                                   &  0.317 & 0.770 & 0.698 & 0.595 & 0.730 & 0.714 & 0.651 & 0.675 & 0.667 & 0.770 & 0.683 & 0.817 & 0.865 & 0.804 & 0.889 & 0.675 & 0.738 & 0.754 & 0.762 & 0.500 & 0.611 & 0.554 & 0.600   \\
\multicolumn{2}{c}{} & \multicolumn{1}{c|}{C2}                                                                                                          &  0.143 & 0.857 & 0.400 & 0.514 & 0.543 & 0.686 & 0.743 & 0.714 & 0.629 & 0.686 & 0.743 & 0.743 & 0.829 & 0.667 & 0.851 & 0.514 & 0.657 & 0.771 & 0.657 & 0.571 & 0.486 & 0.525 & 0.456   \\
\multicolumn{2}{c}{} & \multicolumn{1}{c|}{C3}                                                                                                          &  0.143 & 0.857 & 0.400 & 0.514 & 0.543 & 0.686 & 0.743 & 0.714 & 0.629 & 0.686 & 0.743 & 0.743 & 0.829 & 0.667 & 0.851 & 0.514 & 0.657 & 0.771 & 0.657 & 0.571 & 0.486 & 0.525 & 0.338   \\ \midrule
\multirow{7}{*}{\rotatebox[origin=c]{90}{{\scriptsize Layout}}} & \multicolumn{1}{c}{\multirow{3}{*}{In-Page}} & \multicolumn{1}{c|}{C1}                &  0.254 & 0.725 & 0.551 & 0.696 & 0.551 & 0.645 & 0.681 & 0.638 & 0.681 & 0.761 & 0.768 & 0.775 & 0.833 & 0.816 & 0.832 & 0.623 & 0.739 & 0.746 & 0.775 & 0.478 & 0.645 & 0.502 & 0.573   \\
\multicolumn{2}{c}{} & \multicolumn{1}{c|}{C2}                                                                                                          &  0.185 & 0.758 & 0.371 & 0.492 & 0.500 & 0.492 & 0.621 & 0.677 & 0.718 & 0.790 & 0.653 & 0.782 & 0.911 & 0.869 & 0.897 & 0.548 & 0.653 & 0.694 & 0.710 & 0.565 & 0.573 & 0.373 & 0.527   \\
\multicolumn{2}{c}{} & \multicolumn{1}{c|}{C3}                                                                                                          &  0.119 & 0.595 & 0.310 & 0.357 & 0.405 & 0.310 & 0.524 & 0.643 & 0.643 & 0.524 & 0.643 & 0.595 & 0.643 & 0.533 & 0.756 & 0.524 & 0.476 & 0.571 & 0.571 & 0.381 & 0.452 & 0.208 & 0.558   \\ \cmidrule{2-26}
& \multicolumn{1}{c}{\multirow{3}{*}{Out-Page}} & \multicolumn{1}{c|}{C1}                                                                               &  0.269 & 0.757 & 0.585 & 0.731 & 0.664 & 0.733 & 0.704 & 0.720 & 0.755 & 0.826 & 0.773 & 0.863 & 0.881 & 0.712 & 0.798 & 0.773 & 0.797 & 0.823 & 0.843 & 0.684 & 0.748 & 0.628 & 0.783  \\
\multicolumn{2}{c}{} & \multicolumn{1}{c|}{C2}                                                                                                          &  0.285 & 0.868 & 0.682 & 0.715 & 0.669 & 0.788 & 0.841 & 0.828 & 0.815 & 0.742 & 0.781 & 0.901 & 0.927 & 0.679 & 0.893 & 0.795 & 0.881 & 0.815 & 0.874 & 0.755 & 0.788 & 0.654 & 0.612   \\
\multicolumn{2}{c}{} & \multicolumn{1}{c|}{C3}                                                                                                          &  0.167 & 0.806 & 0.556 & 0.694 & 0.722 & 0.778 & 0.806 & 0.750 & 0.861 & 0.861 & 0.750 & 0.889 & 0.833 & 0.429 & 0.429 & 0.750 & 0.750 & 0.778 & 0.778 & 0.722 & 0.750 & 0.754 & 0.782   \\ \midrule
\multirow{18}{*}{\rotatebox[origin=c]{90}{{\scriptsize NLP Entity}}} & \multicolumn{1}{c}{\multirow{3}{*}{Numeric}} & \multicolumn{1}{c|}{C1}           &  0.235 & 0.864 & 0.864 & 0.901 & 0.975 & 0.840 & 0.605 & 0.605 & 0.753 & 0.877 & 0.951 & 0.975 & 0.988 & 1.000 & 1.000 & 0.901 & 0.951 & 0.901 & 0.951 & 0.617 & 0.765 & 0.526 & 0.727    \\
\multicolumn{2}{c}{} & \multicolumn{1}{c|}{C2}                                                                                                          &  0.239 & 0.935 & 0.978 & 0.870 & 0.870 & 0.913 & 0.848 & 0.761 & 0.826 & 0.957 & 0.957 & 0.957 & 0.978 & 0.974 & 1.000 & 0.913 & 0.935 & 0.913 & 0.913 & 0.674 & 0.783 & 0.533 & 0.715    \\
\multicolumn{2}{c}{} & \multicolumn{1}{c|}{C3}                                                                                                          &  0.000 & 0.000 & 0.000 & 0.000 & 0.000 & 0.000 & 0.000 & 0.000 & 0.000 & 0.000 & 0.000 & 0.000 & 0.000 & 0.000 & 0.000 & 0.000 & 0.000 & 0.000 & 0.000 & 0.000 & 0.000 & 0.147 & 0.061    \\ \cmidrule{2-26}
& \multicolumn{1}{c}{\multirow{3}{*}{Temporal}} & \multicolumn{1}{c|}{C1}                                                                               &  0.273 & 0.727 & 0.394 & 0.394 & 0.606 & 0.636 & 0.606 & 0.788 & 0.636 & 0.758 & 0.606 & 0.697 & 0.667 & 0.885 & 0.962 & 0.545 & 0.576 & 0.606 & 0.697 & 0.455 & 0.394 & 0.432 & 0.497    \\
\multicolumn{2}{c}{} & \multicolumn{1}{c|}{C2}                                                                                                          &  0.359 & 0.803 & 0.556 & 0.453 & 0.479 & 0.504 & 0.658 & 0.855 & 0.863 & 0.641 & 0.573 & 0.855 & 0.932 & 0.800 & 0.859 & 0.607 & 0.641 & 0.718 & 0.778 & 0.744 & 0.658 & 0.750 & 0.751    \\
\multicolumn{2}{c}{} & \multicolumn{1}{c|}{C3}                                                                                                          &  0.170 & 0.702 & 0.553 & 0.660 & 0.702 & 0.660 & 0.723 & 0.617 & 0.660 & 0.596 & 0.681 & 0.681 & 0.681 & 0.414 & 0.828 & 0.681 & 0.660 & 0.766 & 0.745 & 0.723 & 0.702 & 0.581 & 0.618    \\ \cmidrule{2-26}
& \multicolumn{1}{c}{\multirow{3}{*}{Misc}} & \multicolumn{1}{c|}{C1}                                                                                   &  0.310 & 0.814 & 0.540 & 0.823 & 0.743 & 0.735 & 0.805 & 0.805 & 0.823 & 0.876 & 0.814 & 0.858 & 0.912 & 0.940 & 0.952 & 0.743 & 0.814 & 0.832 & 0.841 & 0.699 & 0.858 & 0.645 & 0.827    \\
\multicolumn{2}{c}{} & \multicolumn{1}{c|}{C2}                                                                                                          &  0.184 & 0.821 & 0.400 & 0.732 & 0.658 & 0.689 & 0.774 & 0.711 & 0.716 & 0.900 & 0.811 & 0.874 & 0.932 & 0.856 & 0.906 & 0.700 & 0.837 & 0.758 & 0.784 & 0.653 & 0.726 & 0.664 & 0.759    \\
\multicolumn{2}{c}{} & \multicolumn{1}{c|}{C3}                                                                                                          &  0.244 & 0.644 & 0.422 & 0.444 & 0.467 & 0.489 & 0.556 & 0.578 & 0.756 & 0.622 & 0.622 & 0.711 & 0.644 & 0.600 & 0.560 & 0.600 & 0.600 & 0.644 & 0.600 & 0.333 & 0.489 & 0.218 & 0.586    \\ \cmidrule{2-26}
& \multicolumn{1}{c}{\multirow{3}{*}{Location}} & \multicolumn{1}{c|}{C1}                                                                               &  0.214 & 0.717 & 0.434 & 0.711 & 0.887 & 0.610 & 0.635 & 0.811 & 0.723 & 0.912 & 0.836 & 0.893 & 0.912 & 0.693 & 0.781 & 0.767 & 0.792 & 0.824 & 0.818 & 0.635 & 0.730 & 0.673 & 0.682   \\
\multicolumn{2}{c}{} & \multicolumn{1}{c|}{C2}                                                                                                          &  0.222 & 0.861 & 0.389 & 0.444 & 0.500 & 0.556 & 0.667 & 0.667 & 0.667 & 0.917 & 0.667 & 0.639 & 0.833 & 0.750 & 0.812 & 0.417 & 0.583 & 0.472 & 0.500 & 0.194 & 0.389 & 0.134 & 0.370   \\
\multicolumn{2}{c}{} & \multicolumn{1}{c|}{C3}                                                                                                          &  0.025 & 0.800 & 0.325 & 0.700 & 0.725 & 0.275 & 0.600 & 0.875 & 0.900 & 0.850 & 0.900 & 0.875 & 0.900 & 0.767 & 0.800 & 0.975 & 0.825 & 0.825 & 0.850 & 0.675 & 0.800 & 0.646 & 0.689   \\ \cmidrule{2-26}
& \multicolumn{1}{c}{\multirow{3}{*}{Structure}} & \multicolumn{1}{c|}{C1}                                                                              &  0.293 & 0.698 & 0.624 & 0.659 & 0.259 & 0.741 & 0.741 & 0.580 & 0.712 & 0.678 & 0.654 & 0.766 & 0.800 & 0.571 & 0.654 & 0.678 & 0.727 & 0.771 & 0.800 & 0.639 & 0.683 & 0.579 & 0.563   \\
\multicolumn{2}{c}{} & \multicolumn{1}{c|}{C2}                                                                                                          &  0.224 & 0.783 & 0.609 & 0.559 & 0.540 & 0.671 & 0.752 & 0.770 & 0.776 & 0.602 & 0.677 & 0.826 & 0.901 & 0.742 & 0.887 & 0.714 & 0.807 & 0.814 & 0.870 & 0.739 & 0.714 & 0.924 & 0.796   \\
\multicolumn{2}{c}{} & \multicolumn{1}{c|}{C3}                                                                                                          &  0.127 & 0.667 & 0.402 & 0.402 & 0.451 & 0.578 & 0.686 & 0.706 & 0.716 & 0.676 & 0.647 & 0.706 & 0.725 & 0.431 & 0.681 & 0.480 & 0.490 & 0.569 & 0.588 & 0.490 & 0.500 & 0.489 & 0.636   \\
\bottomrule
\end{tabular}

 \caption{Effect of the corruption type on the Page-Level Accuracy by varying in-context learning strategy and complexity level (addressing RQ2). DUDE dataset.}
 \label{tab:ablationDUDEPage}
\end{sidewaystable*}

\begin{sidewaystable*}[h!]
\tiny
\centering
\setlength{\tabcolsep}{3pt}

\begin{tabular}{@{}p{2mm}cccc|c|cc|cc|cc|cc|cc|cc|cc|cc|cc|cc@{}}
\toprule
\multicolumn{3}{c}{}  & \multicolumn{2}{c}{Phi4} & \multicolumn{1}{c}{Molmo} & \multicolumn{2}{c}{Ovis} & \multicolumn{2}{c}{Llama} & \multicolumn{2}{c}{Llava 34B}  & \multicolumn{2}{c}{Gemma 27B} & \multicolumn{2}{c}{Qwen 2.5 7B} & \multicolumn{2}{c}{Qwen 2.5 72B} & \multicolumn{2}{c}{InternVL 3 9B} & \multicolumn{2}{c}{InternVL 3 78B} & \multicolumn{2}{c}{GPT-4.1-mini} & \multicolumn{2}{c}{O3}\\ \midrule
                       
& & 
  & Explicit & \begin{tabular}[c]{@{}c@{}}OCR\\  Explicit\end{tabular}      
  & Explicit   
  & Explicit & \begin{tabular}[c]{@{}c@{}}OCR\\  Explicit\end{tabular}  
  & Explicit & \begin{tabular}[c]{@{}c@{}}OCR \\ Explicit\end{tabular}          
  & Explicit & \begin{tabular}[c]{@{}c@{}}OCR \\ Explicit\end{tabular}       
  & Explicit & \begin{tabular}[c]{@{}c@{}}OCR \\ Explicit\end{tabular}       
  & Explicit & \begin{tabular}[c]{@{}c@{}}OCR\\  Explicit\end{tabular} 
  & Explicit & \begin{tabular}[c]{@{}c@{}}OCR\\  Explicit\end{tabular} 
  & Explicit & \begin{tabular}[c]{@{}c@{}}OCR \\ Explicit\end{tabular}
  & Explicit & \begin{tabular}[c]{@{}c@{}}OCR \\ Explicit\end{tabular}
  & Explicit & \begin{tabular}[c]{@{}c@{}}OCR \\ Explicit\end{tabular}
  & Explicit & \begin{tabular}[c]{@{}c@{}}OCR \\ Explicit\end{tabular}\\ \midrule

\multicolumn{2}{c}{\multirow{3}{*}{$Acc_D$}}   
& \multicolumn{1}{c|}{C1}                                                                                                                                  & 0.044 & 0.348 & 0.358 & 0.221 & 0.451 & 0.314 & 0.387 & 0.309 & 0.289 & 0.402 & 0.377 & 0.500 & 0.598 & 0.613 & 0.632 & 0.255 & 0.328 & 0.275 & 0.402 & 0.294 & 0.363 & 0.186 & 0.137  \\
\multicolumn{2}{c}{} & \multicolumn{1}{c|}{C2}                                                                                                             & 0.028 & 0.259 & 0.329 & 0.189 & 0.224 & 0.322 & 0.350 & 0.441 & 0.329 & 0.420 & 0.301 & 0.497 & 0.573 & 0.538 & 0.685 & 0.259 & 0.336 & 0.175 & 0.434 & 0.259 & 0.280 & 0.168 & 0.126  \\
\multicolumn{2}{c}{} & \multicolumn{1}{c|}{C3}                                                                                                             & 0.034 & 0.254 & 0.305 & 0.271 & 0.220 & 0.373 & 0.322 & 0.322 & 0.271 & 0.305 & 0.305 & 0.441 & 0.559 & 0.576 & 0.695 & 0.153 & 0.203 & 0.136 & 0.441 & 0.169 & 0.254 & 0.068 & 0.068  \\ \midrule
\multirow{10}{*}{\rotatebox[origin=c]{90}{{\scriptsize Document Element}}} & \multicolumn{1}{c}{\multirow{3}{*}{$<$15\%}} & \multicolumn{1}{c|}{C1}        & 0.033 & 0.328 & 0.367 & 0.228 & 0.444 & 0.317 & 0.394 & 0.278 & 0.267 & 0.383 & 0.344 & 0.494 & 0.594 & 0.600 & 0.622 & 0.244 & 0.322 & 0.261 & 0.394 & 0.289 & 0.356 & 0.178 & 0.133  \\
\multicolumn{2}{c}{} & \multicolumn{1}{c|}{C2}                                                                                                             & 0.031 & 0.248 & 0.341 & 0.194 & 0.240 & 0.318 & 0.349 & 0.450 & 0.326 & 0.442 & 0.279 & 0.519 & 0.566 & 0.550 & 0.690 & 0.271 & 0.349 & 0.171 & 0.426 & 0.271 & 0.287 & 0.186 & 0.132  \\
\multicolumn{2}{c}{} & \multicolumn{1}{c|}{C3}                                                                                                             & 0.031 & 0.248 & 0.341 & 0.194 & 0.240 & 0.318 & 0.349 & 0.450 & 0.326 & 0.442 & 0.279 & 0.519 & 0.566 & 0.550 & 0.690 & 0.271 & 0.349 & 0.171 & 0.426 & 0.271 & 0.287 & 0.186 & 0.132  \\ \cmidrule{2-26}
& \multicolumn{1}{c}{\multirow{3}{*}{15\%-25\%}} & \multicolumn{1}{c|}{C1}                                                                                 & 0.000 & 0.100 & 0.020 & 0.020 & 0.040 & 0.040 & 0.060 & 0.120 & 0.080 & 0.120 & 0.100 & 0.040 & 0.080 & 0.120 & 0.140 & 0.040 & 0.040 & 0.060 & 0.040 & 0.020 & 0.040 & 0.040 & 0.020  \\
\multicolumn{2}{c}{} & \multicolumn{1}{c|}{C2}                                                                                                             & 0.000 & 0.064 & 0.032 & 0.032 & 0.000 & 0.032 & 0.032 & 0.064 & 0.032 & 0.000 & 0.064 & 0.000 & 0.064 & 0.032 & 0.128 & 0.000 & 0.000 & 0.000 & 0.064 & 0.000 & 0.032 & 0.000 & 0.000  \\
\multicolumn{2}{c}{} & \multicolumn{1}{c|}{C3}                                                                                                             & 0.000 & 0.064 & 0.032 & 0.032 & 0.000 & 0.032 & 0.032 & 0.064 & 0.032 & 0.000 & 0.064 & 0.000 & 0.064 & 0.032 & 0.128 & 0.000 & 0.000 & 0.000 & 0.064 & 0.000 & 0.032 & 0.000 & 0.000  \\ \cmidrule{2-26}
& \multicolumn{1}{c}{\multirow{3}{*}{$>$25\%}} & \multicolumn{1}{c|}{C1}                                                                                   & 0.063 & 0.147 & 0.126 & 0.063 & 0.210 & 0.105 & 0.105 & 0.147 & 0.147 & 0.147 & 0.210 & 0.231 & 0.231 & 0.231 & 0.210 & 0.126 & 0.147 & 0.126 & 0.189 & 0.147 & 0.168 & 0.084 & 0.063  \\
\multicolumn{2}{c}{} & \multicolumn{1}{c|}{C2}                                                                                                             & 0.000 & 0.112 & 0.075 & 0.037 & 0.037 & 0.149 & 0.149 & 0.112 & 0.149 & 0.112 & 0.187 & 0.149 & 0.261 & 0.187 & 0.187 & 0.075 & 0.112 & 0.112 & 0.187 & 0.075 & 0.075 & 0.000 & 0.037  \\
\multicolumn{2}{c}{} & \multicolumn{1}{c|}{C3}                                                                                                             & 0.000 & 0.112 & 0.075 & 0.037 & 0.037 & 0.149 & 0.149 & 0.112 & 0.149 & 0.112 & 0.187 & 0.149 & 0.261 & 0.187 & 0.187 & 0.075 & 0.112 & 0.112 & 0.187 & 0.075 & 0.075 & 0.000 & 0.037  \\ \midrule
\multirow{10}{*}{\rotatebox[origin=c]{90}{{\scriptsize Layout}}} & \multicolumn{1}{c}{\multirow{3}{*}{$<$4 pages}} & \multicolumn{1}{c|}{C1}               & 0.029 & 0.487 & 0.327 & 0.279 & 0.344 & 0.344 & 0.472 & 0.527 & 0.345 & 0.619 & 0.498 & 0.550 & 0.654 & 0.735 & 0.796 & 0.268 & 0.371 & 0.359 & 0.482 & 0.309 & 0.373 & 0.223 & 0.149  \\
\multicolumn{2}{c}{} & \multicolumn{1}{c|}{C2}                                                                                                             & 0.058 & 0.215 & 0.321 & 0.156 & 0.152 & 0.302 & 0.390 & 0.654 & 0.271 & 0.450 & 0.307 & 0.477 & 0.592 & 0.578 & 0.783 & 0.280 & 0.338 & 0.091 & 0.419 & 0.227 & 0.232 & 0.181 & 0.091  \\
\multicolumn{2}{c}{} & \multicolumn{1}{c|}{C3}                                                                                                             & 0.073 & 0.227 & 0.199 & 0.136 & 0.172 & 0.267 & 0.168 & 0.321 & 0.235 & 0.266 & 0.266 & 0.266 & 0.420 & 0.430 & 0.537 & 0.176 & 0.108 & 0.181 & 0.387 & 0.023 & 0.126 & 0.023 & 0.000  \\ \cmidrule{2-26}
& \multicolumn{1}{c}{\multirow{3}{*}{4-8 pages}} & \multicolumn{1}{c|}{C1}                                                                                 & 0.125 & 0.250 & 0.250 & 0.312 & 0.250 & 0.350 & 0.287 & 0.188 & 0.250 & 0.512 & 0.350 & 0.450 & 0.550 & 0.613 & 0.512 & 0.250 & 0.350 & 0.250 & 0.550 & 0.450 & 0.613 & 0.350 & 0.250  \\
\multicolumn{2}{c}{} & \multicolumn{1}{c|}{C2}                                                                                                             & 0.000 & 0.381 & 0.508 & 0.381 & 0.381 & 0.452 & 0.397 & 0.127 & 0.381 & 0.579 & 0.508 & 0.651 & 0.579 & 0.579 & 0.635 & 0.381 & 0.381 & 0.381 & 0.437 & 0.468 & 0.437 & 0.341 & 0.286  \\
\multicolumn{2}{c}{} & \multicolumn{1}{c|}{C3}                                                                                                             & 0.000 & 0.000 & 0.000 & 0.000 & 0.000 & 0.000 & 0.000 & 0.000 & 0.000 & 0.000 & 0.000 & 0.000 & 0.000 & 0.000 & 0.000 & 0.000 & 0.000 & 0.000 & 0.000 & 0.000 & 0.000 & 0.000 & 0.000  \\ \cmidrule{2-26}
& \multicolumn{1}{c}{\multirow{3}{*}{$>$8 pages}} & \multicolumn{1}{c|}{C1}                                                                                & 0.047 & 0.298 & 0.456 & 0.151 & 0.616 & 0.331 & 0.264 & 0.149 & 0.281 & 0.200 & 0.327 & 0.506 & 0.569 & 0.466 & 0.503 & 0.295 & 0.316 & 0.211 & 0.355 & 0.303 & 0.358 & 0.143 & 0.205  \\
\multicolumn{2}{c}{} & \multicolumn{1}{c|}{C2}                                                                                                             & 0.018 & 0.179 & 0.196 & 0.018 & 0.161 & 0.167 & 0.232 & 0.036 & 0.196 & 0.277 & 0.196 & 0.310 & 0.420 & 0.375 & 0.482 & 0.161 & 0.196 & 0.179 & 0.342 & 0.161 & 0.241 & 0.018 & 0.036  \\
\multicolumn{2}{c}{} & \multicolumn{1}{c|}{C3}                                                                                                             & 0.000 & 0.094 & 0.219 & 0.062 & 0.104 & 0.302 & 0.208 & 0.125 & 0.000 & 0.000 & 0.198 & 0.292 & 0.333 & 0.375 & 0.333 & 0.094 & 0.125 & 0.000 & 0.135 & 0.125 & 0.167 & 0.094 & 0.125  \\ \midrule
\multirow{18}{*}{\rotatebox[origin=c]{90}{{\scriptsize NLP Entity}}} & \multicolumn{1}{c}{\multirow{3}{*}{Numeric}} & \multicolumn{1}{c|}{C1}              & 0.000 & 0.216 & 0.324 & 0.054 & 0.432 & 0.189 & 0.351 & 0.189 & 0.189 & 0.216 & 0.216 & 0.486 & 0.622 & 0.676 & 0.622 & 0.108 & 0.189 & 0.270 & 0.405 & 0.162 & 0.135 & 0.054 & 0.000  \\
\multicolumn{2}{c}{} & \multicolumn{1}{c|}{C2}                                                                                                             & 0.000 & 0.339 & 0.339 & 0.210 & 0.258 & 0.339 & 0.387 & 0.403 & 0.339 & 0.371 & 0.226 & 0.435 & 0.452 & 0.500 & 0.597 & 0.161 & 0.274 & 0.210 & 0.355 & 0.210 & 0.258 & 0.177 & 0.129  \\
\multicolumn{2}{c}{} & \multicolumn{1}{c|}{C3}                                                                                                             & 0.021 & 0.250 & 0.354 & 0.188 & 0.208 & 0.458 & 0.354 & 0.292 & 0.271 & 0.271 & 0.354 & 0.417 & 0.500 & 0.562 & 0.792 & 0.146 & 0.208 & 0.125 & 0.500 & 0.208 & 0.312 & 0.083 & 0.083  \\ \cmidrule{2-26}
& \multicolumn{1}{c}{\multirow{3}{*}{Temporal}} & \multicolumn{1}{c|}{C1}                                                                                  & 0.130 & 0.435 & 0.609 & 0.304 & 0.696 & 0.565 & 0.565 & 0.304 & 0.522 & 0.304 & 0.435 & 0.652 & 0.870 & 0.696 & 0.739 & 0.391 & 0.435 & 0.304 & 0.609 & 0.522 & 0.565 & 0.391 & 0.478  \\
\multicolumn{2}{c}{} & \multicolumn{1}{c|}{C2}                                                                                                             & 0.267 & 0.533 & 0.467 & 0.200 & 0.400 & 0.600 & 0.533 & 0.400 & 0.667 & 0.467 & 0.467 & 0.533 & 0.667 & 0.600 & 0.800 & 0.200 & 0.333 & 0.267 & 0.667 & 0.400 & 0.333 & 0.133 & 0.200  \\
\multicolumn{2}{c}{} & \multicolumn{1}{c|}{C3}                                                                                                             & 0.000 & 0.556 & 0.333 & 0.333 & 0.333 & 0.444 & 0.556 & 0.333 & 0.556 & 0.444 & 0.444 & 0.778 & 0.889 & 0.667 & 0.889 & 0.556 & 0.667 & 0.111 & 0.667 & 0.444 & 0.667 & 0.333 & 0.444  \\ \cmidrule{2-26}
& \multicolumn{1}{c}{\multirow{3}{*}{Misc}} & \multicolumn{1}{c|}{C1}                                                                                      & 0.023 & 0.322 & 0.299 & 0.253 & 0.506 & 0.333 & 0.402 & 0.345 & 0.241 & 0.402 & 0.356 & 0.471 & 0.529 & 0.598 & 0.609 & 0.230 & 0.322 & 0.230 & 0.345 & 0.276 & 0.402 & 0.195 & 0.126  \\
\multicolumn{2}{c}{} & \multicolumn{1}{c|}{C2}                                                                                                             & 0.007 & 0.125 & 0.236 & 0.139 & 0.132 & 0.257 & 0.292 & 0.410 & 0.222 & 0.354 & 0.236 & 0.451 & 0.514 & 0.479 & 0.653 & 0.201 & 0.257 & 0.104 & 0.340 & 0.208 & 0.243 & 0.125 & 0.069  \\
\multicolumn{2}{c}{} & \multicolumn{1}{c|}{C3}                                                                                                             & 0.038 & 0.205 & 0.244 & 0.282 & 0.192 & 0.333 & 0.282 & 0.321 & 0.282 & 0.256 & 0.333 & 0.308 & 0.462 & 0.487 & 0.590 & 0.103 & 0.141 & 0.128 & 0.397 & 0.103 & 0.192 & 0.013 & 0.000  \\ \cmidrule{2-26}
& \multicolumn{1}{c}{\multirow{3}{*}{Location}} & \multicolumn{1}{c|}{C1}                                                                                  & 0.093 & 0.465 & 0.442 & 0.302 & 0.349 & 0.326 & 0.349 & 0.256 & 0.395 & 0.605 & 0.558 & 0.628 & 0.698 & 0.628 & 0.651 & 0.419 & 0.465 & 0.395 & 0.465 & 0.372 & 0.442 & 0.233 & 0.140  \\
\multicolumn{2}{c}{} & \multicolumn{1}{c|}{C2}                                                                                                             & 0.000 & 0.407 & 0.556 & 0.296 & 0.407 & 0.426 & 0.389 & 0.574 & 0.481 & 0.685 & 0.519 & 0.722 & 0.833 & 0.704 & 0.815 & 0.593 & 0.648 & 0.315 & 0.704 & 0.444 & 0.426 & 0.315 & 0.241  \\
\multicolumn{2}{c}{} & \multicolumn{1}{c|}{C3}                                                                                                             & 0.030 & 0.212 & 0.303 & 0.333 & 0.242 & 0.242 & 0.212 & 0.424 & 0.182 & 0.485 & 0.152 & 0.697 & 0.758 & 0.758 & 0.758 & 0.061 & 0.121 & 0.152 & 0.364 & 0.121 & 0.121 & 0.030 & 0.000  \\ \cmidrule{2-26}
& \multicolumn{1}{c}{\multirow{3}{*}{Structure}} & \multicolumn{1}{c|}{C1}                                                                                 & 0.000 & 0.357 & 0.143 & 0.071 & 0.071 & 0.071 & 0.214 & 0.571 & 0.143 & 0.429 & 0.286 & 0.071 & 0.214 & 0.357 & 0.571 & 0.071 & 0.143 & 0.143 & 0.214 & 0.143 & 0.143 & 0.000 & 0.000  \\
\multicolumn{2}{c}{} & \multicolumn{1}{c|}{C2}                                                                                                             & 0.273 & 0.455 & 0.182 & 0.182 & 0.091 & 0.182 & 0.455 & 0.455 & 0.455 & 0.182 & 0.273 & 0.273 & 0.636 & 0.636 & 0.818 & 0.000 & 0.182 & 0.091 & 0.455 & 0.091 & 0.091 & 0.000 & 0.182  \\
\multicolumn{2}{c}{} & \multicolumn{1}{c|}{C3}                                                                                                             & 0.111 & 0.556 & 0.556 & 0.333 & 0.333 & 0.667 & 0.667 & 0.111 & 0.222 & 0.111 & 0.222 & 0.444 & 0.667 & 0.667 & 0.667 & 0.556 & 0.556 & 0.222 & 0.556 & 0.444 & 0.556 & 0.333 & 0.444  \\
\bottomrule
\end{tabular}

 \caption{Effect of the corruption type on the Document-Level Accuracy by varying in-context learning strategy and complexity level (addressing RQ2 and RQ3). MPDocVQA dataset.}
 \label{tab:ablationMPDoc}
\end{sidewaystable*}

\begin{sidewaystable*}[h!]
\tiny
\centering
\setlength{\tabcolsep}{3pt}
 
\begin{tabular}{@{}p{2mm}cccc|c|cc|cc|cc|cc|cc|cc|cc|cc|cc|cc@{}}
\toprule
\multicolumn{3}{c}{}  & \multicolumn{2}{c}{Phi4} & \multicolumn{1}{c}{Molmo} & \multicolumn{2}{c}{Ovis} & \multicolumn{2}{c}{Llama} & \multicolumn{2}{c}{Llava 34B}  & \multicolumn{2}{c}{Gemma 27B} & \multicolumn{2}{c}{Qwen 2.5 7B} & \multicolumn{2}{c}{Qwen 2.5 72B} & \multicolumn{2}{c}{InternVL 3 9B} & \multicolumn{2}{c}{InternVL 3 78B} & \multicolumn{2}{c}{GPT-4.1-mini} & \multicolumn{2}{c}{O3}\\ \midrule
                       
& & 
  & Explicit & \begin{tabular}[c]{@{}c@{}}OCR\\  Explicit\end{tabular}      
  & Explicit   
  & Explicit & \begin{tabular}[c]{@{}c@{}}OCR\\  Explicit\end{tabular}  
  & Explicit & \begin{tabular}[c]{@{}c@{}}OCR \\ Explicit\end{tabular}          
  & Explicit & \begin{tabular}[c]{@{}c@{}}OCR \\ Explicit\end{tabular}       
  & Explicit & \begin{tabular}[c]{@{}c@{}}OCR \\ Explicit\end{tabular}       
  & Explicit & \begin{tabular}[c]{@{}c@{}}OCR\\  Explicit\end{tabular} 
  & Explicit & \begin{tabular}[c]{@{}c@{}}OCR\\  Explicit\end{tabular} 
  & Explicit & \begin{tabular}[c]{@{}c@{}}OCR \\ Explicit\end{tabular}
  & Explicit & \begin{tabular}[c]{@{}c@{}}OCR \\ Explicit\end{tabular}
  & Explicit & \begin{tabular}[c]{@{}c@{}}OCR \\ Explicit\end{tabular}
  & Explicit & \begin{tabular}[c]{@{}c@{}}OCR \\ Explicit\end{tabular}\\ \midrule

\multicolumn{2}{c}{\multirow{3}{*}{$Acc_P$}}   & \multicolumn{1}{c|}{C1}                                                                                & 0.225 & 0.807 & 0.830 & 0.815 & 0.682 & 0.845 & 0.857 & 0.707 & 0.849 & 0.852 & 0.883 & 0.901 & 0.925 & 0.855 & 0.864 & 0.829 & 0.869 & 0.849 & 0.897 & 0.827 & 0.871 & 0.780 & 0.687     \\
\multicolumn{2}{c}{} & \multicolumn{1}{c|}{C2}                                                                                                          & 0.188 & 0.706 & 0.699 & 0.740 & 0.759 & 0.724 & 0.743 & 0.729 & 0.763 & 0.824 & 0.807 & 0.850 & 0.887 & 0.791 & 0.834 & 0.725 & 0.778 & 0.757 & 0.843 & 0.691 & 0.776 & 0.669 & 0.558     \\
\multicolumn{2}{c}{} & \multicolumn{1}{c|}{C3}                                                                                                          & 0.205 & 0.728 & 0.749 & 0.808 & 0.813 & 0.749 & 0.795 & 0.663 & 0.798 & 0.808 & 0.824 & 0.865 & 0.889 & 0.885 & 0.918 & 0.712 & 0.751 & 0.824 & 0.891 & 0.741 & 0.842 & 0.712 & 0.611     \\ \midrule
\multirow{10}{*}{\rotatebox[origin=c]{90}{{\scriptsize Document Element}}} & \multicolumn{1}{c}{\multirow{3}{*}{0}} & \multicolumn{1}{c|}{C1}           & 0.247 & 0.714 & 0.815 & 0.858 & 0.614 & 0.829 & 0.855 & 0.733 & 0.837 & 0.875 & 0.869 & 0.889 & 0.918 & 0.837 & 0.848 & 0.854 & 0.879 & 0.875 & 0.916 & 0.829 & 0.882 & 0.797 & 0.698     \\
\multicolumn{2}{c}{} & \multicolumn{1}{c|}{C2}                                                                                                          & 0.215 & 0.667 & 0.709 & 0.809 & 0.822 & 0.728 & 0.787 & 0.733 & 0.804 & 0.870 & 0.841 & 0.883 & 0.922 & 0.844 & 0.925 & 0.802 & 0.848 & 0.846 & 0.915 & 0.767 & 0.863 & 0.743 & 0.626     \\
\multicolumn{2}{c}{} & \multicolumn{1}{c|}{C3}                                                                                                          & 0.215 & 0.667 & 0.709 & 0.809 & 0.822 & 0.728 & 0.787 & 0.733 & 0.804 & 0.870 & 0.841 & 0.883 & 0.922 & 0.844 & 0.925 & 0.802 & 0.848 & 0.846 & 0.915 & 0.767 & 0.863 & 0.743 & 0.626      \\ \cmidrule{2-26}
& \multicolumn{1}{c}{\multirow{3}{*}{1}} & \multicolumn{1}{c|}{C1}                                                                                      & 0.232 & 0.897 & 0.850 & 0.826 & 0.703 & 0.876 & 0.868 & 0.698 & 0.879 & 0.847 & 0.915 & 0.932 & 0.944 & 0.863 & 0.863 & 0.844 & 0.888 & 0.861 & 0.900 & 0.842 & 0.888 & 0.777 & 0.685     \\
\multicolumn{2}{c}{} & \multicolumn{1}{c|}{C2}                                                                                                          & 0.146 & 0.729 & 0.649 & 0.625 & 0.653 & 0.688 & 0.653 & 0.715 & 0.681 & 0.747 & 0.743 & 0.795 & 0.830 & 0.870 & 0.861 & 0.604 & 0.663 & 0.604 & 0.733 & 0.562 & 0.635 & 0.549 & 0.444     \\
\multicolumn{2}{c}{} & \multicolumn{1}{c|}{C3}                                                                                                          & 0.146 & 0.729 & 0.649 & 0.625 & 0.653 & 0.688 & 0.653 & 0.715 & 0.681 & 0.747 & 0.743 & 0.795 & 0.830 & 0.870 & 0.861 & 0.604 & 0.663 & 0.604 & 0.733 & 0.562 & 0.635 & 0.549 & 0.444     \\ \cmidrule{2-26}
& \multicolumn{1}{c}{\multirow{3}{*}{$>$1}} & \multicolumn{1}{c|}{C1}                                                                                   & 0.131 & 0.841 & 0.822 & 0.636 & 0.841 & 0.804 & 0.832 & 0.650 & 0.799 & 0.794 & 0.827 & 0.850 & 0.888 & 0.860 & 0.870 & 0.701 & 0.776 & 0.729 & 0.822 & 0.771 & 0.785 & 0.734 & 0.659     \\
\multicolumn{2}{c}{} & \multicolumn{1}{c|}{C2}                                                                                                          & 0.186 & 0.837 & 0.814 & 0.756 & 0.779 & 0.826 & 0.814 & 0.756 & 0.814 & 0.837 & 0.837 & 0.860 & 0.895 & 0.724 & 0.762 & 0.721 & 0.791 & 0.791 & 0.826 & 0.709 & 0.779 & 0.674 & 0.570     \\
\multicolumn{2}{c}{} & \multicolumn{1}{c|}{C3}                                                                                                          & 0.186 & 0.837 & 0.814 & 0.756 & 0.779 & 0.826 & 0.814 & 0.756 & 0.814 & 0.837 & 0.837 & 0.860 & 0.895 & 0.724 & 0.762 & 0.721 & 0.791 & 0.791 & 0.826 & 0.709 & 0.779 & 0.674 & 0.570     \\ \midrule
\multirow{7}{*}{\rotatebox[origin=c]{90}{{\scriptsize Layout}}} & \multicolumn{1}{c}{\multirow{3}{*}{In-Page}} & \multicolumn{1}{c|}{C1}                & 0.223 & 0.750 & 0.695 & 0.719 & 0.512 & 0.734 & 0.781 & 0.719 & 0.730 & 0.836 & 0.820 & 0.855 & 0.898 & 0.863 & 0.878 & 0.695 & 0.770 & 0.754 & 0.832 & 0.703 & 0.746 & 0.664 & 0.473 \\
\multicolumn{2}{c}{} & \multicolumn{1}{c|}{C2}                                                                                                          & 0.142 & 0.606 & 0.582 & 0.551 & 0.609 & 0.591 & 0.618 & 0.717 & 0.612 & 0.698 & 0.668 & 0.766 & 0.791 & 0.692 & 0.764 & 0.566 & 0.625 & 0.563 & 0.708 & 0.495 & 0.625 & 0.511 & 0.425     \\
\multicolumn{2}{c}{} & \multicolumn{1}{c|}{C3}                                                                                                          & 0.210 & 0.609 & 0.572 & 0.696 & 0.710 & 0.572 & 0.681 & 0.652 & 0.645 & 0.754 & 0.739 & 0.775 & 0.819 & 0.836 & 0.895 & 0.551 & 0.587 & 0.717 & 0.826 & 0.536 & 0.717 & 0.500 & 0.406     \\ \cmidrule{2-26}
& \multicolumn{1}{c}{\multirow{3}{*}{Out-Page}} & \multicolumn{1}{c|}{C1}                                                                               & 0.225 & 0.818 & 0.856 & 0.833 & 0.714 & 0.866 & 0.872 & 0.705 & 0.872 & 0.856 & 0.894 & 0.910 & 0.930 & 0.852 & 0.858 & 0.855 & 0.888 & 0.868 & 0.909 & 0.850 & 0.895 & 0.802 & 0.728    \\
\multicolumn{2}{c}{} & \multicolumn{1}{c|}{C2}                                                                                                          & 0.218 & 0.770 & 0.774 & 0.861 & 0.855 & 0.809 & 0.823 & 0.737 & 0.859 & 0.904 & 0.896 & 0.904 & 0.949 & 0.929 & 0.933 & 0.827 & 0.876 & 0.880 & 0.929 & 0.815 & 0.872 & 0.770 & 0.642     \\
\multicolumn{2}{c}{} & \multicolumn{1}{c|}{C3}                                                                                                          & 0.202 & 0.794 & 0.847 & 0.871 & 0.871 & 0.847 & 0.859 & 0.669 & 0.883 & 0.839 & 0.871 & 0.915 & 0.927 & 0.947 & 0.947 & 0.802 & 0.843 & 0.883 & 0.927 & 0.855 & 0.911 & 0.831 & 0.726     \\ \midrule
\multirow{18}{*}{\rotatebox[origin=c]{90}{{\scriptsize NLP Entity}}} & \multicolumn{1}{c}{\multirow{3}{*}{Numeric}} & \multicolumn{1}{c|}{C1}           & 0.277 & 0.802 & 0.845 & 0.808 & 0.652 & 0.854 & 0.878 & 0.710 & 0.838 & 0.857 & 0.878 & 0.912 & 0.942 & 0.877 & 0.873 & 0.811 & 0.860 & 0.866 & 0.905 & 0.799 & 0.838 & 0.784 & 0.631      \\
\multicolumn{2}{c}{} & \multicolumn{1}{c|}{C2}                                                                                                          & 0.258 & 0.753 & 0.771 & 0.760 & 0.784 & 0.773 & 0.776 & 0.714 & 0.742 & 0.836 & 0.789 & 0.854 & 0.857 & 0.754 & 0.789 & 0.695 & 0.740 & 0.745 & 0.794 & 0.703 & 0.745 & 0.732 & 0.612      \\
\multicolumn{2}{c}{} & \multicolumn{1}{c|}{C3}                                                                                                          & 0.239 & 0.799 & 0.855 & 0.836 & 0.843 & 0.871 & 0.874 & 0.720 & 0.855 & 0.814 & 0.877 & 0.912 & 0.925 & 0.904 & 0.956 & 0.805 & 0.821 & 0.830 & 0.912 & 0.808 & 0.865 & 0.811 & 0.714      \\ \cmidrule{2-26}
& \multicolumn{1}{c}{\multirow{3}{*}{Temporal}} & \multicolumn{1}{c|}{C1}                                                                               & 0.240 & 0.957 & 0.974 & 0.908 & 0.962 & 0.974 & 0.951 & 0.770 & 0.962 & 0.862 & 0.954 & 0.980 & 0.992 & 0.957 & 0.974 & 0.921 & 0.941 & 0.918 & 0.957 & 0.951 & 0.959 & 0.949 & 0.934      \\
\multicolumn{2}{c}{} & \multicolumn{1}{c|}{C2}                                                                                                          & 0.611 & 0.903 & 0.806 & 0.819 & 0.875 & 0.889 & 0.889 & 0.833 & 0.931 & 0.861 & 0.889 & 0.903 & 0.931 & 0.868 & 0.925 & 0.778 & 0.861 & 0.750 & 0.931 & 0.847 & 0.861 & 0.750 & 0.708      \\
\multicolumn{2}{c}{} & \multicolumn{1}{c|}{C3}                                                                                                          & 0.183 & 0.963 & 0.927 & 0.908 & 0.899 & 0.899 & 0.954 & 0.752 & 0.963 & 0.798 & 0.954 & 0.982 & 0.991 & 0.968 & 0.989 & 0.954 & 0.954 & 0.927 & 0.963 & 0.945 & 0.972 & 0.945 & 0.927      \\ \cmidrule{2-26}
& \multicolumn{1}{c}{\multirow{3}{*}{Misc}} & \multicolumn{1}{c|}{C1}                                                                                   & 0.175 & 0.728 & 0.761 & 0.813 & 0.655 & 0.788 & 0.827 & 0.633 & 0.800 & 0.841 & 0.843 & 0.859 & 0.889 & 0.809 & 0.809 & 0.790 & 0.846 & 0.822 & 0.876 & 0.799 & 0.860 & 0.760 & 0.661      \\
\multicolumn{2}{c}{} & \multicolumn{1}{c|}{C2}                                                                                                          & 0.137 & 0.633 & 0.638 & 0.758 & 0.761 & 0.693 & 0.718 & 0.688 & 0.734 & 0.818 & 0.808 & 0.818 & 0.873 & 0.785 & 0.832 & 0.705 & 0.767 & 0.758 & 0.831 & 0.688 & 0.781 & 0.665 & 0.535      \\
\multicolumn{2}{c}{} & \multicolumn{1}{c|}{C3}                                                                                                          & 0.186 & 0.631 & 0.618 & 0.765 & 0.780 & 0.628 & 0.719 & 0.629 & 0.756 & 0.834 & 0.788 & 0.814 & 0.842 & 0.861 & 0.891 & 0.603 & 0.659 & 0.818 & 0.877 & 0.665 & 0.829 & 0.601 & 0.475      \\ \cmidrule{2-26}
& \multicolumn{1}{c}{\multirow{3}{*}{Location}} & \multicolumn{1}{c|}{C1}                                                                               & 0.226 & 0.770 & 0.840 & 0.741 & 0.486 & 0.819 & 0.794 & 0.733 & 0.827 & 0.864 & 0.901 & 0.909 & 0.926 & 0.813 & 0.825 & 0.852 & 0.868 & 0.823 & 0.881 & 0.782 & 0.856 & 0.613 & 0.490     \\
\multicolumn{2}{c}{} & \multicolumn{1}{c|}{C2}                                                                                                          & 0.151 & 0.832 & 0.817 & 0.670 & 0.724 & 0.735 & 0.742 & 0.846 & 0.839 & 0.828 & 0.828 & 0.943 & 0.964 & 0.849 & 0.881 & 0.857 & 0.885 & 0.789 & 0.939 & 0.659 & 0.824 & 0.627 & 0.556     \\
\multicolumn{2}{c}{} & \multicolumn{1}{c|}{C3}                                                                                                          & 0.158 & 0.526 & 0.705 & 0.705 & 0.674 & 0.632 & 0.600 & 0.505 & 0.526 & 0.716 & 0.589 & 0.779 & 0.832 & 0.750 & 0.781 & 0.484 & 0.589 & 0.611 & 0.747 & 0.495 & 0.568 & 0.505 & 0.337     \\ \cmidrule{2-26}
& \multicolumn{1}{c}{\multirow{3}{*}{Structure}} & \multicolumn{1}{c|}{C1}                                                                              & 0.297 & 0.766 & 0.453 & 0.578 & 0.094 & 0.609 & 0.688 & 0.875 & 0.734 & 0.828 & 0.750 & 0.719 & 0.734 & 0.444 & 0.556 & 0.625 & 0.672 & 0.688 & 0.734 & 0.625 & 0.656 & 0.547 & 0.453     \\
\multicolumn{2}{c}{} & \multicolumn{1}{c|}{C2}                                                                                                          & 0.171 & 0.686 & 0.314 & 0.429 & 0.486 & 0.571 & 0.743 & 0.800 & 0.771 & 0.714 & 0.629 & 0.771 & 0.886 & 0.789 & 0.895 & 0.429 & 0.457 & 0.600 & 0.743 & 0.543 & 0.429 & 0.257 & 0.257     \\
\multicolumn{2}{c}{} & \multicolumn{1}{c|}{C3}                                                                                                          & 0.263 & 0.960 & 0.960 & 0.939 & 0.939 & 0.960 & 0.970 & 0.717 & 0.919 & 0.747 & 0.929 & 0.949 & 0.970 & 0.963 & 0.963 & 0.960 & 0.960 & 0.929 & 0.960 & 0.949 & 0.960 & 0.939 & 0.939     \\
\bottomrule
\end{tabular}

 \caption{Effect of the corruption type on the Page-Level Accuracy by varying in-context learning strategy and complexity level (addressing RQ2). MPDocVQA dataset.}
 \label{tab:ablationMPPage}
\end{sidewaystable*}

\begin{sidewaystable*}[h!]
\tiny
\centering
\setlength{\tabcolsep}{3.1pt}

\begin{tabular}{@{}p{2mm}cccc|c|cc|cc|cc|cc|cc|cc|cc|cc|cc|cc@{}}
\toprule
\multicolumn{3}{c}{}  & \multicolumn{2}{c}{Phi4} & \multicolumn{1}{c}{Molmo} & \multicolumn{2}{c}{Ovis} & \multicolumn{2}{c}{Llama} & \multicolumn{2}{c}{Llava 34B}  & \multicolumn{2}{c}{Gemma 27B} & \multicolumn{2}{c}{Qwen 2.5 7B} & \multicolumn{2}{c}{Qwen 2.5 72B} & \multicolumn{2}{c}{InternVL 3 9B} & \multicolumn{2}{c}{InternVL 3 78B} & \multicolumn{2}{c}{GPT-4.1-mini} & \multicolumn{2}{c}{O3}\\ \midrule
                       
& & 
  & Explicit & \begin{tabular}[c]{@{}c@{}}OCR\\  Explicit\end{tabular}      
  & Explicit   
  & Explicit & \begin{tabular}[c]{@{}c@{}}OCR\\  Explicit\end{tabular}  
  & Explicit & \begin{tabular}[c]{@{}c@{}}OCR \\ Explicit\end{tabular}          
  & Explicit & \begin{tabular}[c]{@{}c@{}}OCR \\ Explicit\end{tabular}       
  & Explicit & \begin{tabular}[c]{@{}c@{}}OCR \\ Explicit\end{tabular}       
  & Explicit & \begin{tabular}[c]{@{}c@{}}OCR\\  Explicit\end{tabular} 
  & Explicit & \begin{tabular}[c]{@{}c@{}}OCR\\  Explicit\end{tabular} 
  & Explicit & \begin{tabular}[c]{@{}c@{}}OCR \\ Explicit\end{tabular}
  & Explicit & \begin{tabular}[c]{@{}c@{}}OCR \\ Explicit\end{tabular}
  & Explicit & \begin{tabular}[c]{@{}c@{}}OCR \\ Explicit\end{tabular}
  & Explicit & \begin{tabular}[c]{@{}c@{}}OCR \\ Explicit\end{tabular}\\ \midrule

\multirow{18}{*}{\rotatebox[origin=c]{90}{{\scriptsize Document Element}}} & \multicolumn{1}{c}{\multirow{3}{*}{Title}} & \multicolumn{1}{c|}{C1}         &  0.500 & 1.000 & 0.750 & 1.000 & 0.500 & 0.750 & 1.000 & 0.750 & 0.750 & 0.750 & 0.750 & 1.000 & 1.000 & 1.000 & 1.000 & 0.500 & 0.750 & 1.000 & 1.000 & 0.500 & 0.750 & 0.550 & 0.788    \\
\multicolumn{2}{c}{} & \multicolumn{1}{c|}{C2}                                                                                                            &  0.000 & 0.750 & 0.125 & 0.500 & 0.500 & 0.375 & 0.625 & 0.500 & 0.625 & 0.750 & 0.625 & 0.750 & 0.875 & 0.917 & 0.917 & 0.375 & 0.500 & 0.750 & 0.625 & 0.250 & 0.250 & 0.236 & 0.190    \\
\multicolumn{2}{c}{} & \multicolumn{1}{c|}{C3}                                                                                                            &  0.000 & 1.000 & 0.500 & 0.500 & 0.500 & 0.500 & 0.500 & 0.500 & 0.500 & 0.500 & 1.000 & 0.500 & 0.500 & 0.500 & 1.000 & 0.500 & 0.500 & 0.500 & 1.000 & 0.500 & 1.000 & 0.565 & 0.971    \\ \cmidrule{2-26}
& \multicolumn{1}{c}{\multirow{3}{*}{Text}} & \multicolumn{1}{c|}{C1}                                                                                     &  0.149 & 0.649 & 0.486 & 0.730 & 0.662 & 0.608 & 0.662 & 0.689 & 0.622 & 0.811 & 0.730 & 0.757 & 0.824 & 0.804 & 0.813 & 0.581 & 0.770 & 0.743 & 0.797 & 0.419 & 0.635 & 0.571 & 0.575    \\
\multicolumn{2}{c}{} & \multicolumn{1}{c|}{C2}                                                                                                            &  0.179 & 0.731 & 0.358 & 0.478 & 0.522 & 0.478 & 0.567 & 0.746 & 0.746 & 0.746 & 0.567 & 0.776 & 0.910 & 0.880 & 0.890 & 0.507 & 0.612 & 0.687 & 0.687 & 0.582 & 0.567 & 0.559 & 0.752    \\
\multicolumn{2}{c}{} & \multicolumn{1}{c|}{C3}                                                                                                            &  0.059 & 0.588 & 0.294 & 0.529 & 0.588 & 0.176 & 0.471 & 0.706 & 0.706 & 0.647 & 0.765 & 0.765 & 0.706 & 0.783 & 0.826 & 0.824 & 0.588 & 0.824 & 0.765 & 0.529 & 0.706 & 0.506 & 0.705    \\ \cmidrule{2-26}
& \multicolumn{1}{c}{\multirow{3}{*}{Figure}} & \multicolumn{1}{c|}{C1}                                                                                   &  0.464 & 0.857 & 0.607 & 0.679 & 0.536 & 0.714 & 0.750 & 0.571 & 0.857 & 0.679 & 0.714 & 0.750 & 0.893 & 0.944 & 0.917 & 0.643 & 0.643 & 0.714 & 0.750 & 0.500 & 0.679 & 0.658 & 0.573    \\
\multicolumn{2}{c}{} & \multicolumn{1}{c|}{C2}                                                                                                            &  0.296 & 0.815 & 0.333 & 0.333 & 0.259 & 0.481 & 0.667 & 0.667 & 0.630 & 1.000 & 0.630 & 0.815 & 0.852 & 0.792 & 0.917 & 0.667 & 0.593 & 0.519 & 0.704 & 0.296 & 0.519 & 0.373 & 0.601    \\
\multicolumn{2}{c}{} & \multicolumn{1}{c|}{C3}                                                                                                            &  0.000 & 0.778 & 0.000 & 0.000 & 0.111 & 0.111 & 0.556 & 0.667 & 0.778 & 0.444 & 0.778 & 0.556 & 0.667 & 0.545 & 0.818 & 0.667 & 0.667 & 0.667 & 0.556 & 0.000 & 0.444 & 0.047 & 0.322    \\ \cmidrule{2-26}
& \multicolumn{1}{c}{\multirow{3}{*}{Table}} & \multicolumn{1}{c|}{C1}                                                                                    &  0.250 & 0.800 & 0.650 & 0.450 & 0.400 & 0.450 & 0.500 & 0.450 & 0.400 & 0.650 & 0.850 & 0.750 & 0.750 & 0.667 & 0.792 & 0.550 & 0.700 & 0.700 & 0.650 & 0.400 & 0.400 & 0.454 & 0.421    \\
\multicolumn{2}{c}{} & \multicolumn{1}{c|}{C2}                                                                                                            &  0.000 & 0.810 & 0.381 & 0.429 & 0.476 & 0.524 & 0.619 & 0.667 & 0.857 & 0.714 & 0.762 & 0.810 & 0.952 & 1.000 & 1.000 & 0.476 & 0.619 & 0.762 & 0.714 & 0.810 & 0.667 & 0.763 & 0.471    \\
\multicolumn{2}{c}{} & \multicolumn{1}{c|}{C3}                                                                                                            &  0.211 & 0.526 & 0.421 & 0.316 & 0.316 & 0.474 & 0.632 & 0.579 & 0.579 & 0.474 & 0.526 & 0.474 & 0.579 & 0.333 & 0.714 & 0.211 & 0.316 & 0.368 & 0.421 & 0.368 & 0.263 & 0.322 & 0.130    \\ \cmidrule{2-26}
& \multicolumn{1}{c}{\multirow{3}{*}{Abandon}} & \multicolumn{1}{c|}{C1}                                                                                  &  0.286 & 0.714 & 0.571 & 0.857 & 0.357 & 0.786 & 0.786 & 0.643 & 0.857 & 0.786 & 0.857 & 0.857 & 0.857 & 0.625 & 0.625 & 0.857 & 0.857 & 0.857 & 0.857 & 0.714 & 0.857 & 0.738 & 0.877    \\
\multicolumn{2}{c}{} & \multicolumn{1}{c|}{C2}                                                                                                            &  0.333 & 0.556 & 0.667 & 1.000 & 0.889 & 0.778 & 0.667 & 0.333 & 0.444 & 0.889 & 0.889 & 0.778 & 1.000 & 0.800 & 0.800 & 0.667 & 1.000 & 0.889 & 0.889 & 0.667 & 0.778 & 0.475 & 0.852    \\
\multicolumn{2}{c}{} & \multicolumn{1}{c|}{C3}                                                                                                            &  0.000 & 0.429 & 0.429 & 0.286 & 0.286 & 0.286 & 0.571 & 0.571 & 0.571 & 0.429 & 0.571 & 0.429 & 0.571 & 0.556 & 0.889 & 0.429 & 0.571 & 0.286 & 0.429 & 0.429 & 0.571 & 0.256 & 0.589    \\ \midrule
\multirow{14}{*}{\rotatebox[origin=c]{90}{{\scriptsize Layout}}} & \multicolumn{1}{c}{\multirow{3}{*}{Top Left}} & \multicolumn{1}{c|}{C1}                &  0.184 & 0.632 & 0.658 & 0.737 & 0.500 & 0.711 & 0.553 & 0.658 & 0.526 & 0.868 & 0.816 & 0.763 & 0.842 & 0.660 & 0.680 & 0.658 & 0.763 & 0.763 & 0.763 & 0.526 & 0.658 & 0.470 & 0.646    \\
\multicolumn{2}{c}{} & \multicolumn{1}{c|}{C2}                                                                                                            &  0.053 & 0.687 & 0.321 & 0.237 & 0.305 & 0.382 & 0.481 & 0.687 & 0.718 & 0.763 & 0.481 & 0.611 & 0.870 & 0.970 & 0.970 & 0.260 & 0.382 & 0.603 & 0.664 & 0.748 & 0.580 & 0.647 & 0.550    \\
\multicolumn{2}{c}{} & \multicolumn{1}{c|}{C3}                                                                                                            &  0.053 & 0.687 & 0.321 & 0.237 & 0.305 & 0.382 & 0.481 & 0.687 & 0.718 & 0.763 & 0.481 & 0.611 & 0.870 & 0.970 & 0.970 & 0.260 & 0.382 & 0.603 & 0.664 & 0.748 & 0.580 & 0.780 & 0.432    \\ \cmidrule{2-26}
& \multicolumn{1}{c}{\multirow{3}{*}{Top Right}} & \multicolumn{1}{c|}{C1}                                                                                &  0.280 & 0.760 & 0.440 & 0.720 & 0.560 & 0.640 & 0.640 & 0.600 & 0.760 & 0.840 & 0.760 & 0.800 & 0.800 & 0.857 & 0.886 & 0.600 & 0.720 & 0.760 & 0.640 & 0.480 & 0.560 & 0.389 & 0.488    \\
\multicolumn{2}{c}{} & \multicolumn{1}{c|}{C2}                                                                                                            &  0.038 & 0.846 & 0.423 & 0.462 & 0.538 & 0.538 & 0.654 & 0.769 & 0.885 & 0.923 & 0.808 & 0.962 & 1.000 & 0.967 & 0.967 & 0.577 & 0.462 & 0.692 & 0.615 & 0.538 & 0.538 & 0.397 & 0.492    \\
\multicolumn{2}{c}{} & \multicolumn{1}{c|}{C3}                                                                                                            &  0.038 & 0.846 & 0.423 & 0.462 & 0.538 & 0.538 & 0.654 & 0.769 & 0.885 & 0.923 & 0.808 & 0.962 & 1.000 & 0.967 & 0.967 & 0.577 & 0.462 & 0.692 & 0.615 & 0.538 & 0.538 & 0.685 & 0.644    \\ \cmidrule{2-26}
 & \multicolumn{1}{c}{\multirow{3}{*}{Bottom Left}} & \multicolumn{1}{c|}{C1}                                                                             &  0.180 & 0.689 & 0.541 & 0.639 & 0.393 & 0.754 & 0.770 & 0.754 & 0.770 & 0.770 & 0.705 & 0.770 & 0.787 & 0.717 & 0.726 & 0.590 & 0.623 & 0.721 & 0.721 & 0.393 & 0.574 & 0.371 & 0.608    \\
\multicolumn{2}{c}{} & \multicolumn{1}{c|}{C2}                                                                                                            &  0.113 & 0.704 & 0.310 & 0.423 & 0.521 & 0.507 & 0.634 & 0.606 & 0.704 & 0.746 & 0.634 & 0.690 & 0.901 & 0.846 & 0.904 & 0.423 & 0.535 & 0.789 & 0.676 & 0.549 & 0.549 & 0.556 & 0.373    \\
\multicolumn{2}{c}{} & \multicolumn{1}{c|}{C3}                                                                                                            &  0.113 & 0.704 & 0.310 & 0.423 & 0.521 & 0.507 & 0.634 & 0.606 & 0.704 & 0.746 & 0.634 & 0.690 & 0.901 & 0.846 & 0.904 & 0.423 & 0.535 & 0.789 & 0.676 & 0.549 & 0.549 & 0.407 & 0.582    \\ \cmidrule{2-26}
& \multicolumn{1}{c}{\multirow{3}{*}{Bottom Right}} & \multicolumn{1}{c|}{C1}                                                                             &  0.426 & 0.902 & 0.574 & 0.721 & 0.803 & 0.590 & 0.803 & 0.590 & 0.689 & 0.721 & 0.803 & 0.738 & 0.869 & 0.939 & 0.949 & 0.639 & 0.770 & 0.787 & 0.820 & 0.574 & 0.721 & 0.519 & 0.683    \\
\multicolumn{2}{c}{} & \multicolumn{1}{c|}{C2}                                                                                                            &  0.326 & 0.930 & 0.628 & 0.442 & 0.581 & 0.605 & 0.744 & 0.814 & 0.698 & 0.884 & 0.884 & 0.930 & 0.930 & 0.782 & 0.836 & 0.791 & 0.651 & 0.698 & 0.674 & 0.535 & 0.581 & 0.546 & 0.514    \\
\multicolumn{2}{c}{} & \multicolumn{1}{c|}{C3}                                                                                                            &  0.326 & 0.930 & 0.628 & 0.442 & 0.581 & 0.605 & 0.744 & 0.814 & 0.698 & 0.884 & 0.884 & 0.930 & 0.930 & 0.782 & 0.836 & 0.791 & 0.651 & 0.698 & 0.674 & 0.535 & 0.581 & 0.420 & 0.643    \\ 
\bottomrule
\end{tabular}

\caption{Effect of the in-page corruption on the Page-Level Accuracy by varying in-context learning strategy and complexity level (addressing RQ2 and RQ3). DUDE dataset.}
\label{tab:AblDUDEInpage}
\end{sidewaystable*}

\begin{sidewaystable*}[h!]
\tiny
\centering
\setlength{\tabcolsep}{3.1pt}

\begin{tabular}{@{}p{2mm}cccc|c|cc|cc|cc|cc|cc|cc|cc|cc|cc|cc@{}}
\toprule
\multicolumn{3}{c}{}  & \multicolumn{2}{c}{Phi4} & \multicolumn{1}{c}{Molmo} & \multicolumn{2}{c}{Ovis} & \multicolumn{2}{c}{Llama} & \multicolumn{2}{c}{Llava 34B}  & \multicolumn{2}{c}{Gemma 27B} & \multicolumn{2}{c}{Qwen 2.5 7B} & \multicolumn{2}{c}{Qwen 2.5 72B} & \multicolumn{2}{c}{InternVL 3 9B} & \multicolumn{2}{c}{InternVL 3 78B} & \multicolumn{2}{c}{GPT-4.1-mini} & \multicolumn{2}{c}{O3}\\ \midrule
                       
& & 
  & Explicit & \begin{tabular}[c]{@{}c@{}}OCR\\  Explicit\end{tabular}      
  & Explicit   
  & Explicit & \begin{tabular}[c]{@{}c@{}}OCR\\  Explicit\end{tabular}  
  & Explicit & \begin{tabular}[c]{@{}c@{}}OCR \\ Explicit\end{tabular}          
  & Explicit & \begin{tabular}[c]{@{}c@{}}OCR \\ Explicit\end{tabular}       
  & Explicit & \begin{tabular}[c]{@{}c@{}}OCR \\ Explicit\end{tabular}       
  & Explicit & \begin{tabular}[c]{@{}c@{}}OCR\\  Explicit\end{tabular} 
  & Explicit & \begin{tabular}[c]{@{}c@{}}OCR\\  Explicit\end{tabular} 
  & Explicit & \begin{tabular}[c]{@{}c@{}}OCR \\ Explicit\end{tabular}
  & Explicit & \begin{tabular}[c]{@{}c@{}}OCR \\ Explicit\end{tabular}
  & Explicit & \begin{tabular}[c]{@{}c@{}}OCR \\ Explicit\end{tabular}
  & Explicit & \begin{tabular}[c]{@{}c@{}}OCR \\ Explicit\end{tabular}\\ \midrule

\multirow{18}{*}{\rotatebox[origin=c]{90}{{\scriptsize Document Element}}} & \multicolumn{1}{c}{\multirow{3}{*}{Title}} & \multicolumn{1}{c|}{C1}         & 0.000 & 0.250 & 0.250 & 0.250 & 0.250 & 0.250 & 0.500 & 0.750 & 0.250 & 0.750 & 0.250 & 0.250 & 0.750 & 0.667 & 0.667 & 0.250 & 0.250 & 0.250 & 0.250 & 0.250 & 0.250 & 0.250 & 0.250    \\
\multicolumn{2}{c}{} & \multicolumn{1}{c|}{C2}                                                                                                            & 0.000 & 0.267 & 0.333 & 0.333 & 0.267 & 0.133 & 0.533 & 0.533 & 0.667 & 0.267 & 0.667 & 0.267 & 0.600 & 0.467 & 0.800 & 0.067 & 0.267 & 0.267 & 0.533 & 0.467 & 0.267 & 0.333 & 0.267    \\
\multicolumn{2}{c}{} & \multicolumn{1}{c|}{C3}                                                                                                            & 0.118 & 0.588 & 0.588 & 0.471 & 0.529 & 0.647 & 0.647 & 0.765 & 0.765 & 0.824 & 0.882 & 0.824 & 0.882 & 0.826 & 0.826 & 0.588 & 0.588 & 0.471 & 0.706 & 0.294 & 0.588 & 0.294 & 0.412    \\ \cmidrule{2-26}
& \multicolumn{1}{c}{\multirow{3}{*}{Text}} & \multicolumn{1}{c|}{C1}                                                                                     & 0.212 & 0.715 & 0.676 & 0.782 & 0.559 & 0.721 & 0.788 & 0.726 & 0.715 & 0.844 & 0.782 & 0.844 & 0.894 & 0.873 & 0.891 & 0.709 & 0.788 & 0.765 & 0.827 & 0.704 & 0.760 & 0.665 & 0.480    \\
\multicolumn{2}{c}{} & \multicolumn{1}{c|}{C2}                                                                                                            & 0.193 & 0.665 & 0.628 & 0.642 & 0.693 & 0.647 & 0.697 & 0.775 & 0.688 & 0.789 & 0.711 & 0.835 & 0.853 & 0.778 & 0.856 & 0.679 & 0.720 & 0.656 & 0.803 & 0.583 & 0.725 & 0.624 & 0.550    \\
\multicolumn{2}{c}{} & \multicolumn{1}{c|}{C3}                                                                                                            & 0.232 & 0.580 & 0.536 & 0.723 & 0.732 & 0.527 & 0.661 & 0.607 & 0.625 & 0.750 & 0.696 & 0.768 & 0.804 & 0.828 & 0.873 & 0.518 & 0.545 & 0.786 & 0.839 & 0.527 & 0.723 & 0.491 & 0.348    \\ \cmidrule{2-26}
& \multicolumn{1}{c}{\multirow{3}{*}{Figure}} & \multicolumn{1}{c|}{C1}                                                                                   & 0.258 & 0.774 & 0.613 & 0.387 & 0.355 & 0.677 & 0.677 & 0.710 & 0.710 & 0.774 & 0.935 & 0.839 & 0.806 & 0.667 & 0.688 & 0.645 & 0.645 & 0.645 & 0.806 & 0.581 & 0.677 & 0.581 & 0.484    \\
\multicolumn{2}{c}{} & \multicolumn{1}{c|}{C2}                                                                                                            & 0.070 & 0.721 & 0.581 & 0.326 & 0.535 & 0.674 & 0.651 & 0.721 & 0.698 & 0.512 & 0.674 & 0.791 & 0.884 & 0.623 & 0.721 & 0.651 & 0.744 & 0.628 & 0.791 & 0.372 & 0.628 & 0.279 & 0.279    \\
\multicolumn{2}{c}{} & \multicolumn{1}{c|}{C3}                                                                                                            & 0.364 & 0.364 & 0.455 & 0.364 & 0.364 & 0.455 & 0.545 & 0.545 & 0.364 & 0.636 & 0.545 & 0.727 & 0.818 & 0.818 & 0.818 & 0.273 & 0.364 & 0.455 & 0.636 & 0.364 & 0.273 & 0.273 & 0.273    \\ \cmidrule{2-26}
& \multicolumn{1}{c}{\multirow{3}{*}{Table}} & \multicolumn{1}{c|}{C1}                                                                                    & 0.103 & 0.931 & 0.862 & 0.655 & 0.759 & 0.828 & 0.897 & 0.759 & 0.897 & 0.828 & 0.966 & 1.000 & 1.000 & 0.942 & 0.942 & 0.655 & 0.793 & 0.724 & 0.966 & 0.793 & 0.793 & 0.759 & 0.621    \\
\multicolumn{2}{c}{} & \multicolumn{1}{c|}{C2}                                                                                                            & 0.046 & 0.369 & 0.492 & 0.446 & 0.477 & 0.385 & 0.369 & 0.446 & 0.323 & 0.554 & 0.554 & 0.569 & 0.569 & 0.518 & 0.542 & 0.077 & 0.262 & 0.108 & 0.354 & 0.323 & 0.308 & 0.292 & 0.154    \\
\multicolumn{2}{c}{} & \multicolumn{1}{c|}{C3}                                                                                                            & 0.043 & 0.783 & 0.783 & 0.609 & 0.652 & 0.826 & 0.739 & 0.870 & 0.696 & 0.826 & 0.913 & 0.870 & 0.957 & 0.861 & 0.972 & 0.739 & 0.826 & 0.435 & 0.739 & 0.565 & 0.739 & 0.565 & 0.609    \\ \cmidrule{2-26}
& \multicolumn{1}{c}{\multirow{3}{*}{Abandon}} & \multicolumn{1}{c|}{C1}                                                                                  & 0.471 & 0.941 & 0.824 & 0.882 & 0.000 & 0.882 & 0.824 & 0.647 & 0.824 & 0.882 & 0.882 & 0.882 & 0.941 & 1.000 & 1.000 & 0.706 & 0.824 & 0.941 & 0.882 & 0.882 & 0.765 & 0.706 & 0.176    \\
\multicolumn{2}{c}{} & \multicolumn{1}{c|}{C2}                                                                                                            & 0.300 & 0.700 & 0.600 & 0.400 & 0.500 & 0.600 & 0.800 & 1.000 & 0.800 & 0.700 & 0.500 & 0.700 & 0.700 & 0.600 & 1.000 & 0.600 & 0.600 & 0.700 & 0.700 & 0.500 & 0.500 & 0.400 & 0.300    \\
\multicolumn{2}{c}{} & \multicolumn{1}{c|}{C3}                                                                                                            & 0.667 & 0.667 & 1.000 & 0.667 & 0.667 & 1.000 & 1.000 & 0.667 & 1.000 & 1.000 & 0.667 & 0.667 & 1.000 & 0.667 & 0.667 & 0.667 & 1.000 & 1.000 & 1.000 & 0.667 & 0.667 & 0.667 & 0.667    \\ \midrule
\multirow{14}{*}{\rotatebox[origin=c]{90}{{\scriptsize Layout}}} & \multicolumn{1}{c}{\multirow{3}{*}{Top Left}} & \multicolumn{1}{c|}{C1}                & 0.152 & 0.780 & 0.644 & 0.674 & 0.682 & 0.689 & 0.765 & 0.780 & 0.788 & 0.833 & 0.818 & 0.818 & 0.856 & 0.857 & 0.883 & 0.644 & 0.765 & 0.614 & 0.780 & 0.712 & 0.735 & 0.652 & 0.500    \\
\multicolumn{2}{c}{} & \multicolumn{1}{c|}{C2}                                                                                                            & 0.102 & 0.408 & 0.453 & 0.445 & 0.524 & 0.427 & 0.455 & 0.654 & 0.448 & 0.584 & 0.647 & 0.599 & 0.610 & 0.526 & 0.613 & 0.298 & 0.416 & 0.259 & 0.497 & 0.411 & 0.469 & 0.387 & 0.296    \\
\multicolumn{2}{c}{} & \multicolumn{1}{c|}{C3}                                                                                                            & 0.102 & 0.408 & 0.453 & 0.445 & 0.524 & 0.427 & 0.455 & 0.654 & 0.448 & 0.584 & 0.647 & 0.599 & 0.610 & 0.526 & 0.613 & 0.298 & 0.416 & 0.259 & 0.497 & 0.411 & 0.469 & 0.387 & 0.296    \\ \cmidrule{2-26}
& \multicolumn{1}{c}{\multirow{3}{*}{Top Right}} & \multicolumn{1}{c|}{C1}                                                                                & 0.420 & 0.820 & 0.760 & 0.760 & 0.520 & 0.780 & 0.900 & 0.760 & 0.860 & 0.900 & 0.880 & 0.920 & 0.940 & 0.911 & 0.924 & 0.760 & 0.800 & 0.840 & 0.900 & 0.800 & 0.840 & 0.820 & 0.740    \\
\multicolumn{2}{c}{} & \multicolumn{1}{c|}{C2}                                                                                                            & 0.324 & 0.794 & 0.735 & 0.696 & 0.716 & 0.765 & 0.784 & 0.637 & 0.716 & 0.784 & 0.706 & 0.882 & 0.902 & 0.827 & 0.860 & 0.608 & 0.745 & 0.588 & 0.794 & 0.627 & 0.667 & 0.569 & 0.578    \\
\multicolumn{2}{c}{} & \multicolumn{1}{c|}{C3}                                                                                                            & 0.324 & 0.794 & 0.735 & 0.696 & 0.716 & 0.765 & 0.784 & 0.637 & 0.716 & 0.784 & 0.706 & 0.882 & 0.902 & 0.827 & 0.860 & 0.608 & 0.745 & 0.588 & 0.794 & 0.627 & 0.667 & 0.569 & 0.578    \\ \cmidrule{2-26}
 & \multicolumn{1}{c}{\multirow{3}{*}{Bottom Left}} & \multicolumn{1}{c|}{C1}                                                                             & 0.347 & 0.810 & 0.777 & 0.793 & 0.421 & 0.777 & 0.785 & 0.686 & 0.752 & 0.826 & 0.818 & 0.884 & 0.934 & 0.918 & 0.927 & 0.736 & 0.818 & 0.793 & 0.843 & 0.760 & 0.818 & 0.678 & 0.364    \\
\multicolumn{2}{c}{} & \multicolumn{1}{c|}{C2}                                                                                                            & 0.096 & 0.664 & 0.528 & 0.504 & 0.600 & 0.624 & 0.608 & 0.728 & 0.616 & 0.600 & 0.592 & 0.728 & 0.800 & 0.685 & 0.751 & 0.600 & 0.616 & 0.512 & 0.704 & 0.424 & 0.584 & 0.576 & 0.488    \\
\multicolumn{2}{c}{} & \multicolumn{1}{c|}{C3}                                                                                                            & 0.096 & 0.664 & 0.528 & 0.504 & 0.600 & 0.624 & 0.608 & 0.728 & 0.616 & 0.600 & 0.592 & 0.728 & 0.800 & 0.685 & 0.751 & 0.600 & 0.616 & 0.512 & 0.704 & 0.424 & 0.584 & 0.576 & 0.488    \\ \cmidrule{2-26}
& \multicolumn{1}{c}{\multirow{3}{*}{Bottom Right}} & \multicolumn{1}{c|}{C1}                                                                             & 0.119 & 0.712 & 0.814 & 0.678 & 0.441 & 0.847 & 0.797 & 0.695 & 0.627 & 0.864 & 0.847 & 0.949 & 0.949 & 0.832 & 0.832 & 0.780 & 0.797 & 0.881 & 0.915 & 0.729 & 0.729 & 0.729 & 0.475    \\
\multicolumn{2}{c}{} & \multicolumn{1}{c|}{C2}                                                                                                            & 0.229 & 0.780 & 0.789 & 0.651 & 0.716 & 0.789 & 0.826 & 0.752 & 0.734 & 0.771 & 0.743 & 0.908 & 0.936 & 0.878 & 0.900 & 0.716 & 0.844 & 0.789 & 0.917 & 0.651 & 0.807 & 0.615 & 0.615    \\
\multicolumn{2}{c}{} & \multicolumn{1}{c|}{C3}                                                                                                            & 0.229 & 0.780 & 0.789 & 0.651 & 0.716 & 0.789 & 0.826 & 0.752 & 0.734 & 0.771 & 0.743 & 0.908 & 0.936 & 0.878 & 0.900 & 0.716 & 0.844 & 0.789 & 0.917 & 0.651 & 0.807 & 0.615 & 0.615    \\ 
\bottomrule
\end{tabular}

 \caption{Effect of the in-page corruption on the Page-Level Accuracy by varying in-context learning strategy and complexity level (addressing RQ2 and RQ3). MPDocVQA dataset.}
 \label{tab:AblMPInpage}
\end{sidewaystable*}

\end{document}